\documentclass[acmtog,nonacm]{acmart} 
\acmSubmissionID{203}
\usepackage{amsmath,tikz}
\usepackage{booktabs} % For formal tables
\usepackage{overpic}
\usepackage{float}
\usepackage{subfig}
\usepackage{multirow}
\usepackage{ulem}
\usepackage{pdfpages}
\acmJournal{TOG}
\setcitestyle{square}

\newcommand{\sysName}{\textit{SketchHairSalon}}
\begin{document}

% Title. 
% If your title is long, consider \title[short title]{full title} - "short title" will be used for running heads.
\title{\sysName: Deep Sketch-based Hair Image Synthesis }

% Authors.
\author{Chufeng Xiao}
\affiliation{%
  \institution{School of Creative Media, City University of Hong Kong}}
\email{chufeng.xiao@my.cityu.edu.hk}

\author{Deng Yu}
\affiliation{%
  \institution{School of Creative Media, City University of Hong Kong}}
\email{deng.yu@my.cityu.edu.hk}

\author{Xiaoguang Han}
\affiliation{%
  \institution{SSE, The Chinese University of Hong Kong, Shenzhen}}
\email{hanxiaoguang@cuhk.edu.cn}

\author{Youyi Zheng}
\affiliation{%
  \institution{State Key Lab of CAD\&CG, Zhejiang University}}
\email{youyizheng@zju.edu.cn}

\author{Hongbo Fu}
\authornote{Corresponding author.}
\affiliation{%
  \institution{School of Creative Media, City University of Hong Kong}}
\email{hongbofu@cityu.edu.hk}

% \footnote{$\dag$ Corresponding author.}

% abstract
\begin{abstract}
Recent deep generative models allow real-time generation of hair images from sketch inputs.
Existing solutions often require a user-provided binary mask to specify a target hair shape. This not only costs users extra labor but also fails to capture complicated hair boundaries. Those solutions usually encode hair structures via orientation maps, which, however, are not very effective to encode complex structures. 
We observe that colored hair sketches already implicitly define target hair shapes as well as hair appearance and are more flexible to depict hair structures than orientation maps. Based on these observations, we present \sysName, a two-stage framework for generating realistic hair images directly from freehand sketches depicting desired hair structure and appearance. At the first stage, we train a network to predict a hair matte from an input hair sketch, with an optional set of non-hair strokes. At the second stage, another network is trained to synthesize the structure and appearance of hair images from the input sketch and the generated matte.
To make the networks in the two stages aware of long-term dependency of strokes, we apply self-attention modules to them. To train these networks, we present {a new dataset containing thousands of annotated hair sketch-image pairs} %a new moderately large dataset, containing diverse hairstyles with annotated hair sketch-image pairs 
and corresponding hair mattes. 
Two efficient methods for sketch completion are proposed to automatically complete repetitive braided parts and hair strokes, respectively, thus reducing the workload of users.
Based on the trained networks and the two sketch completion strategies, we build an intuitive interface to allow even novice users to design visually pleasing hair images exhibiting various hair structures and appearance via freehand sketches.
The qualitative and quantitative evaluations show the advantages of the proposed system over the existing or alternative solutions.
\end{abstract}

\authorsaddresses{Authors’ addresses: Chufeng Xiao, School of Creative Media, City University of Hong Kong, chufeng.xiao@my.cityu.edu.hk; Deng Yu, School of Creative Media, City University  of  Hong Kong,  deng.yu@my.cityu.edu.hk;  Xiaoguang  Han,  SSE,  The  Chinese University of Hong Kong, Shenzhen, hanxiaoguang@cuhk.edu.cn; Youyi Zheng, State Key Lab of CAD\&CG, Zhejiang University, youyizheng@zju.edu.cn; Hongbo Fu, School of Creative Media, City University of Hong Kong, hongbofu@cityu.edu.hk.}

% <concept_id>10010147.10010371.10010372</concept_id>
% <concept_desc>Computing methodologies~Rendering</concept_desc>
% <concept_id>10010147.10010371.10010372.10010374</concept_id>
% <concept_desc>Computing methodologies~Ray tracing</concept_desc>

\ccsdesc[500]{Computing methodologies}
\ccsdesc[500]{Image processing}

\keywords{image-to-image translation, sketch-based image synthesis, hair image synthesis}

\begin{teaserfigure}{
    \centering
    \subfloat{
        \hspace{-2mm}
        \begin{minipage}{\linewidth}
        \includegraphics[width=\linewidth]{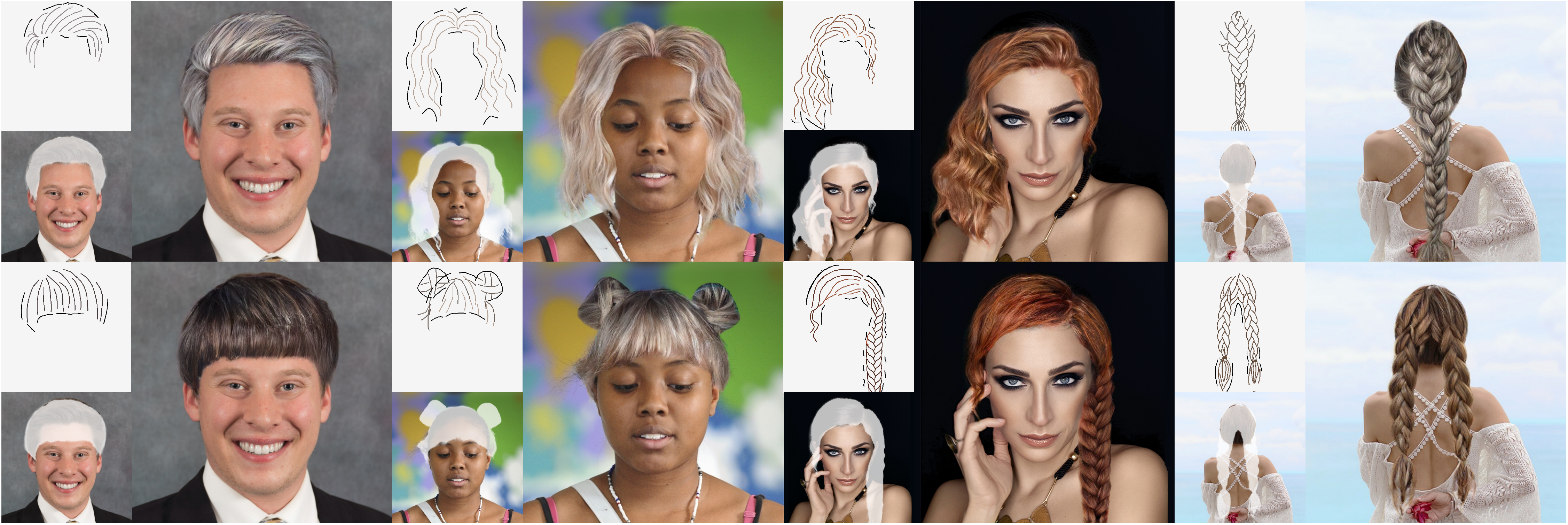}
        \end{minipage}
    }
    \caption{Our \sysName~system allows users to easily create {photo-}realistic hair images with various hairstyles {(e.g., straight, wavy, braided)} from freehand sketches ({Left-Top in each example}), containing colored hair strokes and non-hair strokes (in black). Our two-stage framework automatically generates both hair mattes ({Left-Bottom in each example}) and {hair} images (Right in each example) directly from such sketches. {Original images courtesy of Jacob Rabin, Peteselfchoose, Apostolos Vamvouras, and Ralf21cn.}}
    \label{fig:teaser}
    }
\end{teaserfigure}
\maketitle

\section{Introduction}

% 1. Background
Generating realistic hair images from scratch benefits various applications like hairstyle design, face image generation, portrait manipulation, etc. Sketching provides a simple way to depict complex hair structures as well as local and global features of diverse hairstyles (e.g., straight, wavy, braided). Thus, sketches have been adopted as input by recent sketch-based hair image synthesis techniques \cite{qiu2019two,tan2020michigan,olszewski2020intuitive}, which rely on powerful deep generative models (e.g., \cite{pix2pix2017,wang2018high,SPADE2019CVPR}) to achieve impressive results.

% 2. Problems with existing solutions
These existing hair image synthesis techniques map sparse sketch inputs to orientation maps as a medium fed into image-to-image translation networks. To some extent, such deep networks actually predict texture and shading information from orientation maps to produce realistic images. However, although the use of orientation maps keeps rich local features for hair texture generation, they might wash away some global information, e.g., coherence and occlusion of hair wisps. Thus, the existing techniques based on orientation maps generate less satisfactory results for hairstyles with complex hair structures. 
In addition, these methods require users to provide a binary mask to clearly indicate a target shape, i.e., a hair region within a face image. On the one hand, it is fussy and boring for users to create a hair mask with rich details. On the other hand, a binary mask sets a hard constraint on boundaries of the hair region, and thus is difficult to capture soft boundaries of hair, which naturally diffuses to the background in real hair images.

% 3. Motivation and key ideas
To address these issues, our key idea is to force deep networks to directly focus on sketch inputs, since we observe that a hair sketch itself contains enough information to depict the structure, appearance, and shape of a desired hairstyle on the local and global levels.
For example, for a wavy hairstyle, one single stroke is able to represent one local {and} coherent hair wisp, while two strokes can be used to form a T-junction and represent occlusion that one hair wisp locally occludes {the other} wisp.
Colored strokes are able to indicate the local appearance of a hair image. Also, a sketch depicting the structure of a hairstyle already implicitly defines the global shape of a hair region. We  believe that it is better to automatically infer local and soft details along {boundaries of hair regions}, %the boundary of the hair region, 
since such details are difficult and time-consuming to specify by users. In this case, a hair matte is more appropriate than a binary mask to describe the hair region due to its support for soft boundaries. 

% 4. Method
With the above key observations, we present \sysName, a novel deep generative framework for synthesizing realistic hair images directly from {a small set of colored strokes}. We focus on the generation of 2D hair geometry (i.e., %its 
shape and structure) {and appearance}. It consists of two key stages: \textit{sketch-to-matte generation} and \textit{sketch-to-image generation}.
The first stage focuses on hair matte generation from an input hair sketch, in order to reduce the ambiguity of sketch-to-hair generation. Users may optionally input non-hair strokes, which are used as extra conditions to guide matte generation. The second stage manages to synthesize a {photo-}realistic hair image, given the input sketch and the generated hair matte.
Due to the long-range dependency among hair strokes, we apply self-attention modules \cite{fu2019dual} to the networks in both stages for learning more correspondence among strokes.
Colored strokes are used to control the local appearance of hair images at the second stage. To train the networks in the two stages, we present a new hair sketch-image dataset, which contains thousands of hair images with the corresponding manually annotated hair sketches to depict the underlying hair structures. Each hair image is also associated with an automatically generated hair matte.

% 5. Evaluation
Based on the trained networks, we provide an intuitive interface for users to design their desired hairstyles over human face images, purely via sketch inputs. We present two efficient methods to respectively auto-complete repetitive unbraided strokes and braided knots given user-drawn strokes, thus reducing the workload of users. Our system is able to produce photo-realistic and high-quality hair images ($512 \times 512$) with most of the common hairstyles {that can be} easily depicted by sketches, including straight, wavy, braided, as well as their combinations (Figure \ref{fig:teaser}). Extensive experiments have been conducted to evaluate our dataset and networks in comparison with the existing and alternative ones, both qualitatively and quantitatively. The results show that our proposed method outperforms the other solutions in terms of both visual naturalness and faithfulness to hair sketches.
Also, the usability and controllability of our system are confirmed through two user studies. 

In summary, this work makes the following main contributions: 
\begin{itemize}
    \item A novel system for non-professional users to create photo-realistic hair images with various hair geometry, structure, and appearance from freehand sketches; 
    \item A two-stage framework for first generating hair mattes from sketches and then synthesizing hair images from sketches and generated mattes; 
    \item An intuitive sketch-based interface for hairstyle design, with two novel hair sketch auto-completion features; 
    \item A new dataset containing {4,500} % moderately large dataset of 
    hair sketch-image pairs, with the corresponding hair mattes. 
\end{itemize}
 \section{Related Work}
In this section, we review prior works that are closely related to our method, namely, hair modeling and rendering, conditional hair image synthesis, as well as sketch-based hair image generation. 

\subsection{Hair Modeling and Rendering}
Hair modeling {has been extensively explored} in the computer graphics community. 
Most previous works have focused on reconstructing 3D hair models from either real images \cite{wei2005modeling,jakob2009capturing,zhou2018hairnet,zhang2018modeling,yang2019dynamic} or user-specified sketches \cite{mao2004sketch,fu2007sketching,hu2015single,shen2020deepsketchhair}.
For example, given a single-view hair image for hair modeling, Chai et al. \shortcite{chai2013dynamic} utilize a few strokes to guide hair directions to reduce the ambiguity of growing hair strands from {the hair image}. 
Hu et al. \shortcite{hu2014capturing} present a data-driven method targeted at braided hairstyle reconstruction, through data fitting with a database of parametric braid models. It inspires us to deform the parametric models following sparse user inputs to auto-complete braided %\hbc{braid or braided? to be consistent across the paper}\cfc{braided. Made them consistent for sketches and images.} 
hair sketches, which are tedious to create from scratch.

To make users manipulate hair models freely, Xing et al. \shortcite{xing2019hairbrush} provide an interactive system for authoring 3D hair structures in virtual reality. Due to the requirement of 3D inputs, it is difficult to apply their technique to our problem.
Shen et al. \shortcite{shen2020deepsketchhair} introduce a deep learning based framework for strand-level hair modeling based on 2D sketches to produce plausible 3D hairstyles.
Although their method can generate high-quality results with realistic appearance and layering effects to some extent, mainly due to the use of 2D and 3D orientation fields as intermediate hair representations, their method is not capable of modeling hairstyles with complex structures like braided hairstyles. In addition, an additional step for re-rendering 3D hairstyles back to 2D hair images is needed to extend these approaches to our task. Several neural rendering methods like \cite{wei2018real,chai2020neural} adopt similar ideas to control hair image generation based on hair rendering and image encoding, i.e., 3D models for structures (via orientation maps) and reference images for appearance.

\subsection{Conditional Hair Image Synthesis}
Recent deep generative models like conditional Generative Adversarial Networks (GANs) show amazing ability on image generation based on different types of inputs. 
For example, Isola et al. \shortcite{pix2pix2017} introduce a general framework with conditional GANs for image-to-image translation. 
Park et al. \shortcite{SPADE2019CVPR} propose spatially-adaptive normalization, a simple but effective layer for realistic image synthesis given a sparse semantic layout. 
These impressive models have been widely applied to hair image generation given various inputs. 
Extended from the pix2pix %\hbc{in the paper, sometimes you used Pix2Pix and sometimes pix2pix; be consistent} %\cfc{fixed and double-checked the others.} 
method \cite{pix2pix2017}, our solution addresses the problem of sketch-based hairstyle design by using a novel two-stage framework for sketch-to-matte generation and sketch-to-image generation. 

Diverse types of inputs (e.g., abstract attribute{s}, guided mask{s}, reference{s}, sketches, etc.) have been used as conditions for hair image generation.
Many attribute-level portrait editing methods (e.g., \cite{choi2018stargan,lample2017fader,xiao2018elegant}) seek to disentangle semantic vectors and then interpolate them in semantic directions in the latent space.
StyleGAN \cite{karras2019style} further improves the generation quality significantly by adding an intermediate latent space to lessen the entangled degree of the single input latent space. 
Such attribute-conditioned methods can only provide users with high-level control on hairstyles due to their highly abstract definitions. Thus they are not suitable for our task, which aims for precise control on hair geometry and appearance during the design process.

Several mask-level methods provide global-shape
control for portrait image manipulation via semantic label masks, including one for the hair region. For example,
Gu et al. \shortcite{gu2019mask} divide face images into five components and separately encode their embeddings for face image editing and component swapping. 
MaskGAN \cite{lee2020maskgan} is proposed to learn face manipulation over mask manifold from exemplars. Zhu et al. \shortcite{zhu2020sean} introduce semantic region-adaptive normalization for GANs to control styles in a specified mask region. These methods require users to provide masks for editing and transferring indicated components of human faces. The hair masks used in these works trim hair regions with hard boundaries, and thus not very effective for synthesizing hair strands naturally diffusing to the background in real images. We show that it is possible to automatically infer desired hair mattes (with soft boundaries) from strokes depicting hair structures.

\subsection{Sketch-based Hair Image Generation}
Sketch-based interfaces have been explored to achieve fine-grained geometry control for hair image generation and editing. For example, Chen et al. \shortcite{chen2006generative} introduce a 2D generative sketch model for hair analysis and synthesis. 
However, their method requires a {large} set of manually-defined parameters for extracting {a} sketch representation over given hair images.
DeepFaceDrawing \cite{chenDeepFaceDrawing2020} takes sketch inputs as soft constraints for retrieval and interpolation upon the sketch level, while DeepFaceEditing \cite{DFE2021} sets sketches as an intuitive geometry representation to edit face images, with decoupled control of their geometry and appearance. Although they perform well for synthesizing main face components such as mouth and eyes, the two approaches do not work well for synthesizing hairstyles mainly due to their complex structures. 

Olszewski et al. \shortcite{olszewski2020intuitive} introduce a two-stage deep framework mainly for realistic synthesis and editing of facial hair, i.e., beard. To train the {two networks in their framework} with sketch inputs, they randomly produce strokes following orientation maps extracted from hair images. However, unlike beard, scalp hair images have more complex shapes and global structures, and their underlying strokes are hard to automatically extract from real images (Figure \ref{fig:dataset} (c)). Thus, we believe there is a domain gap between automatically extracted strokes from hair images and manually annotated sketches (see comparisons in Section \ref{exp:ablation}). 
Qiu et al. \shortcite{qiu2019two} also use orientation maps as a medium between two phases for hair image synthesis conditioned on sketches.
MichiGAN \cite{tan2020michigan} proposes a disentangling manner for portrait hair manipulation based on multiple inputs, of which smoothed orientation maps are used to control local hair structures. They provide an interface for users to manipulate hair structures by editing orientation maps via stroke inputs.
However, for complex hairstyles, orientation maps easily wash away some global structures, e.g., coherence and occlusion of hair wisps (Figure \ref{fig:ablation_input}).
To address this issue, we present a new dataset of hair images with manually annotated hair sketches and propose to learn hair images with complex hairstyles directly from sketches. In addition, the above methods require users to explicitly specify hair regions, and this process is usually time-consuming, especially when desired hairstyles are complicated. Our approach automatically infers hair mattes from hair sketches.

\begin{figure}[htb]{
    \includegraphics[width=.99\linewidth]{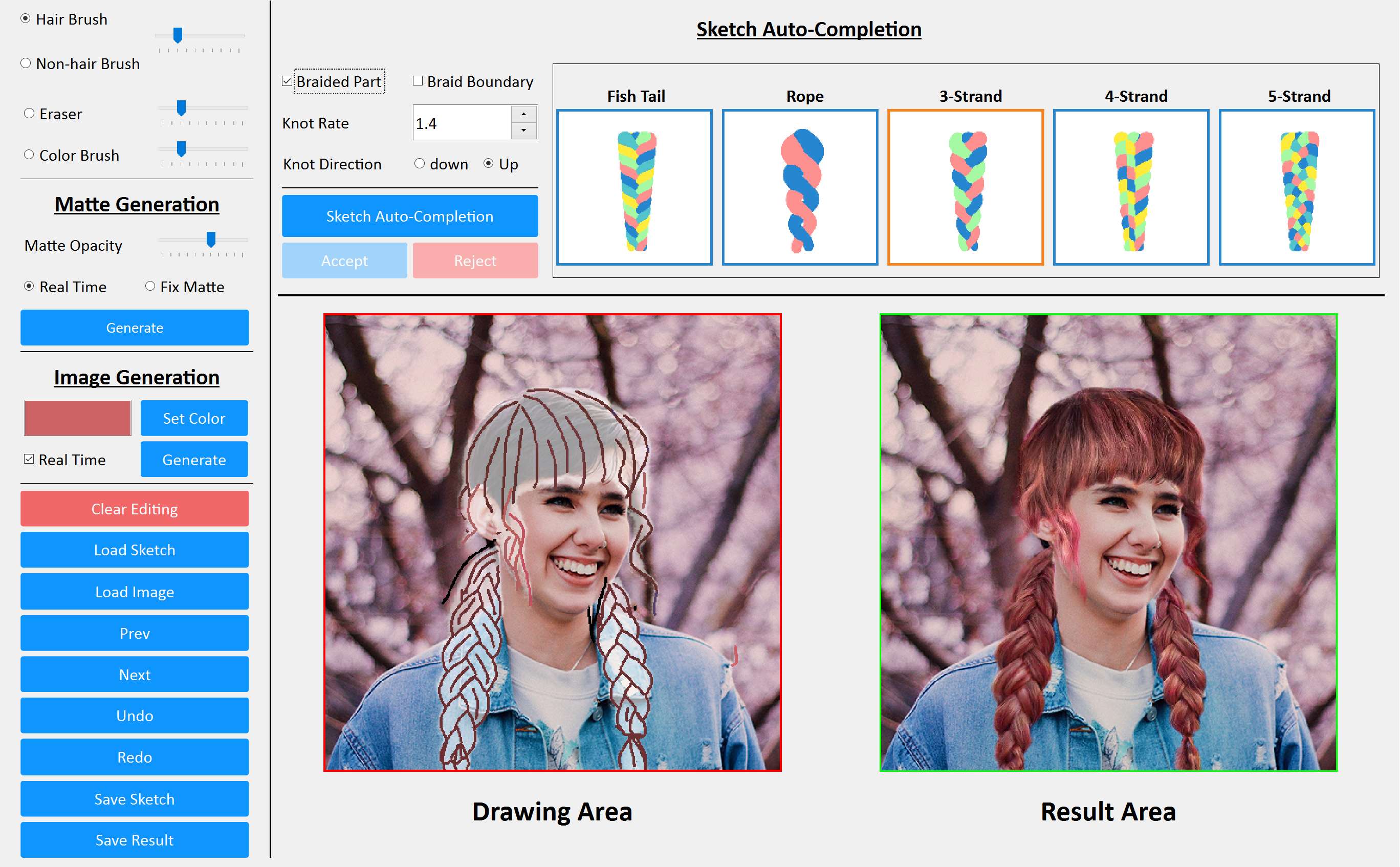}
    \caption{A screenshot of our sketching interface for realistic hairstyle design. The colored lines indicate hair strokes, while the black ones represent non-hair strokes. This example includes both the braided and unbraided parts.  
    Note that the braid sketch is automatically generated given a small number of user-specified strokes.
    Both the sketch-to-matte and sketch-to-image stages
    are completely controlled via sketching.
    {Original image courtesy of Elijah M. Henderson.}
    }
    \label{fig:interface}
}
\end{figure}
\section{User Interface}
\label{sec:interface}
The main goal of this work is to develop a sketch-based system for realistic hairstyle design. We first introduce our interface (as shown in Figure \ref{fig:interface}) from a user's perspective in this section and will describe the underlying {algorithm} in Section~\ref{sec:alg}. Our interface is designed based on a key observation that a hairstyle can be determined by three main factors: shape, structure, and appearance, which can be specified all via colored sketches. To use our system, the user can firstly load a portrait image and then do sketching surrounding {its} head region for hairstyle design. Our system will produce {an updated hair matte and hairstyle image} in an interactive rate according to each interaction. The details are described in the following aspects.   

{\paragraph{Hair Structure Specification.} The user draws \textit{hair strokes} to indicate the structure of a desired hairstyle.} Our system supports diverse structures, such as straight, wavy, braided, and their combinations.

\paragraph{Hair Shape Refinement.} 
As discussed previously, we prefer to use a hair matte to accurately represent the shape of a hair region. Given the hair strokes for depicting a desired hair structure, our system reuses them for automatically inferring a hair matte. The inferred matte is displayed semi-transparently over the image. The user can \textit{optionally} use \textit{non-hair strokes} (e.g., the black strokes in Figure \ref{fig:teaser}), which are located outside and surrounding the desired hair region, to refine the matte generation result. The hair and non-hair strokes can be drawn in an arbitrary order. After the matte is generated, the user can fix it and draw hair strokes only to control the structure of a desired hairstyle. 
Note that, hair structure specification and hair shape refinement can be interweaved to polish a desired hairstyle iteratively. 

\paragraph{Hair Appearance Specification.} Users are allowed to choose specific stroke colors to control the hair appearance when drawing individual hair strokes. They may also use \textit{color brush} in the interface to re-colorize the strokes for hair dyeing. We observe that the color space of realistic hair images is limited to a certain space, meaning that not all the RGB colors are valid for synthesizing photo-realistic results. To guarantee the generation quality, we set up a retrieval way to pick a valid color close to a user's choice. Specifically, we first create a hair color database by collecting all the stroke colors in our hair sketch-image dataset, totally over 70k colors. Each stroke's color is {calculated as} the average color of all the pixels along the stroke in its associated hair image. When the user chooses one color, our system finds 20 nearest colors to that by conducting a KNN search in the CIELab space. Each stroke is then randomly assigned {one} of the top-20 colors at each input from the user. This simple mechanism helps highlight the junction structures, where occlusion relations can be inferred from two or more different color strokes, for hair image generation.

\paragraph{Sketch Auto-completion.} To reduce users' load, considering some repetitive patterns of hair sketches, we propose two sketch auto-completion methods for braided and unbraided hairstyles (Section \ref{sec:auto-comp}). For unbraided sketches, we derive %the 
extra strokes by diffusing the existing strokes. For braided sketches, we first construct five parametric braided models inspired by Hu et al. \shortcite{hu2014capturing}, then deform them to follow the user-drawn strokes indicating a rough shape of a desired braid, and finally extract the colored strokes. During interaction, users can separately let the interface auto-complete braided and unbraided sketches by selecting a specific auto-completion mode. Our interface provides a completion suggestion, and then  users can choose to accept, reject, or refine the automatically inferred sketches.

\section{Dataset Preparation}
\label{sec:data}
To train our model in a supervised manner, we need a considerably large dataset containing hair sketch-image pairs with diverse hairstyles. Unfortunately, there exists no such dataset with manually annotated hair sketches.
Qiu et al. \shortcite{qiu2019two} present a small-scale hair sketch-image dataset, including 640 pairs. However, its scale is not big enough and it only covers limited types of hairstyles (mainly straight hairstyle).

\begin{figure}
\centering
    \setlength{\fboxrule}{0.1pt}
    \setlength{\fboxsep}{-0.01cm}
    % \hspace{-2mm}
    \begin{tabular}{c}
    \vspace{-1mm}
            \hspace{-3mm}
            {\includegraphics[width=0.248\linewidth]{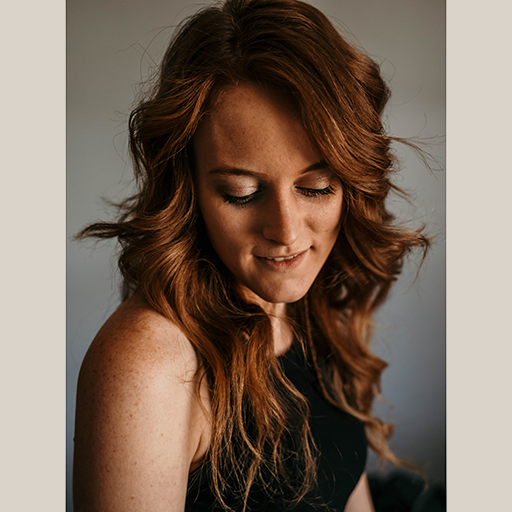}}
            \hspace{-1.6mm}
            {\includegraphics[width=0.249\linewidth]{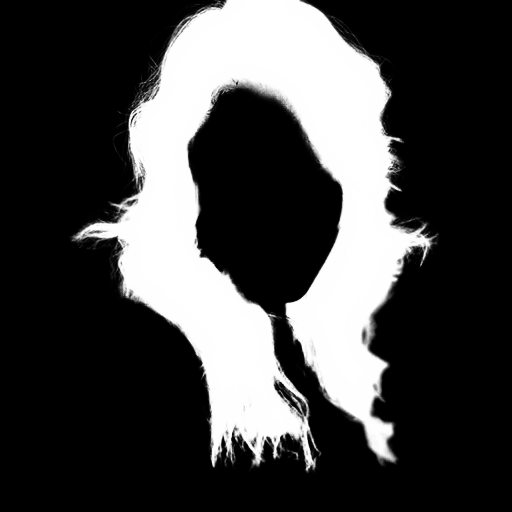}}
            \hspace{-1.6mm}
            {\includegraphics[width=0.249\linewidth]{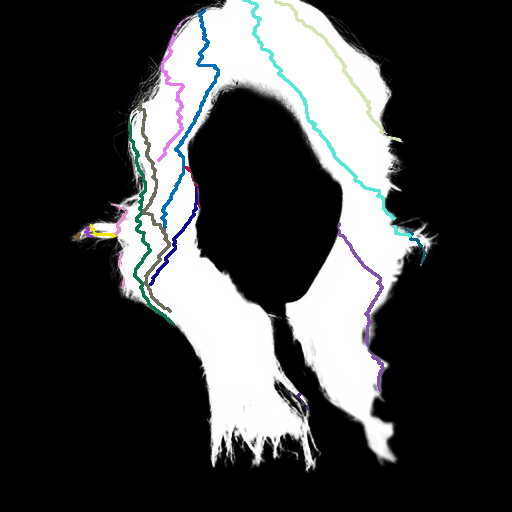}}
            \hspace{-1.6mm}
            {\includegraphics[width=0.249\linewidth]{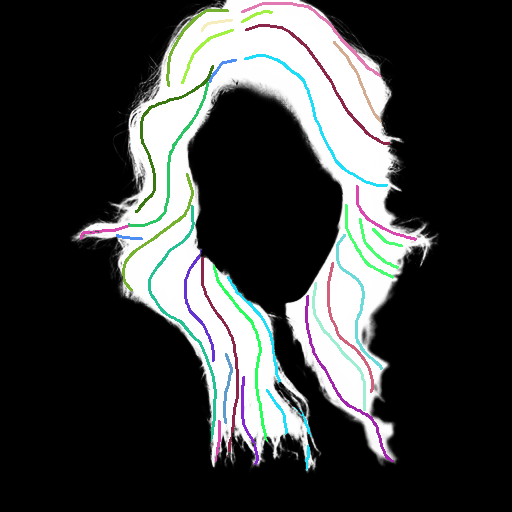}} \\
        
            \hspace{-3mm}
            {\includegraphics[width=0.249\linewidth]{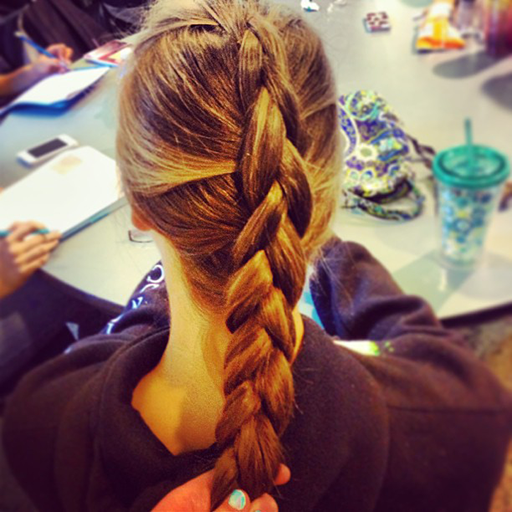}} 
            
            \hspace{-1.5mm}
            {\includegraphics[width=0.249\linewidth]{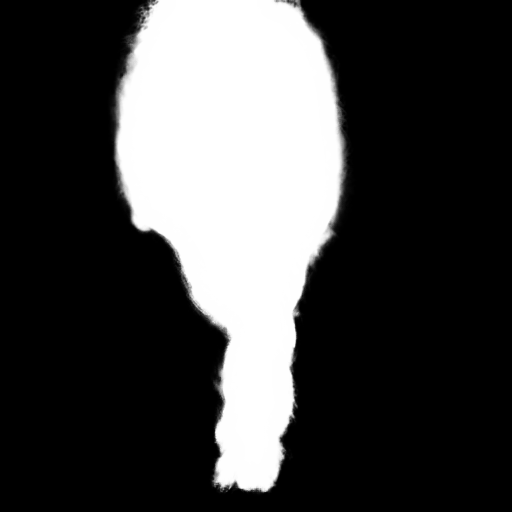}} 
            
            \hspace{-1.6mm}
            {\includegraphics[width=0.249\linewidth]{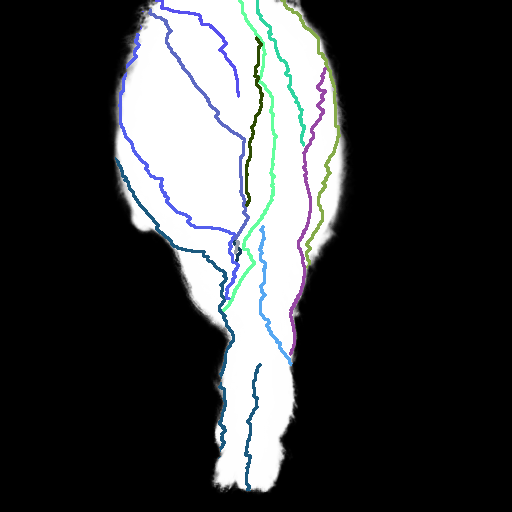}}
            
            \hspace{-1.6mm}
            {\includegraphics[width=0.249\linewidth]{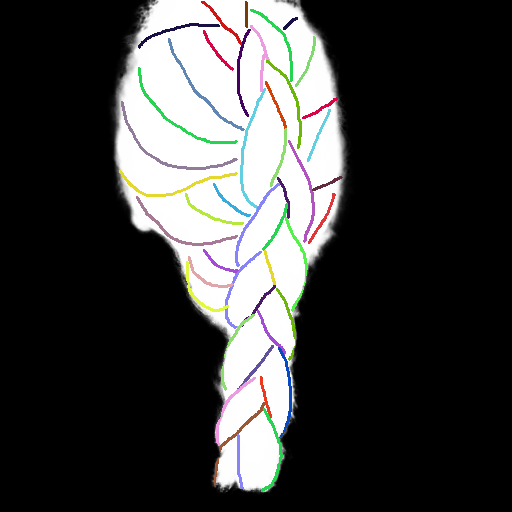}}\\
            \vspace{4mm}
             (a) \qquad \qquad \quad (b) \quad \qquad \qquad (c) \qquad \qquad \qquad (d) \\
    \end{tabular}
    \vspace{-4mm}
    \caption{Illustration of hair matte and hair sketch preparation. Given {each} hair image (a), {its} hair matte (b) is produced via the hair segmentation method \cite{muhammad2018hair}{, followed by} the image matting algorithm \cite{li2020natural}. Upon the matted image (b), (d) is our manually drawn sketch {depicting the hair structures}, while (c) is generated by a synthetic approach \cite{olszewski2020intuitive}. {Original images courtesy of Nicole Geri and Wicker Paradise.}}
    \label{fig:dataset}
\end{figure}

Alternatively, it is possible to automatically extract hair strokes from hair images. 
For example, Olszewski et al. \shortcite{olszewski2020intuitive} propose an automatic stroke extraction approach for facial hair (i.e., beard). It simulates {user-drawn} strokes by tracing from randomly sampled points {and} following orientation maps filtered from hair images. 
MichiGAN \cite{tan2020michigan} adopts a similar approach to prepare their training data for the sketch-to-orientation inpainting task.
However, unlike facial hair images, scalp hair images in our scenario have larger hair regions and involve more complex structures such as wavy and braided hairstyles. The automatically extracted strokes from orientation maps might not faithfully respect the underlying structures of hair images, especially for braided styles (see Figure \ref{fig:dataset} (c-bottom)).  
It is expected that utilizing such synthesized sketches and the counterpart hair images might confuse our hair image generation network and lead to ambiguous hair generation (Figure \ref{fig:ablation_variants} (b)), caused by the inconsistency between the synthesized sketches and hair images. 

To address this issue, we construct a hair sketch-image dataset with manually sketch annotation. Specifically, we first collect thousands of high-quality portrait images (resized to $512 \times 512$) containing hair regions from the Internet, covering the front-view, side-view, and back-view.
The hair regions are then segmented coarsely via a hair segmentation method by Muhammad et al. \shortcite{muhammad2018hair}. 
To capture the soft boundaries of hair regions, we adopt hair mattes to indicate the shape of hairstyles {and automatically extract hair mattes (Figure \ref{fig:dataset} (b)) from the segmented hair regions using a state-of-the-art image matting approach \cite{li2020natural}}.

Given the matted hair images, we asked three {student helpers} to help manually trace a {sparset set of} strokes on top of each hair image, with each stroke representing a wisp of hair strands and the strokes together capturing the structure of the {underlying} hairstyle, such as curling, occluding, etc.
Note that for braided hairstyles, this tracing manner leads to extracting the edges of each braided wisp (Figure \ref{fig:dataset} (d-bottom)).
Since our sketch is a stroke-based representation, we may assign different colors to individual strokes to capture and specify local appearance of a hairstyle. For training, we  derive the color of each stroke from the corresponding hair image.
Compared to the synthetic sketches \cite{olszewski2020intuitive} (e.g., those in Figure \ref{fig:dataset} (c)), our manually drawn sketches (Figure \ref{fig:dataset} (d)) capture the underlying structure of each hair image more faithfully.
We will compare image generation results based on our presented dataset and pairs of synthetic sketch-image pairs in Section \ref{exp:ablation}.

In total, we create 4K sketch-image pairs (including 3K {for} non-braided type and 1K {for} braided type) for training, and 500 pairs (including 400 {for} non-braided type and 100 {for} braided type) for testing. Each pair is associated with a corresponding hair matte. To improve the generalization ability of our system, we augment the training data by translating each hair region, rotating {it} around its center in the range of [$-15^\circ$,$15^\circ$], and horizontal flip. Please find more examples and statistic{s} about the dataset in the supplemental materials.

\begin{figure}
\centering
    \setlength{\fboxrule}{0.1pt}
    \setlength{\fboxsep}{-0.01cm}
    % \hspace{-2mm}
    \begin{tabular}{c}
    \hspace{-3mm}
            \subfloat[]{\framebox
            {\includegraphics[width=0.249\linewidth]{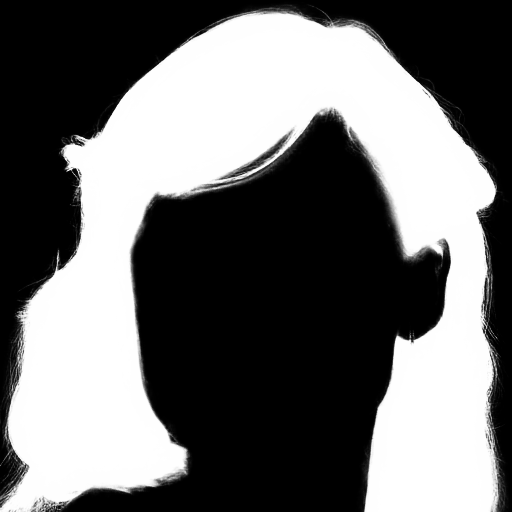}} 
            }
            \hspace{-1.5mm} 
            \subfloat[]
            {\includegraphics[width=0.249\linewidth]{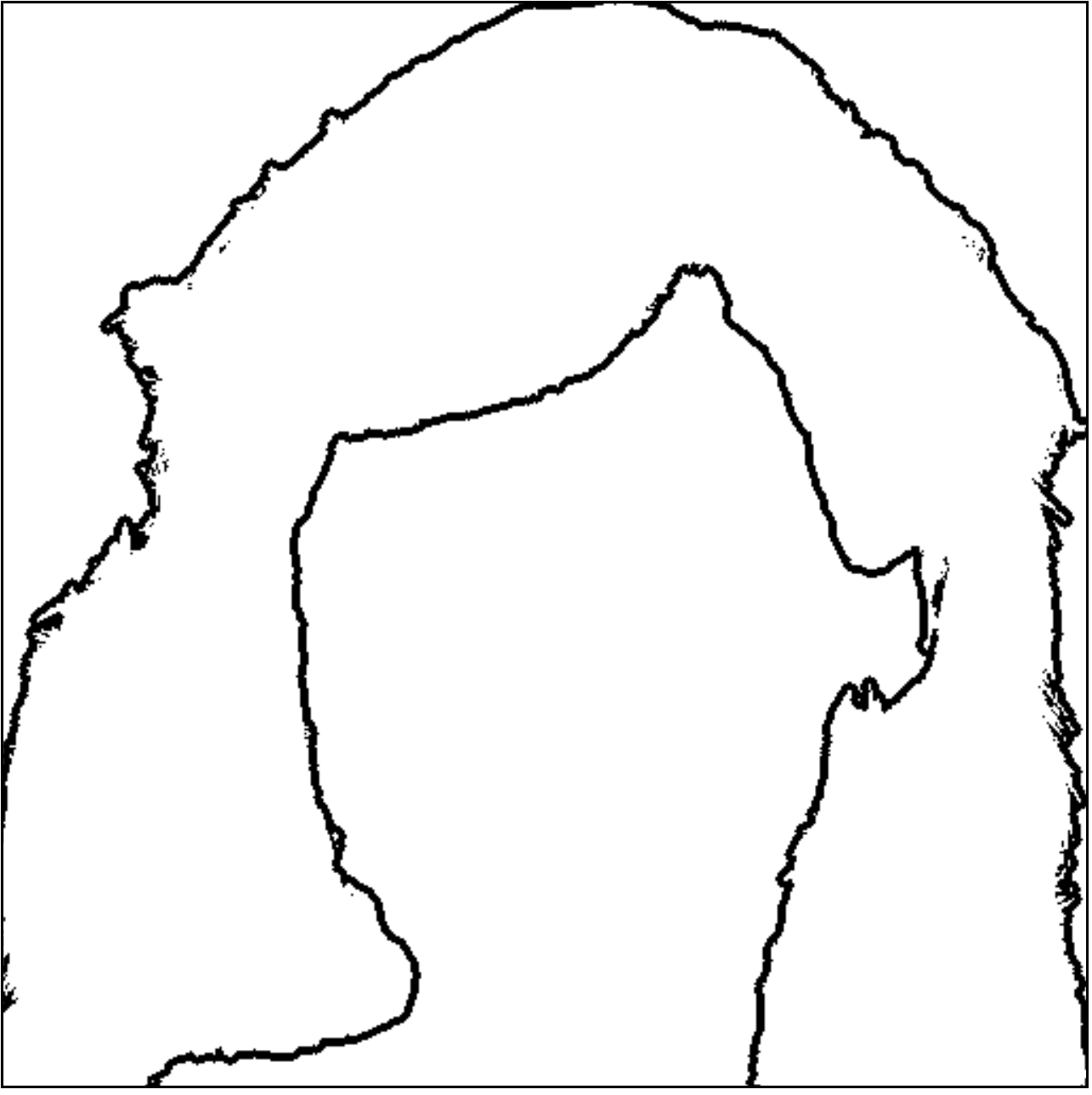}} 
            \hspace{-1.57mm}
            \subfloat[]
            {\includegraphics[width=0.249\linewidth]{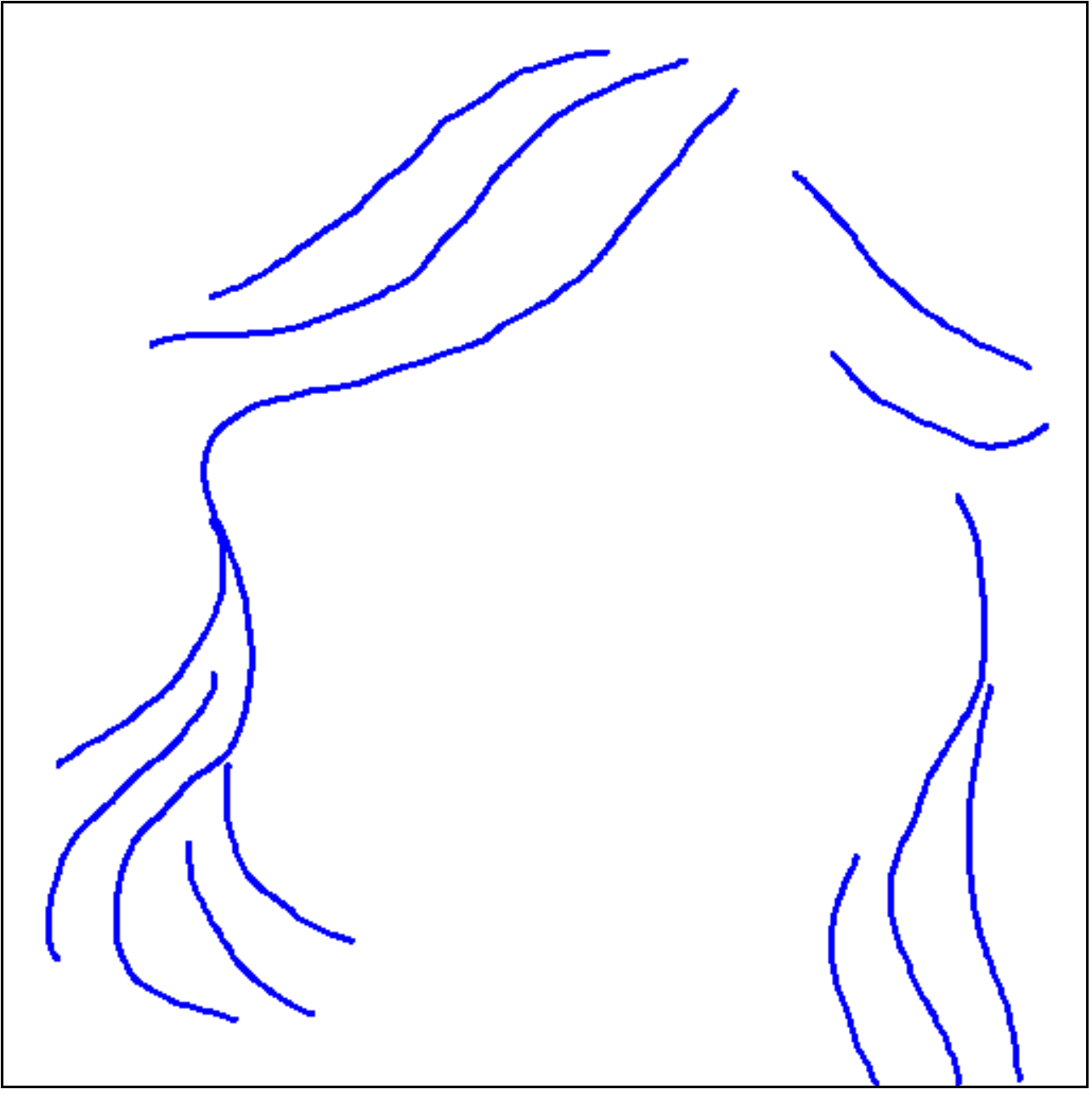}}
            \hspace{-1.57mm}
            \subfloat[]
            {\includegraphics[width=0.249\linewidth]{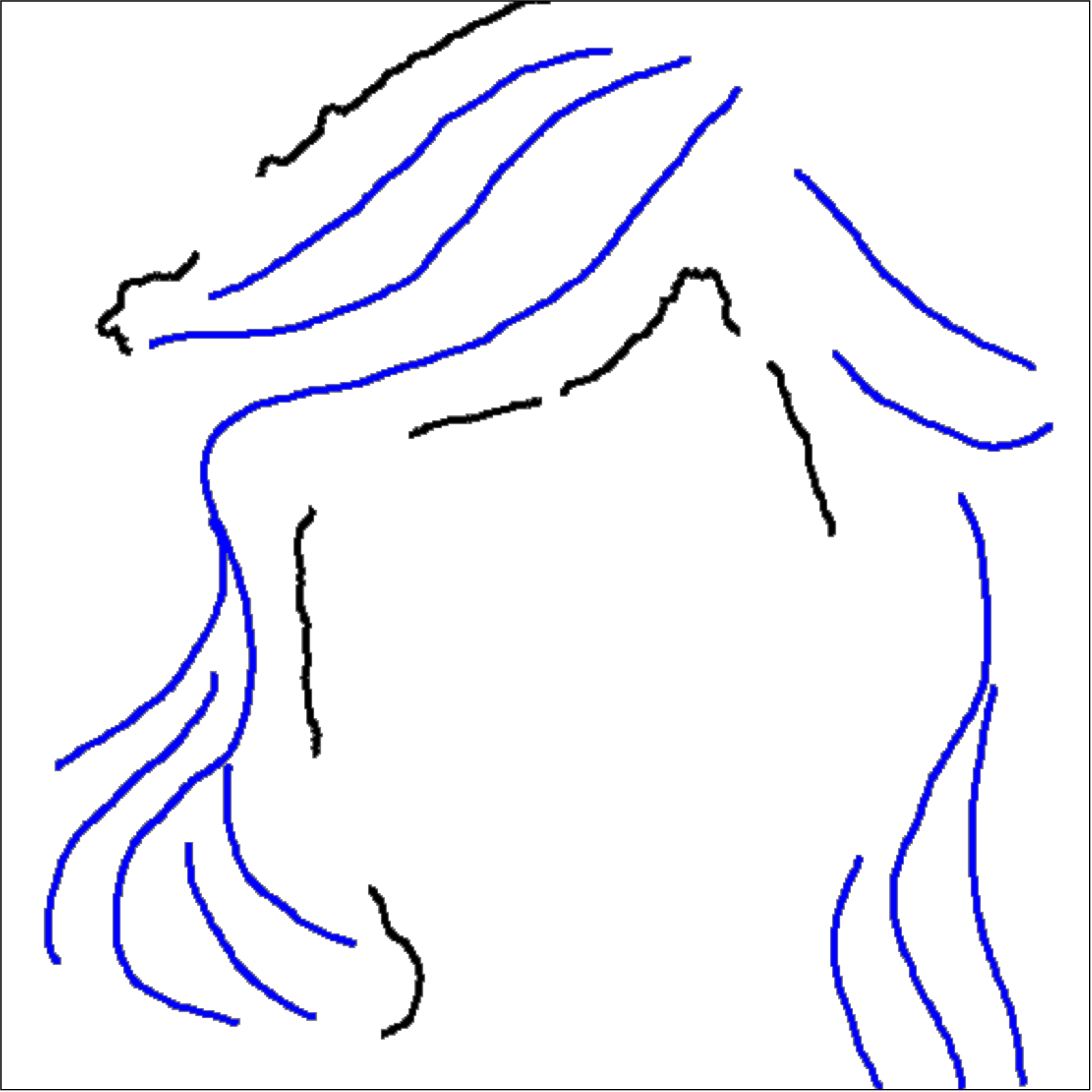}} 
    \end{tabular}

    \caption{Data preparation for training S2M-Net. Given a hair matte (a), non-hair strokes are generated randomly around the contour (b) of the target hair region, and are then appended into a hair sketch (c) as the input (d) fed into S2M-Net.}
    \label{fig:s2m_input}
\end{figure}

\begin{figure*}[t]{
    \includegraphics[width=.99\linewidth]{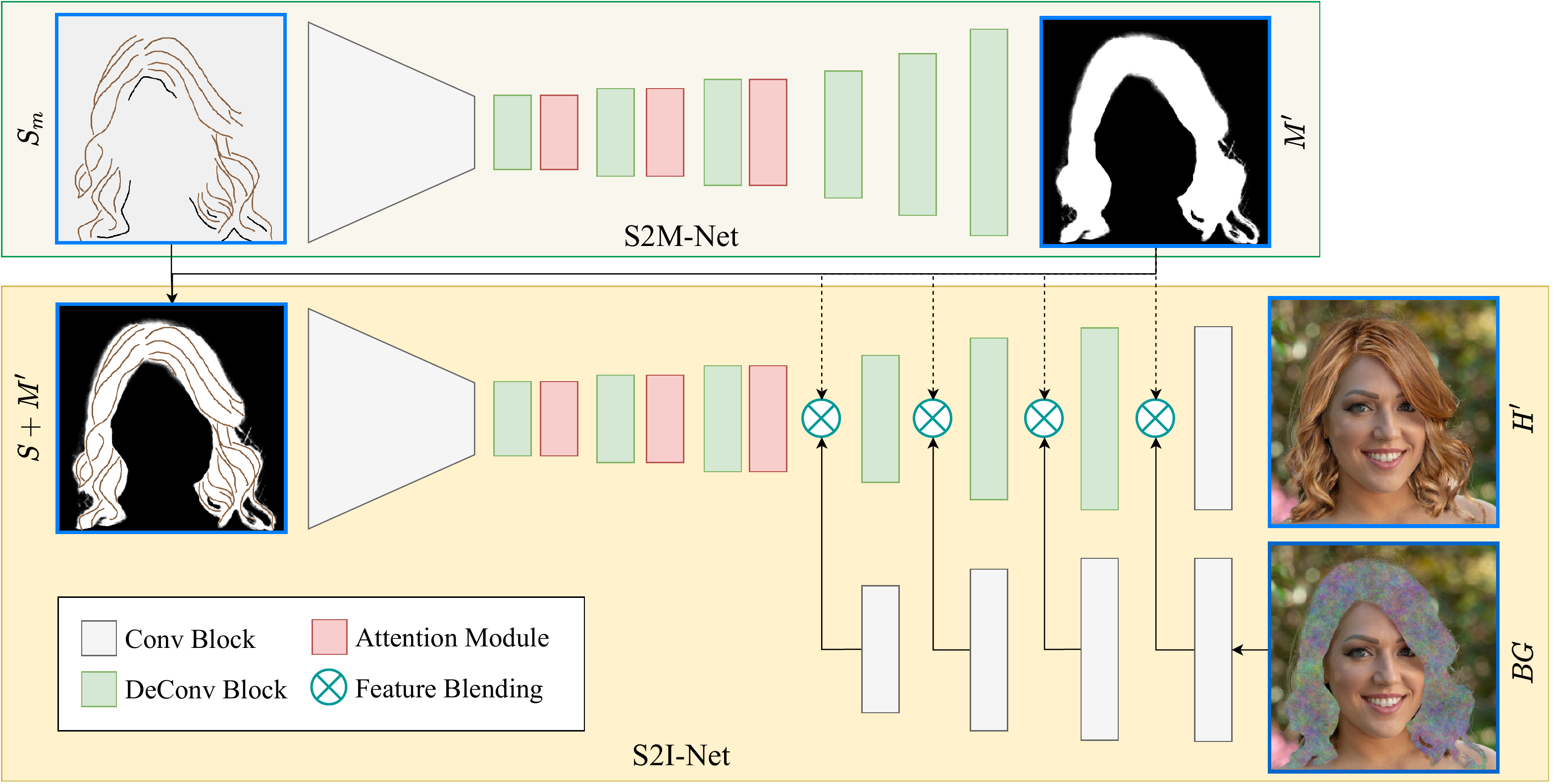}
    \caption{
    {
    %\hbc{Maybe consider adding the notation on top of the image: $S_m$ to $M'$, $S + M'$ to $H'$, 'BG'?}\cfc{Updated.}
    The pipeline of our \sysName}, a two-stage framework for sketch-based hair image synthesis from coarse to fine. Given a hair sketch, the first stage S2M-Net focuses on hair matte generation {(Top)}. Then the generated matte representing a target hair shape is fused with the hair sketch indicating the local structure of a certain hairstyle. Based on that, the second stage S2I-Net manages to synthesize a realistic hair image {(Bottom)}, by {blending the background region  on the feature levels guided by the generated matte}. {The background input (in the bottom-right corner) is derived by replacing the hair region of the original image with Gaussian noise.} Both S2M-Net and S2I-Net apply the self-attention modules to capture the long-range dependencies of the input sketch. Note that {we use} skip connections{, which are omitted in the figure} for simplicity.
    {Original image courtesy of Charcharius.}
    } 
    \label{fig:network}
}
\end{figure*}

\section{Algorithm}
\label{sec:alg}
{Now} we discuss the {algorithm} behind the user interface. The key problem is how to covert {a sparse set of colored strokes} to a photo-realistic hair image. In the following, we first give an overview about our algorithm and then describe the algorithm details.   

\paragraph{Overview}  As illustrated in Figure \ref{fig:network}, our framework consists of two main {{networks}:} the sketch-to-matte network (S2M-Net for short) and the {sketch-to-image} network (S2I-Net for short). The second network also relies on the matte generated by the first network but we call it S2I-Net for simplicity. 

Our first step is to predict a hair matte from an input hair sketch (containing both hair and (optional) non-hair strokes) via the so-called S2M-Net, {which has} an encoder-decoder architecture (Section \ref{sec:S2M}). To make the network focus on the long-term dependency of hair strokes, we apply self-attention modules \cite{fu2019dual} to our network. S2I-Net (Section \ref{sec:S2I}) aims to synthesize a realistic hair image which faithfully respects the structure and appearance depicted by the colored hair strokes and the hair shape depicted by the {generated} matte. 
We adopt a similar network architecture to S2M-Net, but also incorporate the background blending module, since users often need to design hairstyles on top of an existing portrait image and the designed hairstyles should naturally blend to the background region of the original image. 

Below we first introduce our S2M-Net (Section \ref{sec:S2M}) and then S2I-Net (Section \ref{sec:S2I}). Finally, we present the {two} approaches for hair sketch auto-completion (Section \ref{sec:auto-comp}).

\subsection{Hair Sketch-to-Matte Synthesis}
\label{sec:S2M}
S2M-Net takes as input a sketch map {$S_m \in \mathbb{R}^{512 \times 512 \times 1}$}, which contains both hair and non-hair strokes, by setting the colored strokes in our dataset {with one color (e.g., blue) and non-strokes with black color} (visualized like the one in Figure \ref{fig:s2m_input} (d)), and outputs a plausible hair matte $M' \in \mathbb{R}^{512 \times 512 \times 1}$ {(like the one in Figure \ref{fig:s2m_input} (a))} defining the hair shape faithful to a few strokes depicting a desired hair structure. % (like the one in Figure \ref{fig:s2m_input} (a)). 
To achieve this, we adopt an encoder-decoder network with self-attention modules.  
To prepare the dataset for training S2M-Net, we first extract hair contours (Figure \ref{fig:s2m_input} (b)) from ground-truth hair mattes via their distance maps. {The hair contours are slightly pushed away (randomly set from 3 to 8 pixels) from the hair areas}. Then, we derive non-hair strokes by randomly erasing most of the hair contours, in order to balance the density of the non-hair strokes and hair strokes during training.
It means only a small number of non-hair strokes are retained, since we want to set the non-hair input only as an option for users. 
The stroke width is randomly set ranging from 3 to 15 pixels to define the size of a non-hair area and avoid over-fitting. Finally, the non-hair strokes and the hair strokes are fused together in the sketch {map,} %maps 
denoted as $S_m$ (illustrated in Figure \ref{fig:s2m_input} (d)) before fed into S2M-Net.

\textbf{Network Architecture.} Figure \ref{fig:network}  (Top) shows our proposed sketch-to-matte network (S2M-Net). We adopt an encoder-decoder generator with the self-attention modules \cite{fu2019dual}, extended from the image-to-image translation model \cite{pix2pix2017}.
In the first three layers of the decoder, three self-attention modules are repeatedly applied following each deconvolution layer to focus on global and high-level translation. We do not insert any self-attention modules in the latter layers considering the exponential increase of attention computation as the spatial size of feature maps enlarges. In addition, the latter layers of the decoder should get involved to low-level detail generation from the shallow layers of the encoder, but not global features. Please find the details of the network architecture and parameter settings in the supplemental materials.

\textbf{Why Self-attention?} 
Hair sketches only contain a sparse set of colored strokes indicating the local and global appearance and geometry (e.g., occlusion{s} at local T-junction structures and global directions of hair strands). Thus, they {form} long-range dependencies. 
However, convolution layers only perceive their local areas. 
Larger receptive fields to completely cover the long-range strokes would require the network to be deeper, requiring a larger-scale dataset for training many more weights. 
To alleviate the issue, we apply self-attention modules \cite{fu2019dual} to the S2M-Net for extracting long-range correspondence of hair strokes. They introduce few parameters but effectively enhance the representation capacity of the network. We will evaluate the effectiveness of attention modules for our sketch-to-matte task in Section \ref{exp:matte}. The visualization of the attention maps produced by our S2M-Net is shown in the supplemental materials.

\textbf{Loss Function.} Since the input $S_m$ and output $M'$ 
%\hbc{$M'$? It seems $M$ used in Equation 1 is not introduced?}\cfc{Fixed and added a description for $M$.}
are of lower frequency compared with real images, we train S2M-Net by simply using $L_1$ loss and the adversarial loss of patch-GAN \cite{pix2pix2017}, performing as a condition-GAN framework. The total loss $L_M(M',M)$ is defined as:
\begin{equation}
    L_{S2M}(S_m,M',M)={\lambda_1}L_1(M',M)+{\lambda_2}L_{adv}(S_m,M',M),
\end{equation}
where {$M$ is the ground-truth matte}, $\lambda_1$ and $\lambda_2$ are respectively the weights of the $L_1$ loss and adversarial loss $L_{adv}$ ($\lambda_1=100$ and $\lambda_2=1$ in our implementation). To avoid over-fitting, the non-hair strokes in $S_m$ are randomly generated at each iteration to provide more paired data.   

\subsection{Hair Sketch-to-Image Synthesis}
\label{sec:S2I}
After S2M-Net, we get a synthesized hair matte $M'$ explicitly indicating a target hair shape. As illustrated in Figure \ref{fig:network} (Bottom), our hair sketch-to-image network (S2I-Net) is designed to predict a realistic hair image $H' \in \mathbb{R}^{512 \times 512 \times 3}$ based on the hair sketch $S \in \mathbb{R}^{512 \times 512 \times 3}$ and the hair matte $M'$. 
Inspired by Scribbler \cite{sangkloy2017scribbler}, we adopt color{ed} sketches to represent both hair structure and appearance. {In contrast, $S_m$ in Section \ref{sec:S2M} has a single channel only.}

\textbf{Color-coded Hair Stroke Inputs.}
Before training S2I-Net, we pre-process the input sketch $S$ to be more expressive and appropriate for geometry learning. Thanks to our manually annotated hair dataset, we have hair sketches with separately annotated strokes, which often have slightly different colors even for neighboring strokes.
In this way, such sketches can highlight the special structures of diverse hairstyles, e.g., long and coherent wisps, occluding areas, {braided knots}, etc. To encourage the network to learn the correspondence from the colored strokes to the hair image appearance and augment hair sketch-image pairs, at each training iteration, each stroke is assigned the color of a pixel randomly picked from the corresponding stroke region of the paired hair image (Figure \ref{fig:network}). %\hbc{maybe refer to Fig. 3?}\cfc{The colors in Fig.3 are randomized while those in Fig.5 are from the image.}).
During testing, users can decide each stroke color to control the local hair appearance.

\textbf{Network Architecture.}
As shown in Figure \ref{fig:network} (Bottom), S2I-Net is similar to the S2M-Net. The key difference is its incorporation of a background blending module. Like the setting of MichiGAN \cite{tan2020michigan}, we do not input the background region together with the sketch to guarantee the most capacity of the network for learning hair sketch-to-image translation. Thus, the background region $BG$ is blended with the synthesized hair region on the feature {level} guided by the hair matte $M'$, denoted as
\begin{equation}
    F_i=F_i^h\cdot M_i+F_i^{BG}\cdot(1-M_i),
\end{equation}
where $F_i$ is the blended feature of the $(i+1)$th last layer of the main branch between the hair feature $F_i^h$ and the background feature $F_i^{BG}$, while $M_i$ is the nearest-downsampled matte from {$M$} to fit the feature size. The background input is derived by replacing the hair region %\hbc{here the hair region is masked by $M'$ or is the original hair region?}\cfc{During training, it is the original hair matte, while during testing it is the generated matte. I changed $M'$ to $M$ above.} 
of the original image with Gaussian noise. At the main branch of S2I-Net, we only blend the background region at the last four layers. Please find the details of the network architecture and parameter settings in the supplemental materials.

\begin{figure}
\centering
    % \setlength{\fboxrule}{0.1pt}
    % \setlength{\fboxsep}{-0.01cm}
    % \hspace{-2mm}
    \begin{tabular}{c}
    \vspace{-1mm}
    \hspace{-2mm}
    {\includegraphics[width=0.33\linewidth]{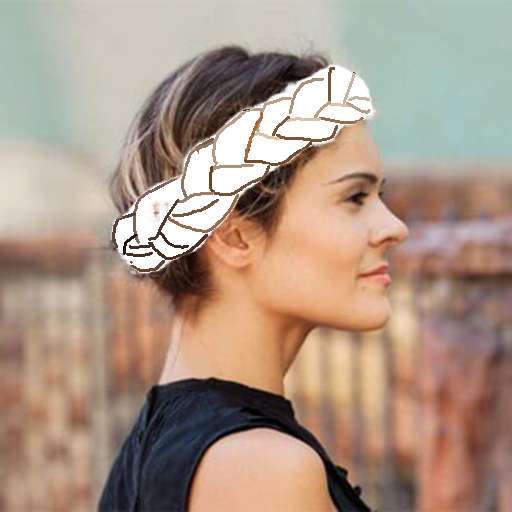}} 
    \hspace{-1.6mm}
    {\includegraphics[width=0.33\linewidth]{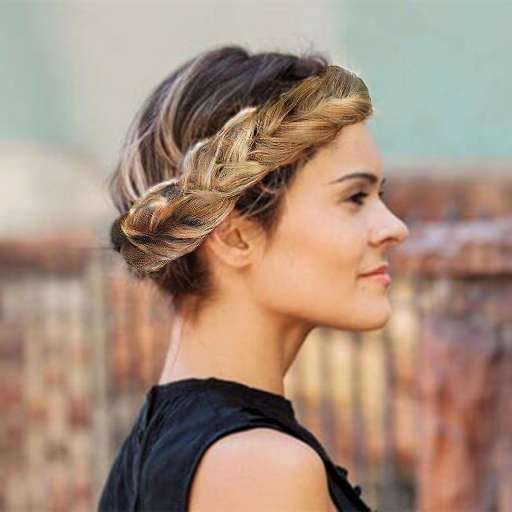}} 
    \hspace{-1.6mm}
    {\includegraphics[width=0.33\linewidth]{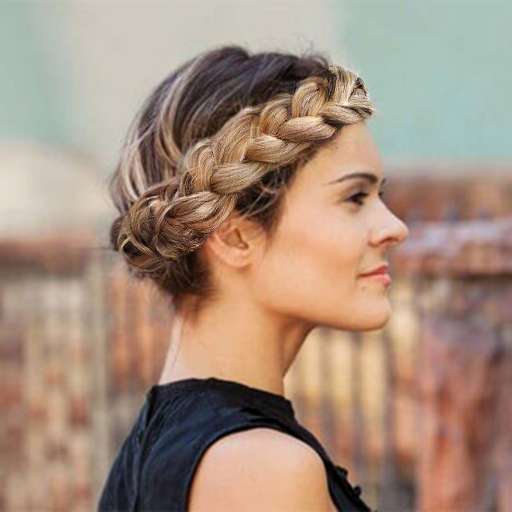}} \hspace{1mm}
    \\
    \hspace{-3.5mm}
    {\includegraphics[width=0.33\linewidth]{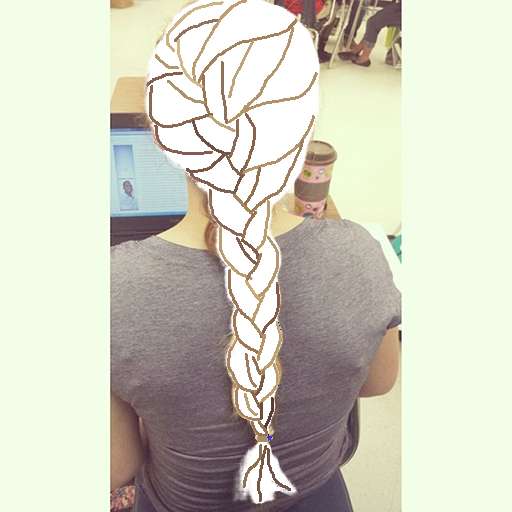}}
    \hspace{-1.6mm}
    {\includegraphics[width=0.33\linewidth]{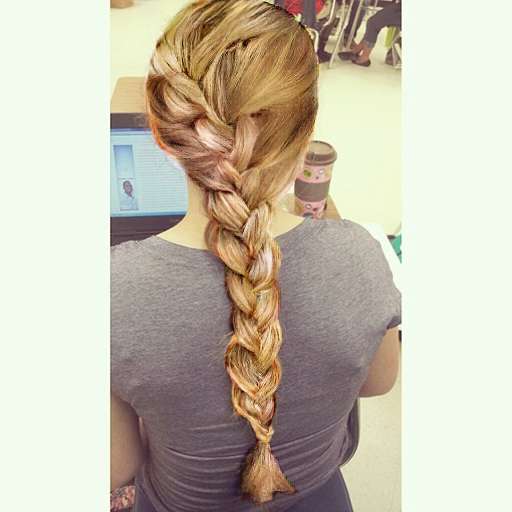}} 
    \hspace{-1.6mm}
    {\includegraphics[width=0.33\linewidth]{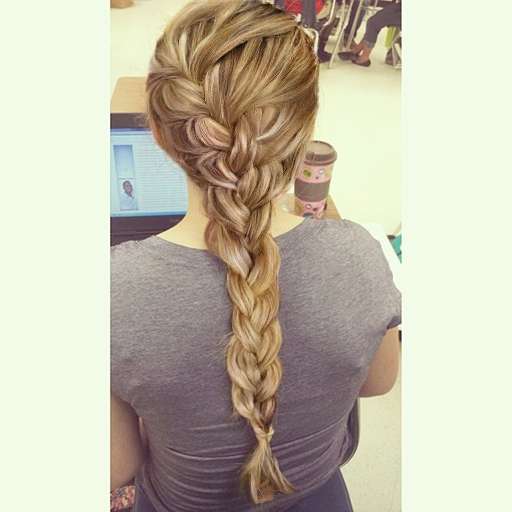}}\\
    \vspace{4mm}
    (a) Input Sketch \qquad (b) Joint Training \quad (c) Separate Training 
    \end{tabular}
    \vspace{-4mm}
    \caption{Comparison of joint training and separate training. (a) is the image combining the background input and {the} sketch input for S2I-Net for visualization. (b) and (c) are the testing braid %\hbc{braided hairstyle} 
    results respective{ly} from the model joint{ly} trained on {the} braided and unbraided datasets and the ones separately trained on the two type of datasets. Please zoom in to better examine their quality. {Original images courtesy of Parekh Cards and Wicker Paradise.}}
    \label{fig_tog_vs_sep}
\end{figure}

% \begin{figure}
% \centering
%     \setlength{\fboxrule}{0.1pt}
%     \setlength{\fboxsep}{-0.01cm}
%     % \hspace{-2mm}
%     \begin{tabular}{ccc}
%     \vspace{-1mm}
%             \hspace{-2mm} 
%             % \framebox
%             {\includegraphics[width=0.33\linewidth]{fig/results_v2/shape_loss/sk/braid_2830}} 
%             & 
%             \hspace{-4mm}
%             % \framebox
%             {\includegraphics[width=0.33\linewidth]{fig/results_v2/shape_loss/wo/braid_2830_combine_wo.pdf}} 
%             & 
%             \hspace{-4mm}
%             % \framebox
%             {\includegraphics[width=0.33\linewidth]{fig/results_v2/shape_loss/w/braid_2830_combine_w.pdf}}\\
            
%             \hspace{-2mm} 
%             {\includegraphics[width=0.33\linewidth]{fig/results_v2/shape_loss/sk/braid_3026}} 
%             & 
%             \hspace{-4mm}
%             % \framebox
%             {\includegraphics[width=0.33\linewidth]{fig/results_v2/shape_loss/wo/braid_3026_combine_wo.pdf}} 
%             & 
%             \hspace{-4mm}
%             % \framebox
%             {\includegraphics[width=0.33\linewidth]{fig/results_v2/shape_loss/w/braid_3026_combine_w.pdf}}
%             \\
%             \vspace{4mm}
%     (a) Input Sketch & (b) w/o ShapeLoss &  (c) w ShapeLoss
%     \end{tabular}
%     \vspace{-4mm}
%     \caption{{Comparison of the results generated by the model with and without {the} shape loss. It is obvious that {the} shape loss can {greatly} help S2I-Net to learn {a} clear shape for the braided part.}}
%     \label{fig_shapeloss}
% \end{figure}
\begin{figure}
\centering
    \setlength{\fboxrule}{0.1pt}
    \setlength{\fboxsep}{-0.01cm}
    % \hspace{-2mm}
    \begin{tabular}{c}
    \vspace{-1mm}
            \hspace{-2mm} 
            % \framebox
            {\includegraphics[width=0.33\linewidth]{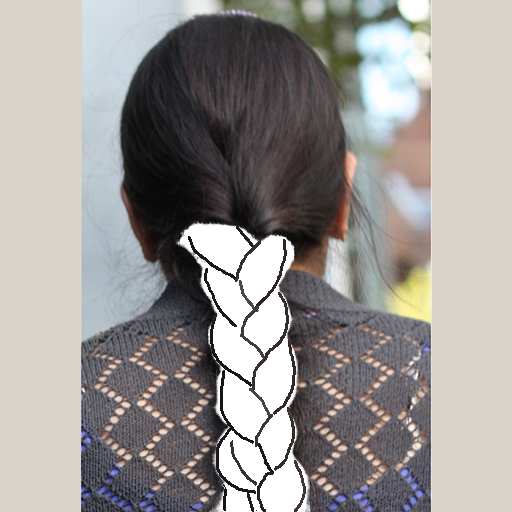}} 
            \hspace{-2mm}
            % \framebox
            {\includegraphics[width=0.33\linewidth]{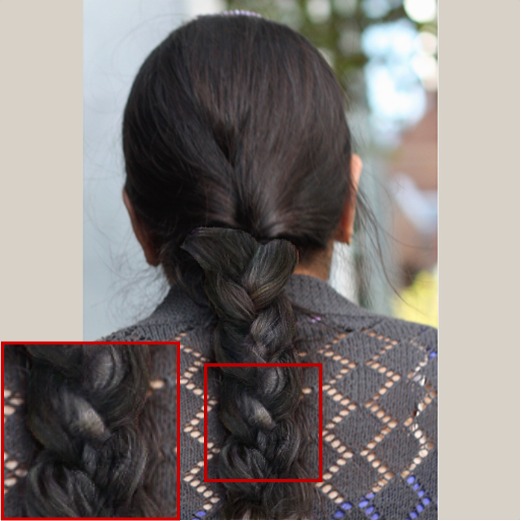}}
            \hspace{-2mm}
            % \framebox
            {\includegraphics[width=0.33\linewidth]{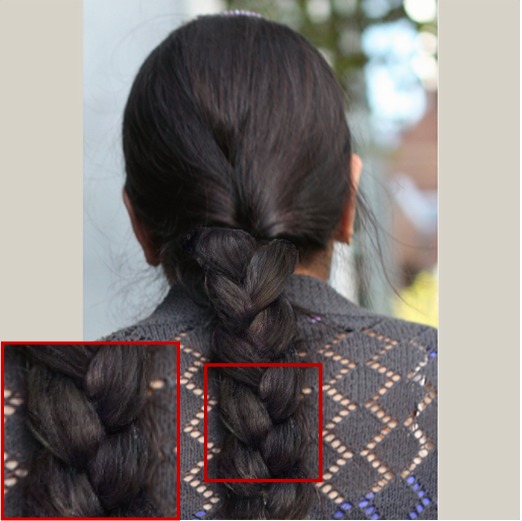}}\\
            \hspace{-2mm} 
            {\includegraphics[width=0.33\linewidth]{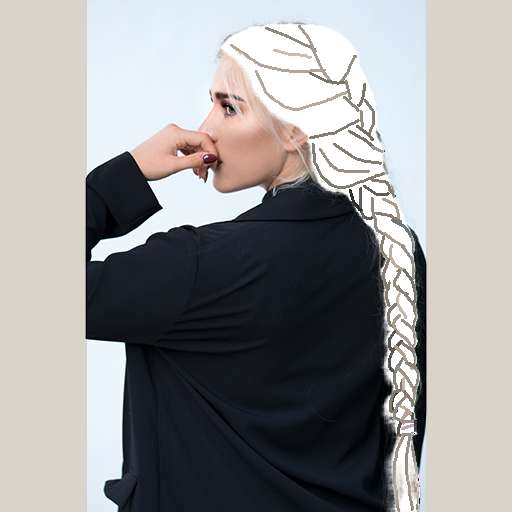}} 
            \hspace{-2mm}
            % \framebox
            {\includegraphics[width=0.33\linewidth]{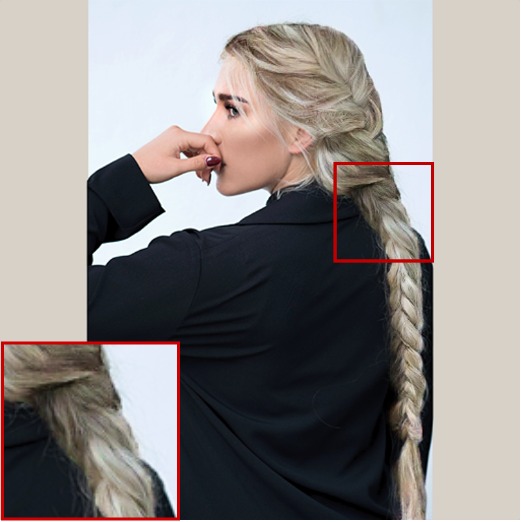}}
            \hspace{-2mm}
            % \framebox
            {\includegraphics[width=0.33\linewidth]{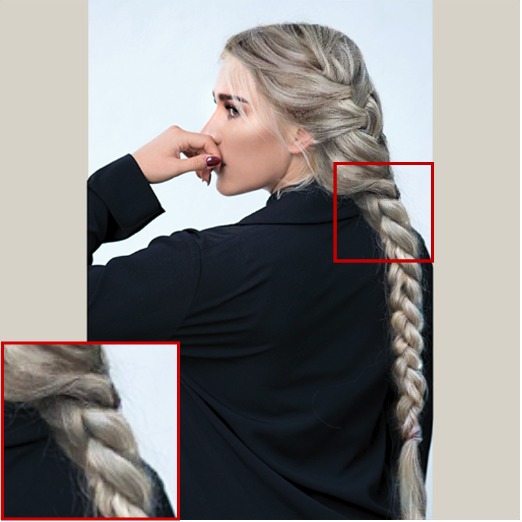}}
            \\
            \vspace{4mm}
    (a) Input Sketch \quad (b) w/o Shape Loss \quad (c) w\ Shape Loss \quad
    \end{tabular}
    \vspace{-4mm}
    \caption{Comparison of the results generated by the model with and without {the} shape loss. It is obvious that {the} shape loss can {greatly} help S2I-Net to learn {a} clear shape for the braided part. {Original images courtesy of Robyn Jay and Pxfuel.}}
    \label{fig_shapeloss}
\end{figure}

\textbf{Loss Function.}
To train our S2I-Net, we adopt a combination of several types of losses for this complicated problem. For the generator, we use {an} $L_1$ loss to guarantee the pixel-wise quality of the generated image $H'$. Beside that, a patch-GAN is adopted to provide an adversarial loss
\cite{pix2pix2017} 
%\hbc{ref?}\cfc{Shall we cite the vanilla GAN paper? This adversarial loss of the patch-GAN is set by pix2pix \cite{pix2pix2017}.} 
to improve the naturalness and local details, denoted as $L_{cGAN}$. We find that the perceptual loss $L_{per}$ \cite{dosovitskiy2016generating} can effectively enhance hair texture synthesis, and thus also incorporate it for training the generator.

Since a braided sketch and a non-braided sketch are actually two different representation types: the former with structure lines while the latter with flow lines. Training them together easily leads to more ambiguity for S2I-Net, as shown in Figure \ref{fig_tog_vs_sep}. Thus, we train two S2I-Nets separately for the braided and unbraided hairstyles. For braided hairstyles, the network needs to learn a general shape from the braided sketch. However, only $L_1$ loss would make S2I-Net resort to spreading the color texture around the sketch, but not forming the clear knot shape, as shown in  Figure \ref{fig_shapeloss} (b). To address this issue, we introduce a shape reconstruction loss for braided hairstyle generation. The shape loss first smooths {both} the result $H'$ and ground truth $H$ via a Gaussian filter and then compute{s} the $L_1$ loss of their hair regions, denoted as
\begin{equation}
    L_{shape}(H',H)=L_1(M' \cdot g(H'),M' \cdot g(H)),
\end{equation}
where g($\cdot$) is a Gaussian filter with the kernel size $10$ and $\sigma =10$. With the shape loss, S2I-Net is able to restore a clearer braided shape, as shown in Figure \ref{fig_shapeloss} (c). We also tried to apply the shape loss for generating unbraided hairstyles, but found that the quality is not significantly improved compared to that without the shape loss. For simplicity, we keep the shape loss for both braided and unbraided hairstyles. 

In summary, our final loss function combining all of the above losses can be written as
\begin{equation}
    \begin{aligned}
    L_{S2I}({S},H',H)=&{\lambda_1}L_1(H',H)+{\lambda_2}L_{cGAN}({S},H',H)\\
                      &+{\lambda_3}L_{per}(H',H)+{{\lambda_4}L_{shape}(H',H)},
    \end{aligned}
\end{equation}
where we {empirically} set the loss weights $\lambda_1$, $\lambda_3$, $\lambda_4$=100 and $\lambda_2$ = 1 in our implementation. % {empirically}.

\begin{figure}
\centering
    \setlength{\fboxrule}{0.1pt}
    \setlength{\fboxsep}{-0.01cm}
    \hspace{-2mm}
    \begin{tabular}{ccccc}
    \vspace{-1mm}
            % \framebox
            {\includegraphics[width=.95\linewidth]{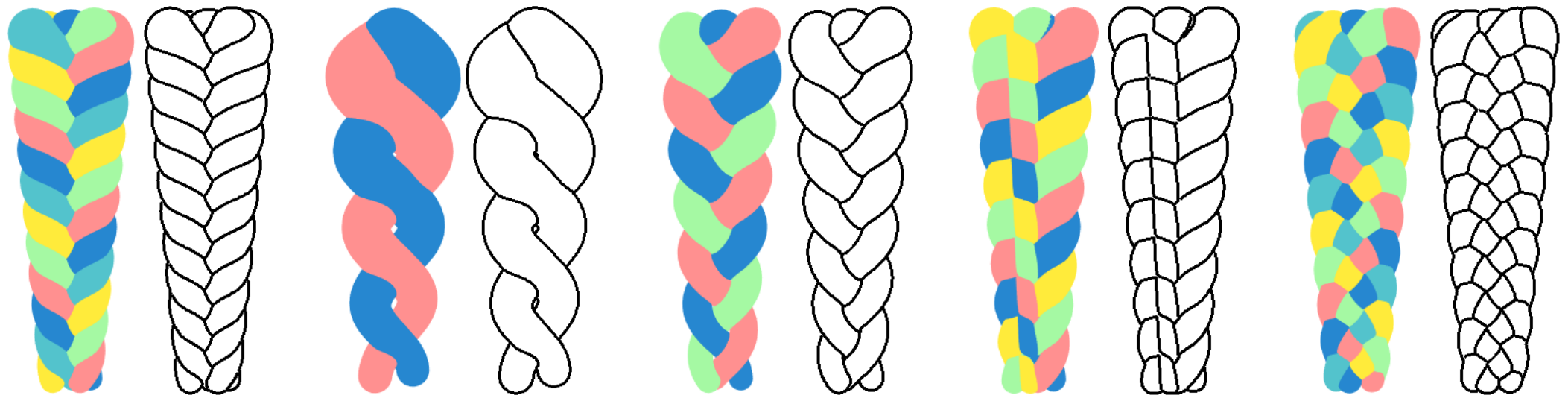}} 
    \end{tabular}
    \caption{{Five braid templates and their corresponding edge maps. The hairstyles from left to right are fish-tail, rope, three-strand, four-strand, and five-strand.}}
    \label{fig:braid_models}
\end{figure}

\begin{figure}[htb]{
    \centering
    \vspace{-4mm}
    \subfloat[User-guided Braid Model]{
        \includegraphics[width=.44\linewidth]{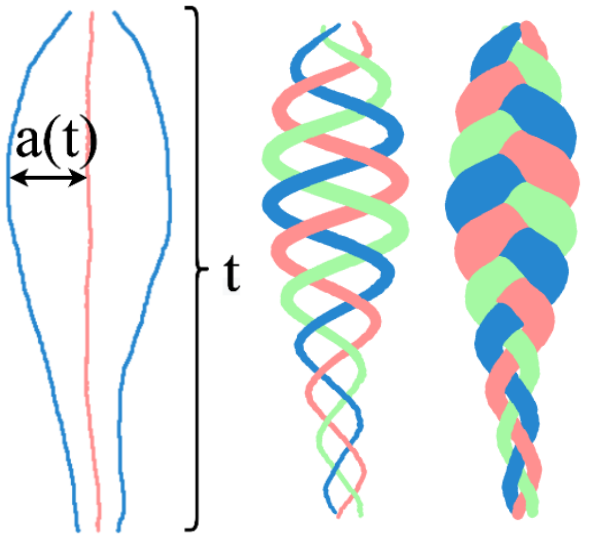}
    }
    \hspace{4mm}
    \subfloat[Various $w$ Settings]{
        \includegraphics[width=.392\linewidth]{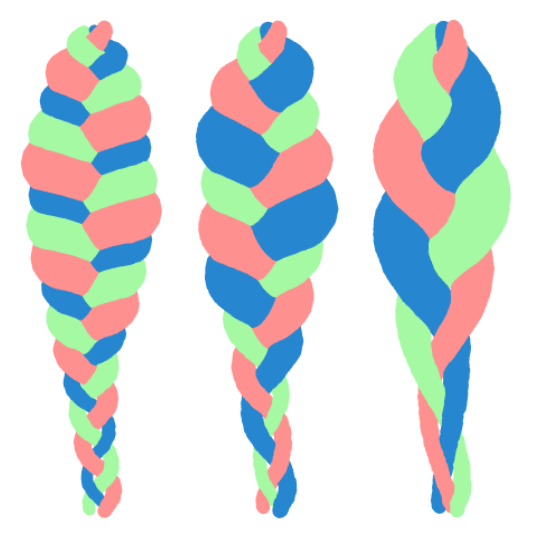}
    }
    \caption{{(a) The generated three-strand braid model (Right) is expanded from the three center-lines (Middle) guided by {the} {user-specified} %users' 
    rough boundary lines (the blue strokes at the left). (b) $w$ is respectively set as $1.5, 1$, and $0.5$ to  change the knot number and direction of the 3-strand model ($w=-1$) in (a).}}
    \label{fig:braid_comp_pipeline}
    }
\end{figure}

% \begin{figure}[htb]{
%     \centering
%     \subfloat[Color Palette]{
%         \includegraphics[width=.34\linewidth]{fig/results_v2/braid_models/sk/color_palette_0.png}
%     }\hspace{-4mm}
%     \subfloat[Color Sketch]{
%         \includegraphics[width=.34\linewidth]{fig/results_v2/braid_models/sk/color_edge_0.png}
%     }\hspace{-4mm}
%     \subfloat[Generated Image]{
%         \includegraphics[width=.34\linewidth]{fig/results_v2/braid_models/res/braid_16_0_rgb=99_74_62.png}
%     }\\
%     \subfloat[]{
%         \includegraphics[width=.34\linewidth]{fig/results_v2/braid_models/res/braid_16_2_rgb=99_86_131.png}
%     }\hspace{-4mm}
%     \subfloat[]{
%         \includegraphics[width=.34\linewidth]{fig/results_v2/braid_models/res/braid_16_3_rgb=34_25_31.png}
%     }\hspace{-4mm}
%     \subfloat[]{
%         \includegraphics[width=.34\linewidth]{fig/results_v2/braid_models/res/braid_16_6_rgb=161_121_76.png}
%     }
%     \caption{\cf{To be updated.}}
%     \label{fig:braid_comp_pipeline2}
%     }
% \end{figure}

\begin{figure}[htb]{
    \centering
    \vspace{-4mm}
    \subfloat[Colorization Pipeline]{
        \includegraphics[width=.44\linewidth]{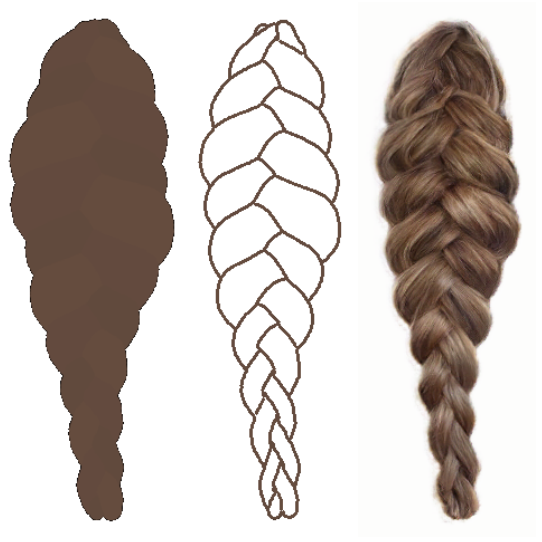}
    }
    \hspace{4mm}
    \subfloat[Diverse Appearance]{
        \includegraphics[width=.44\linewidth]{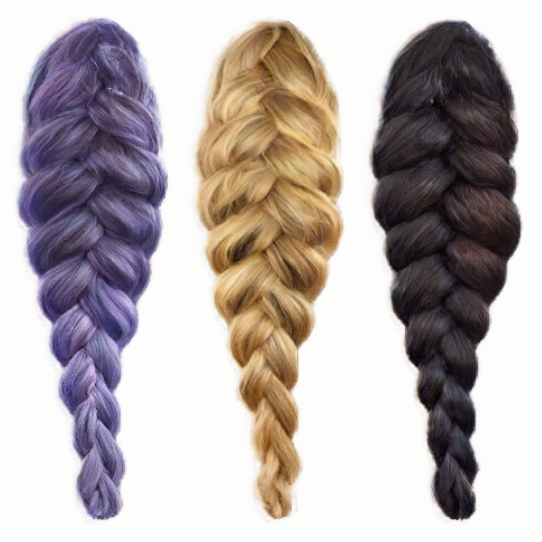}
    }
    \caption{{(a) A braided sketch (Middle) is colorized based on a generated color palette {(Left)}, and fed into S2I-net to produce a braided image (Right). (b) Re-colorizing the braided sketch based on this manner can vary the hair images with diverse appearance.}}
    \label{fig:braid_comp_pipeline2}
    }
\end{figure}

\subsection{{Sketch Auto-Completion}} 
\label{sec:auto-comp}
We observe that most of hairstyles have simple but large regions with similar local structures that need to be filled with repetitive hair strokes to reduce the ambiguity for our S2I-Net. For example, a braided hairstyle typically has many repeated knots, while an unbraided hairstyle has the property that hair stokes often share local orientations with their neighbors. When designing {a} hairstyle, it is tedious to require users to draw {a complete set of hair strokes}. % the complete hair sketches.
To reduce the user's load, we propose two sketch auto-completion methods given sparse strokes for the braided and unbraided hairstyles respectively.

\textbf{For Braided Hairstyles}. 
Some existing auto-completion techniques \cite{tu2020continuous,hsu2020autocomplete} perform well for auto-completing repetitive patterns given a small pattern or exemplar elements. However, for braided sketches, since most of users are even not good at creating such exemplar elements, the above auto-completion methods might not be very effective here. Instead, we adopt a procedural approach.

Inspired by Hu et al. \shortcite{hu2014capturing}, we first construct parametric braided 3D models (Figure \ref{fig:braid_models}), based on the braid theory \cite{braidTheory1947}. Then we introduce a simple method for users to control a desired braided shape via two roughly drawn strokes (the blue strokes in Figure \ref{fig:braid_comp_pipeline} (a) Left), which indicate the position and boundary of a target braided sketch.
Based on the two boundary strokes, we get $a(t)$ width, range $t$, and the translating parameters $\triangle X$, $\triangle Y$ for determining the final shape of a braided model with a specific structure. 
Here, we take {a} three-strand braid as an example. The middle one in Figure \ref{fig:braid_comp_pipeline} (a) illustrates three intertwining center-lines before expanded to tubes until inter-tube penetrations occur, forming {a} 3-strand braided model{ (Figure \ref{fig:braid_comp_pipeline} (a) Right), guided by the two user-specified boundary strokes in (Figure \ref{fig:braid_comp_pipeline} (a) Left)}. The reshaped three-strand braided model can be described as
\begin{equation}
    \left\{\begin{array}{l}L_0:x=a(t)\sin(wt)+\triangle x,\;y=\triangle y,z=b\sin(2wt)\\L_1:x=a(t)\sin(wt+2\pi/3)+\triangle x,y=\triangle y,z=b\sin(2(wt+2\pi/3))\\L_2:x=a(t)\sin(wt+4\pi/3)+\triangle x,y=\triangle y,z=b\sin(2(wt+4\pi/3))\end{array},\right.
\end{equation}
and
\begin{equation}
    \left\{\begin{array}{l}\triangle Y=(B_{Y0}+B_{Y1})/2\\\triangle X=(B_{X0}+B_{X1})/2\\a(t)=\left|B_{X0}-B_{X1}\right|/2\\t=\lbrack0,\left|\triangle Y\right|\rbrack\end{array},\right.
\end{equation}
where $\triangle x \in \triangle X$, $\triangle y \in \triangle Y$, $B_0$ and $B_1$ are the two boundaries, while $w$ can control the number of braided knots and the knot direction, as illustrated in Figure \ref{fig:braid_comp_pipeline} (a) Right and (b). A larger value of $w$ leads to more knots, and the negative values make the knots point up while the positive ones make them point down. After constructing the reshaped 3D braided models for braids, we extract the braid edge as the braided sketch using the Canny algorithm \cite{canny1986computational}. The colored braided sketch (Figure \ref{fig:braid_comp_pipeline2} {(a) Middle}) can be derived by setting edge colors upon the color palette (Figure \ref{fig:braid_comp_pipeline2} (a) {Left}), which is generated by assigning the 3D tubes with the user-desired colors. Conditioned on the auto-completed sketch, {photo-}realistic braided images can be synthesized via our S2M-Net and S2I-Net. Figure \ref{fig:braid_comp_pipeline2} (a) Right and (b) show the generated images with different appearance by changing the sketch colors. In our current implementation, we provide users with five different braided hairstyles (Figure \ref{fig:braid_models}), including fishtail, rope, three-strand, four-strand, five-strand. The definitions for the 3D lines of them are provided in the supplemental materials.

\textbf{For Unbraided Hairstyles}. 
Our main idea is that the input strokes can be diffused to the rest of the hair region since users prefer to achieve similar structures near the input strokes. Thus, we duplicate some strokes close to the input strokes to encourage the generated hair image to be faithful to the input sketch.

\begin{figure}[htb]{
    \centering
    \hspace{-2mm}
    \subfloat[]{
        \includegraphics[width=.246\linewidth]{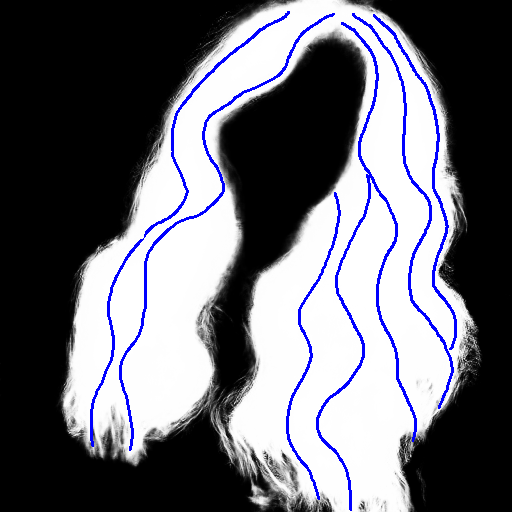}
    } \hspace{-2.2mm}
    \subfloat[]{
        \includegraphics[width=.246\linewidth]{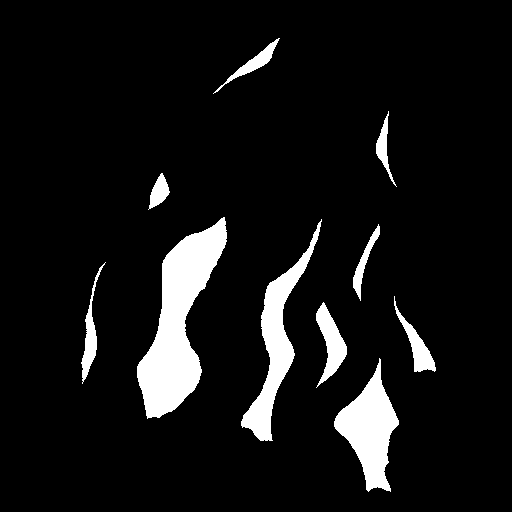}
    }\hspace{-2.2mm}
    \subfloat[]{
        \includegraphics[width=.246\linewidth]{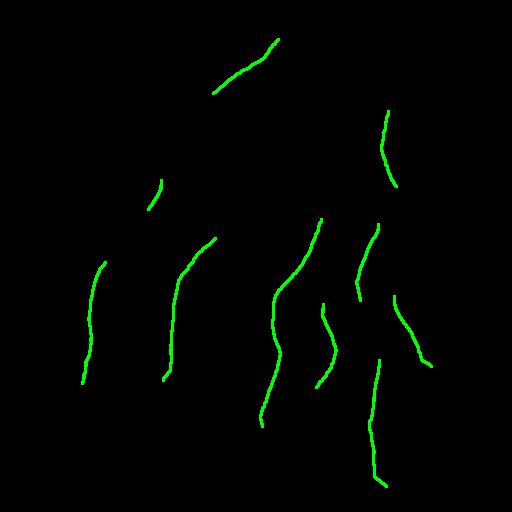}
    }\hspace{-2.2mm}
    \subfloat[]{
        \includegraphics[width=.246\linewidth]{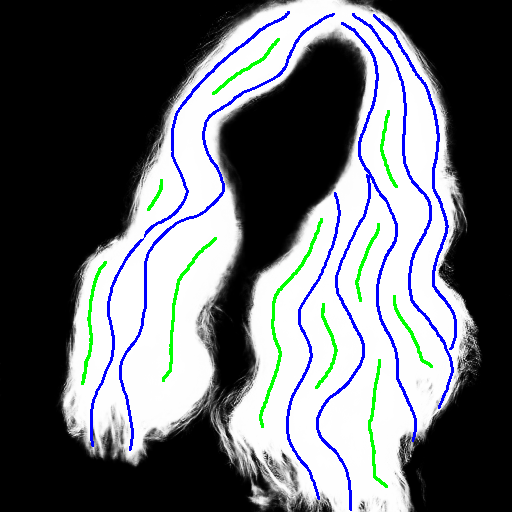}
    }
    \caption{{The pipeline of our method for {unbraided} hair sketch auto-completion. Given the input sketch (a), the medial-axis extraction algorithm \cite{10.1006/cgip.1994.1042} extracts the extra strokes (c) from the subtracted map {(b)}. (d) is the completed sketch, of which the blue {and green strokes are the user-specified and automatically generated strokes, respectively}. %  original inputs from users, while the green strokes are generated automatically.
    }
    }
    \label{fig:refine_pipeline}
    }
\end{figure}

% \begin{figure}[htb]{
%     \centering
%     \subfloat[Input Strokes]{
%         \includegraphics[width=.315\linewidth]{fig/results_final/sk_refine/pipeline/incompleted_sketch}
%     }
%     \subfloat[Subtracted Mask]{
%         \includegraphics[width=.315\linewidth]{fig/results_final/sk_refine/pipeline/subtracted_hair_matte}
%     }
%     \subfloat[Contour Map]{
%         \includegraphics[width=.315\linewidth]{fig/results_final/sk_refine/pipeline/contour_map}
%     }\\ \vspace{-2mm}
%     \subfloat[Filtered Map]{
%         \includegraphics[width=.315\linewidth]{fig/results_final/sk_refine/pipeline/filtered_contour_map}
%     }
%     \subfloat[{Added Strokes}]{
%         \includegraphics[width=.315\linewidth]{fig/results_final/sk_refine/pipeline/addtion_stokes}
%     }
%     \subfloat[Completed Sketch]{
%         \includegraphics[width=.315\linewidth]{fig/results_final/sk_refine/pipeline/completed_sketch}
%     }
%     \caption{The pipeline of our method for hair sketch {auto-}completion. The blue strokes are the original inputs from users, while the green strokes are generated automatically.}
%     \label{fig:refine_pipeline}
%     }
% \end{figure}

\begin{figure}[htb]{
    \centering
    \vspace{-2mm}
    \subfloat[Original]{
        \begin{minipage}{.34\linewidth}
            \includegraphics[width=\linewidth]{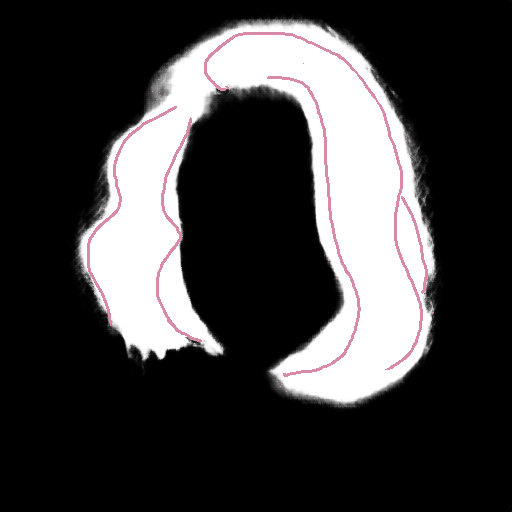}
            \includegraphics[width=\linewidth]{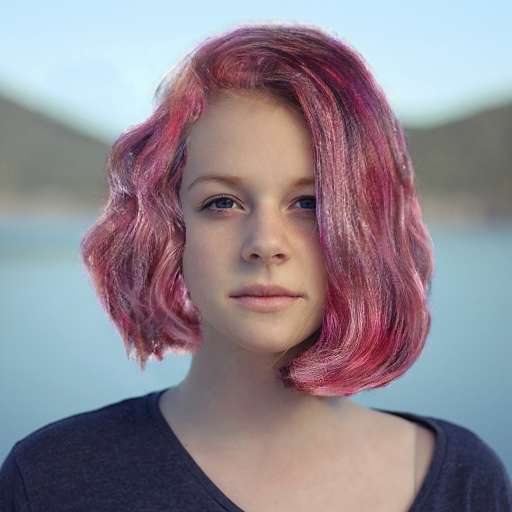}
        \end{minipage}
    }
    \hspace{-4mm}%{-2.2mm}
    \subfloat[Auto-tracing]{
        \begin{minipage}{.34\linewidth}
            \includegraphics[width=\linewidth]{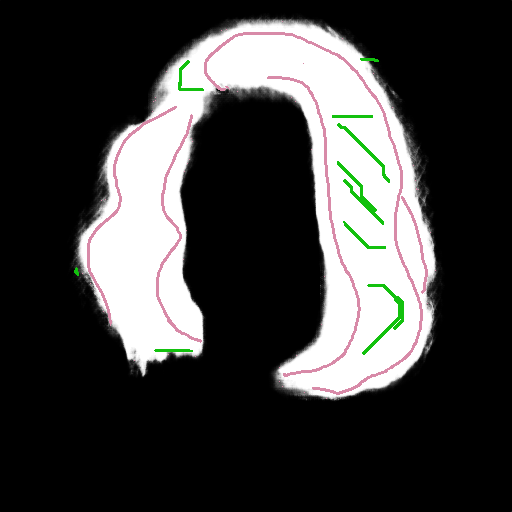}
            \includegraphics[width=\linewidth]{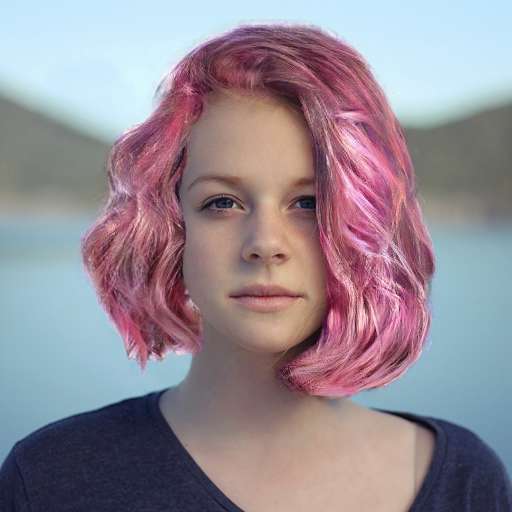}
        \end{minipage}
    }
    \hspace{-4mm}
    \subfloat[Ours]{
        \begin{minipage}{.34\linewidth}
            \includegraphics[width=\linewidth]{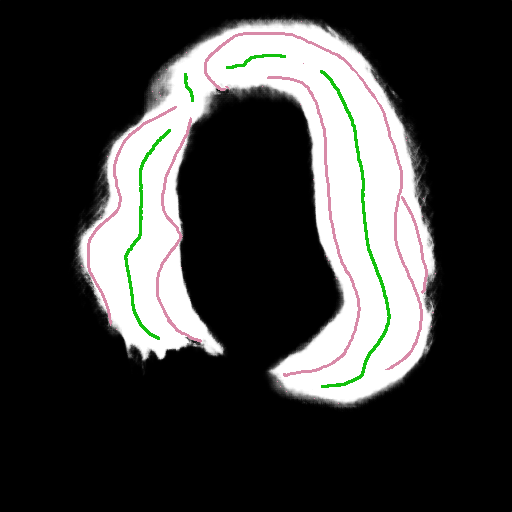}
            \includegraphics[width=\linewidth]{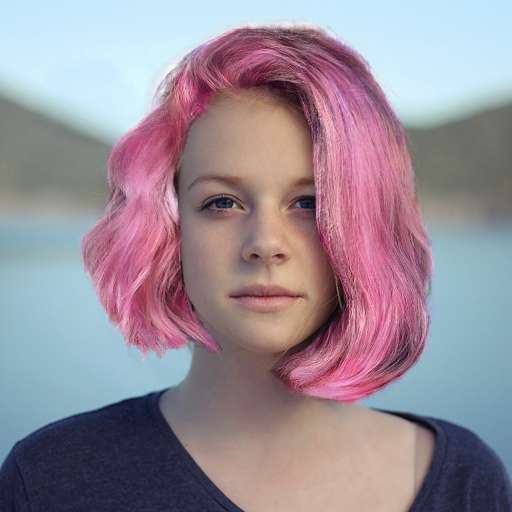}
        \end{minipage}
    }
    \caption{{Comparison of our method and the auto-tracing method \cite{olszewski2020intuitive} for unbraided sketch auto-completion. The top row is the group of the original sketch input and the two generated sketches. 
    % The strokes are highlighted in different colors, but not the actual colors for S2I-Net. The blue strokes are user-specified, while the green ones are auto-generated based on the corresponding methods.
    The original strokes and the generated strokes are all assigned with the same color (pink) for S2I-Net, while the latter are highlighted in green in this example. 
    The bottom row shows the hair images generated by our S2I-Net. {Original image courtesy of Free-Photos.}}
    }
    \label{fig:sk_refine}
    }
\end{figure}

To achieve this, we first apply a dilation operation with $15\times 15$ kernel size to the input strokes (Figure \ref{fig:refine_pipeline} (a)) and subtract the binary mask (trimmed from the matte) from the dilated regions, resulting in a subtracted map (Figure \ref{fig:refine_pipeline} (b)). This map contains {possibly many separate} %many separating 
regions for extracting extra strokes. Given the candidate regions, we use a medial-axis extraction algorithm \cite{10.1006/cgip.1994.1042} to automatically fill strokes (Figure \ref{fig:refine_pipeline} (c)). Finally, the auto-completed sketch (Figure \ref{fig:refine_pipeline} (d)) is derived by blending the input sketch (Figure \ref{fig:refine_pipeline} (a)) with the {automatically filled} strokes (Figure \ref{fig:refine_pipeline} (c)). In this way, the supplemented strokes provide the additional clues for our network to better synthesize details in the regions which are not well covered by the user-specified strokes. 

Figure \ref{fig:sk_refine} (c) shows {an example with}  the auto-completed sketch and the generated hair image. Most of the added strokes follow the orientation of the user-specified strokes. We also compare our results with the auto-tracing based method \cite{olszewski2020intuitive}, in which a synthetic sketch is generated based on an automatically predicted orientation map from the given sketch (similar to \cite{tan2020michigan}). As shown in Figure \ref{fig:sk_refine} (b) and (c), our auto-completion method is able to better keep the structures depicted by the user-specified strokes than this alternative approach. Note that in this example all the strokes (including original and generated) are assigned with the same color. We find that the sparse input (Figure \ref{fig:sk_refine} (a)) would cause appearance ambiguity for S2I-Net in large hair regions, while the additional strokes (Figure \ref{fig:sk_refine} (c)) help produce more consistent appearance with the input color.

\section{Experiment}

\begin{figure*}[htb]{
    \centering
    \hspace{-3mm}
    \subfloat[Input Sketch \& GT]{
        \begin{minipage}{.155\linewidth}
            \framebox{\includegraphics[width=.94\linewidth]{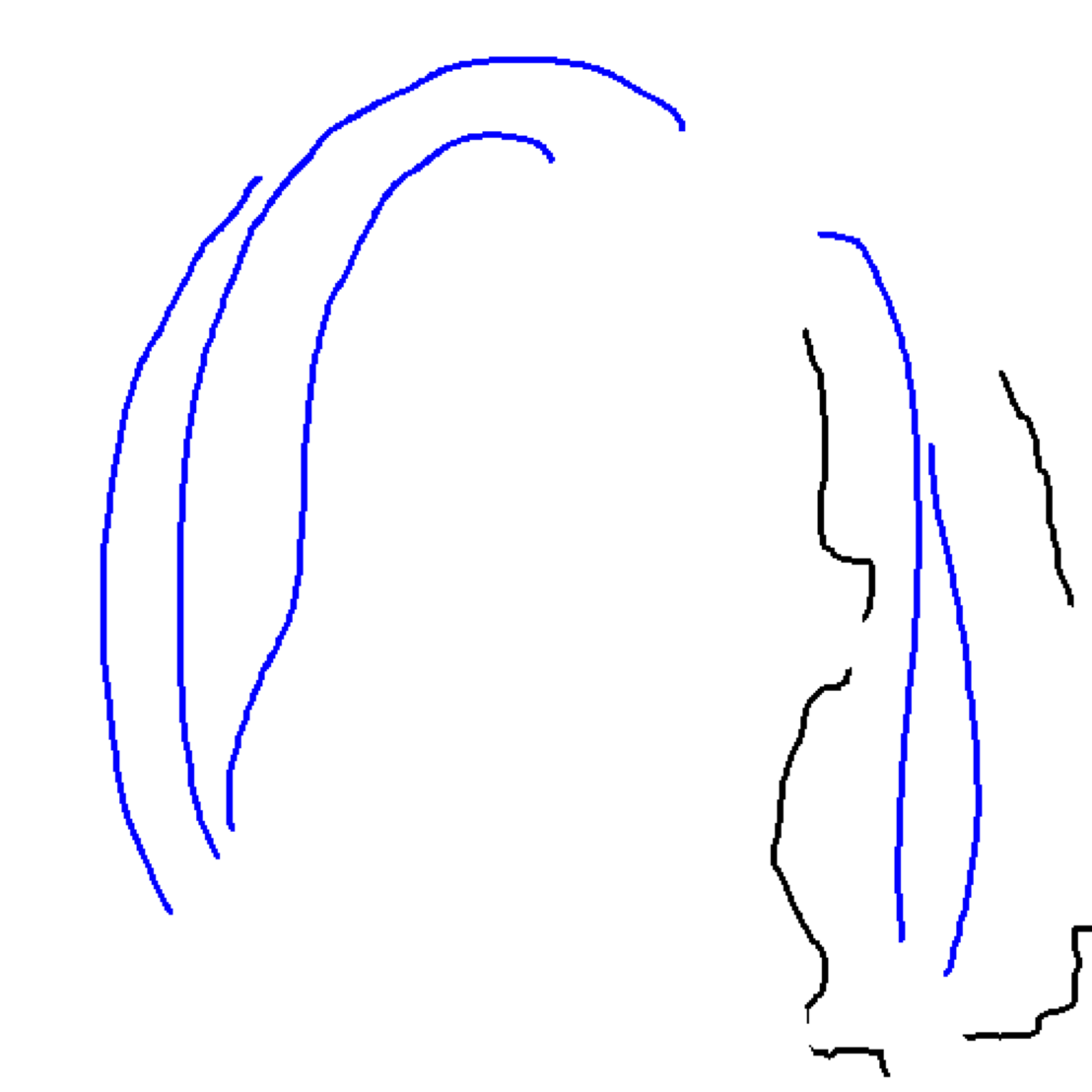}}
            \includegraphics[width=1.03\linewidth]{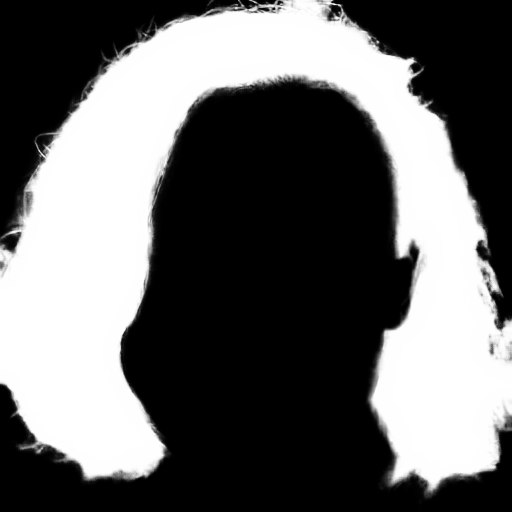}
        \end{minipage}
    }
    \hspace{-2.1mm}
    \subfloat[Generated Mattes]{
        \begin{minipage}{.155\linewidth}
            \centering
            \includegraphics[width=1.027\linewidth]{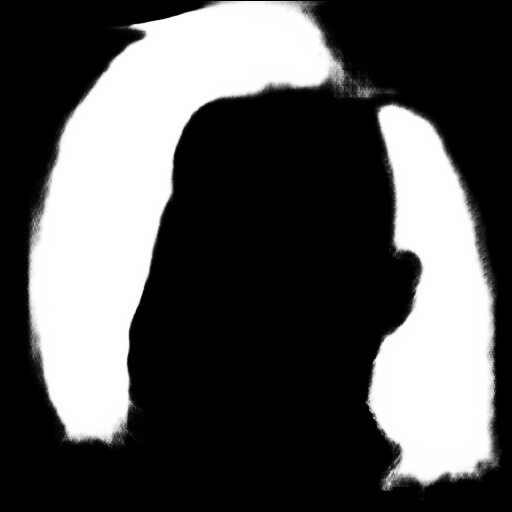}\\
            \includegraphics[width=1.027\linewidth]{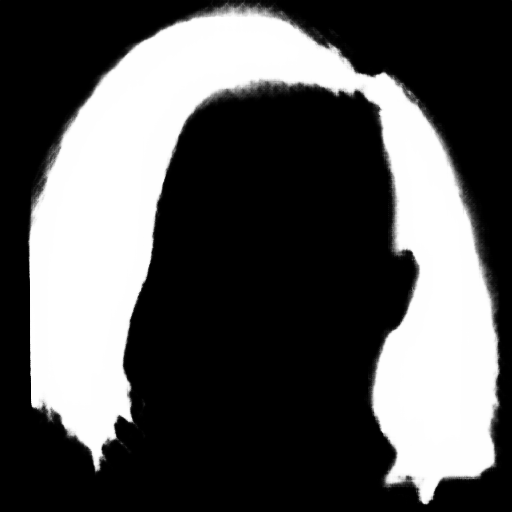}
        \end{minipage}
    }
    \hspace{-2.12mm}
     \subfloat[Difference Maps]{
        \begin{minipage}{.155\linewidth}
            \centering
            \framebox{\includegraphics[width=.94\linewidth]{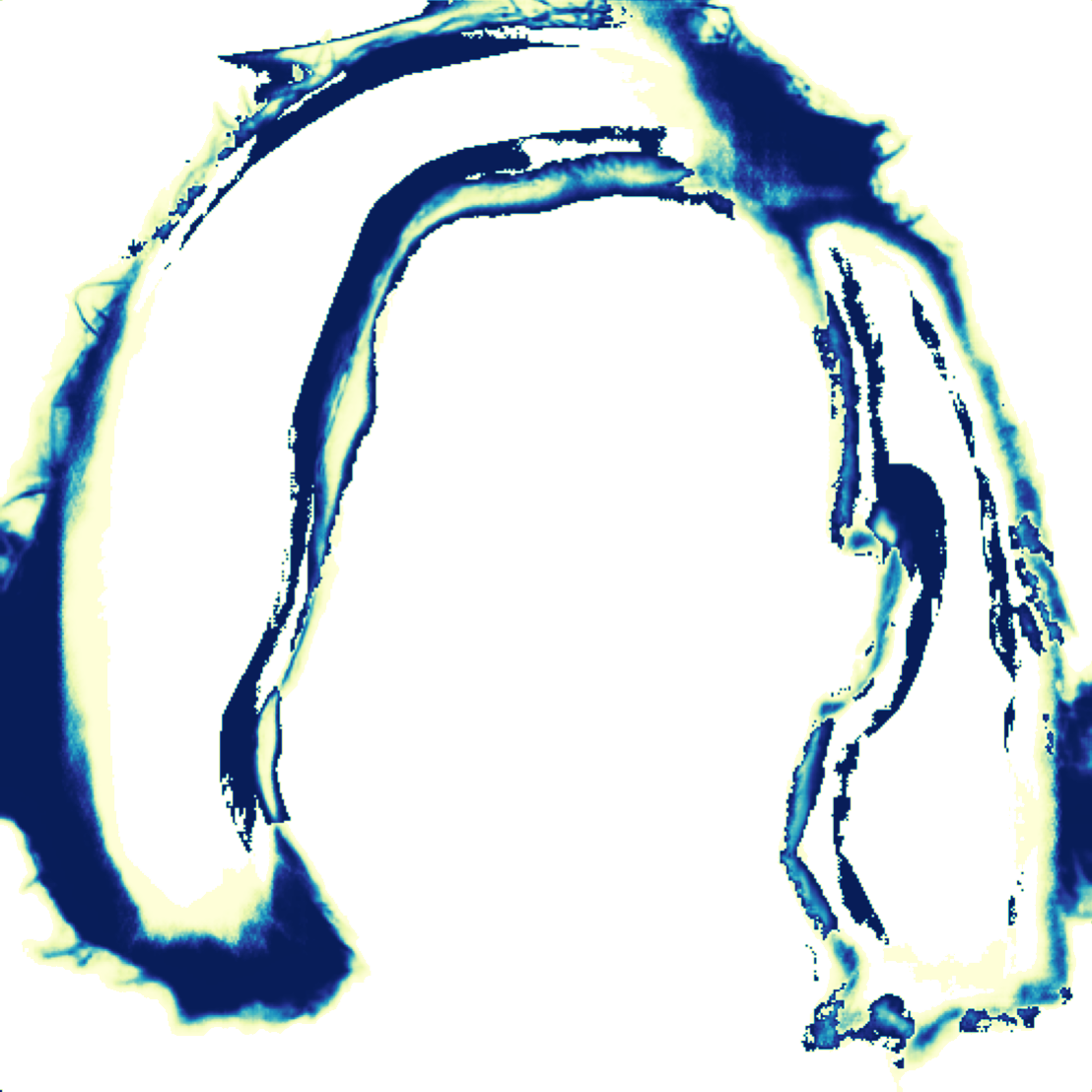}}\\
            \framebox{\includegraphics[width=.94\linewidth]{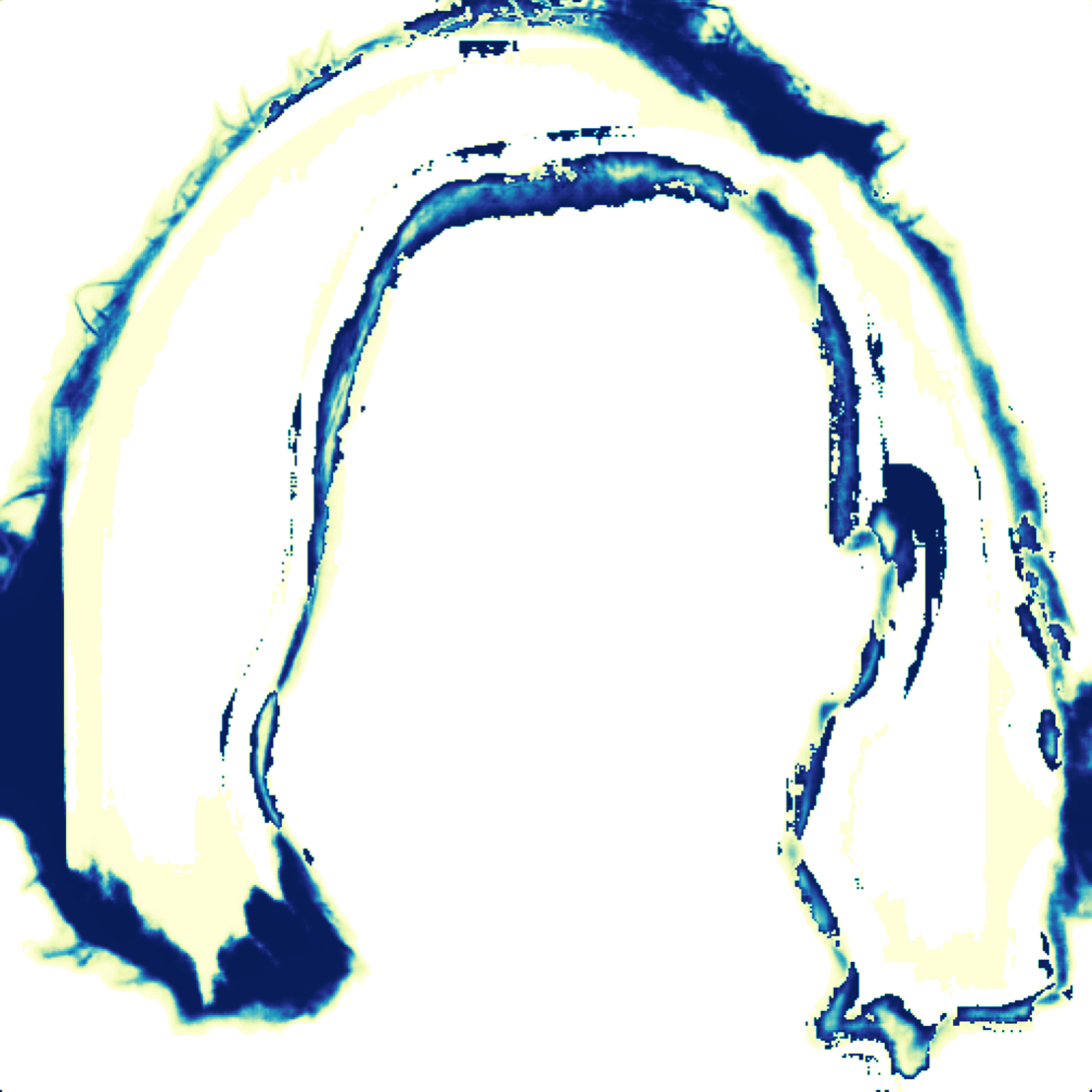}}
        \end{minipage}
    }
    \hspace{2mm}
    \subfloat[Input Sketch \& GT]{
        \begin{minipage}{.155\linewidth}
            \framebox{\includegraphics[width=.94\linewidth]{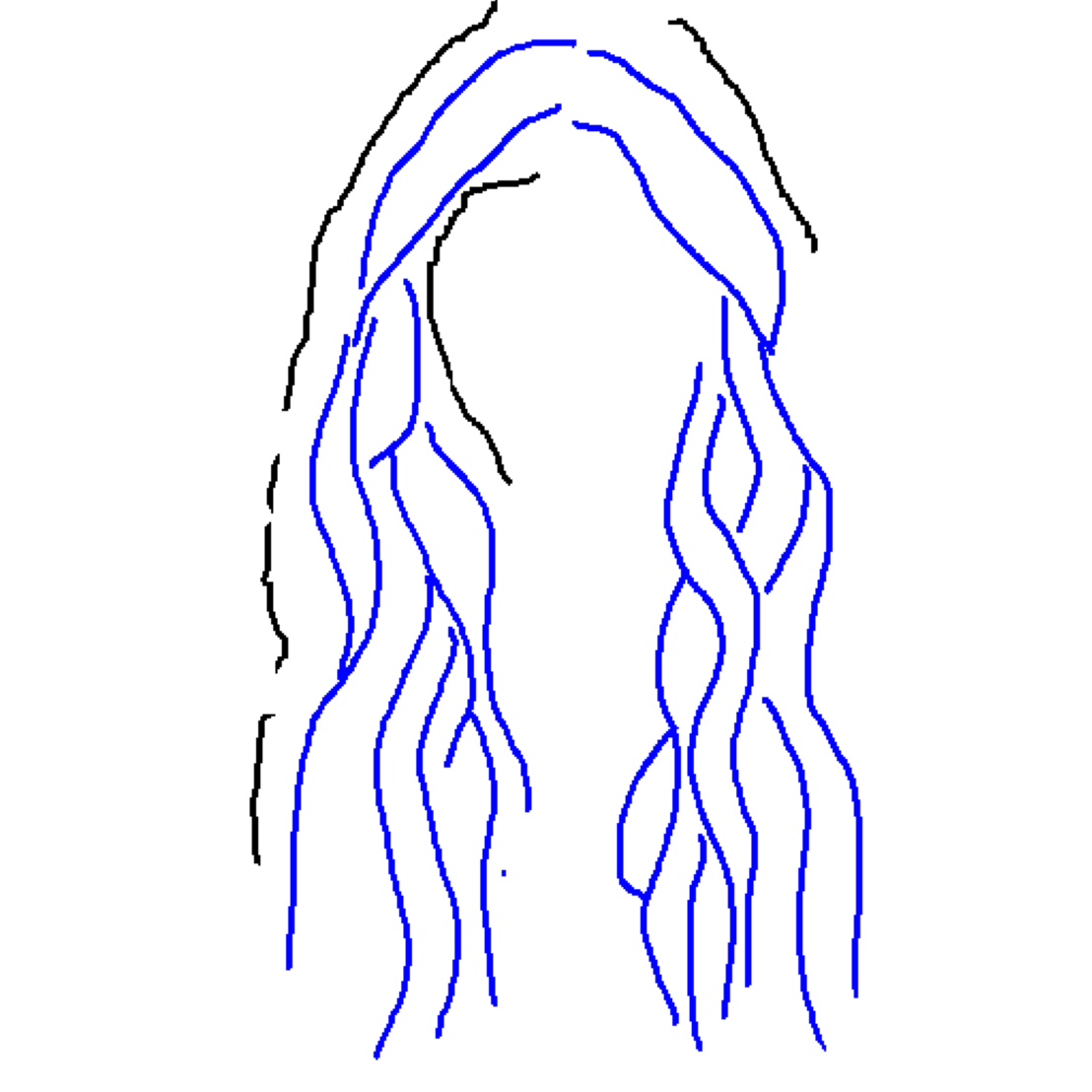}}
            \includegraphics[width=1.03\linewidth]{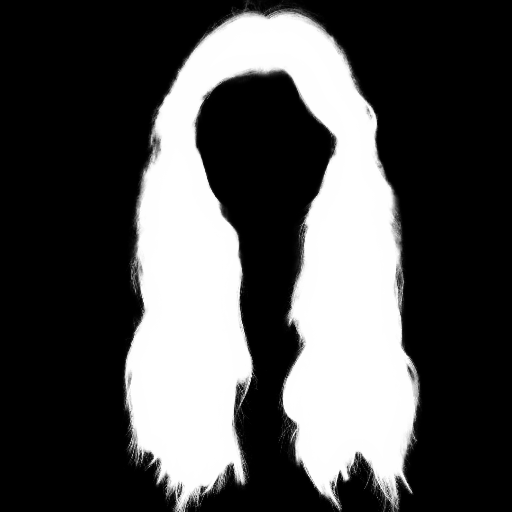}
        \end{minipage}
    }
    \vspace{-2mm}
    \hspace{-2.5mm}
    \subfloat[Generated Mattes]{
        \begin{minipage}{.155\linewidth}
            \centering
          \includegraphics[width=1.027\linewidth]{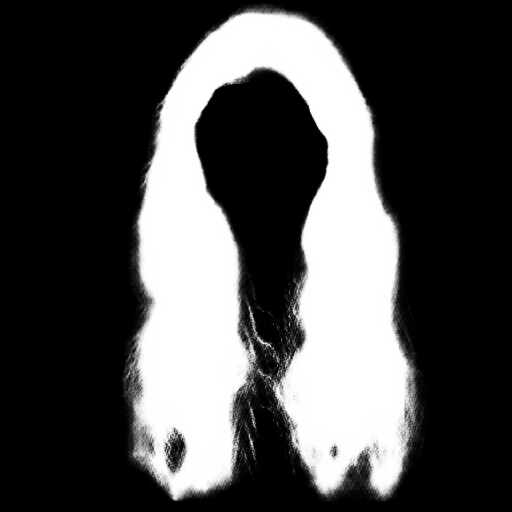}\\
          \includegraphics[width=1.027\linewidth]{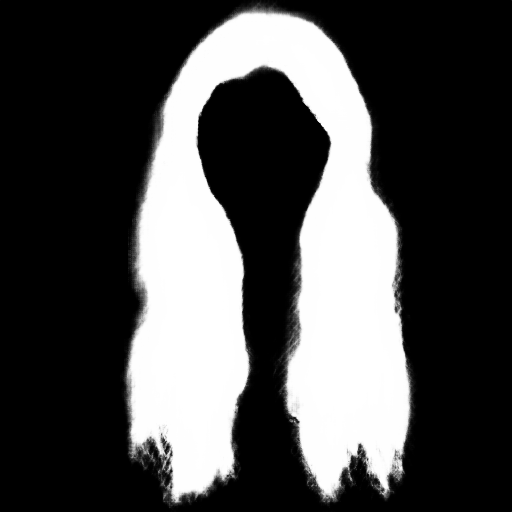}
              \end{minipage}
    }
    \hspace{-2.12mm}
    \subfloat[Difference Maps]{
        \begin{minipage}{.155\linewidth}
            \centering
            \framebox{\includegraphics[width=.94\linewidth]{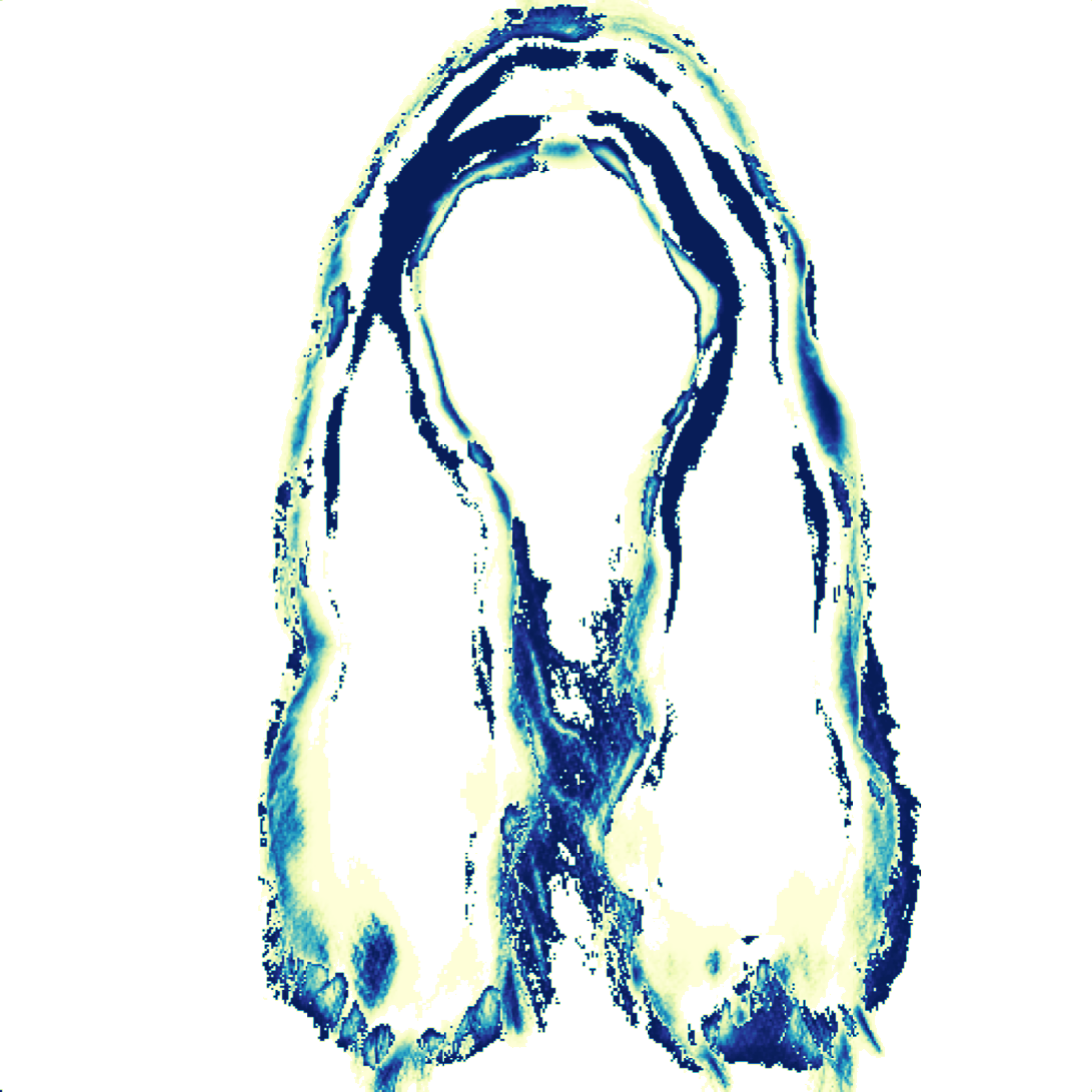}}\\
            \framebox{\includegraphics[width=.94\linewidth]{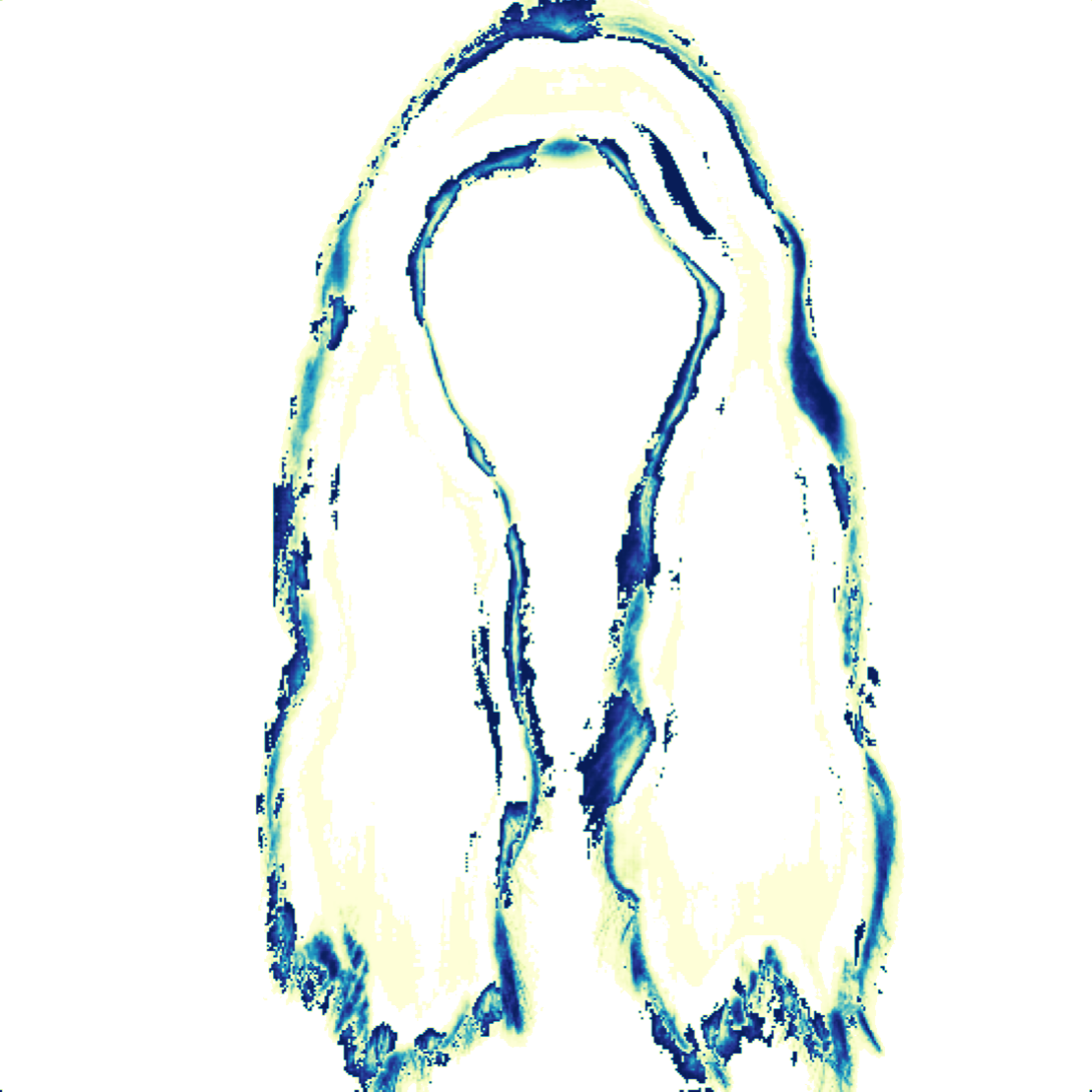}}
        \end{minipage}
    }
    \caption{{Comparison of the mattes generated by the model without and with the self-attention modules. Except for (a) {and (d)}, the top row corresponds {to} the model without the self-attention modules, while the bottom row corresponds {to} the model with those. At each group of left and right, (b) \& (e) are the generated mattes given the sketches ((a) \& (d) Top), while (c) \& (f) are the difference map{s} between the mattes and the ground truth ((a) \& (d) Bottom). In the difference maps, larger blue regions mean higher values of difference from the ground truth.}}
    \label{fig:matte_comparison}
    }
\end{figure*}

\begin{figure*}
    \centering
     \begin{minipage}{.03\linewidth}
        \flushleft{
            \rotatebox{90}{\quad \large{Ground Truth \qquad \qquad Ours \quad \qquad \qquad MichiGAN \qquad \qquad \quad HIS \qquad \qquad \qquad pix2pix \quad \qquad Generated Matte \qquad Input Sketch \quad}}
        }
     \end{minipage}
     \hspace{-2.5mm}
    \begin{minipage}{.975\linewidth}
        \includegraphics[width=\linewidth]{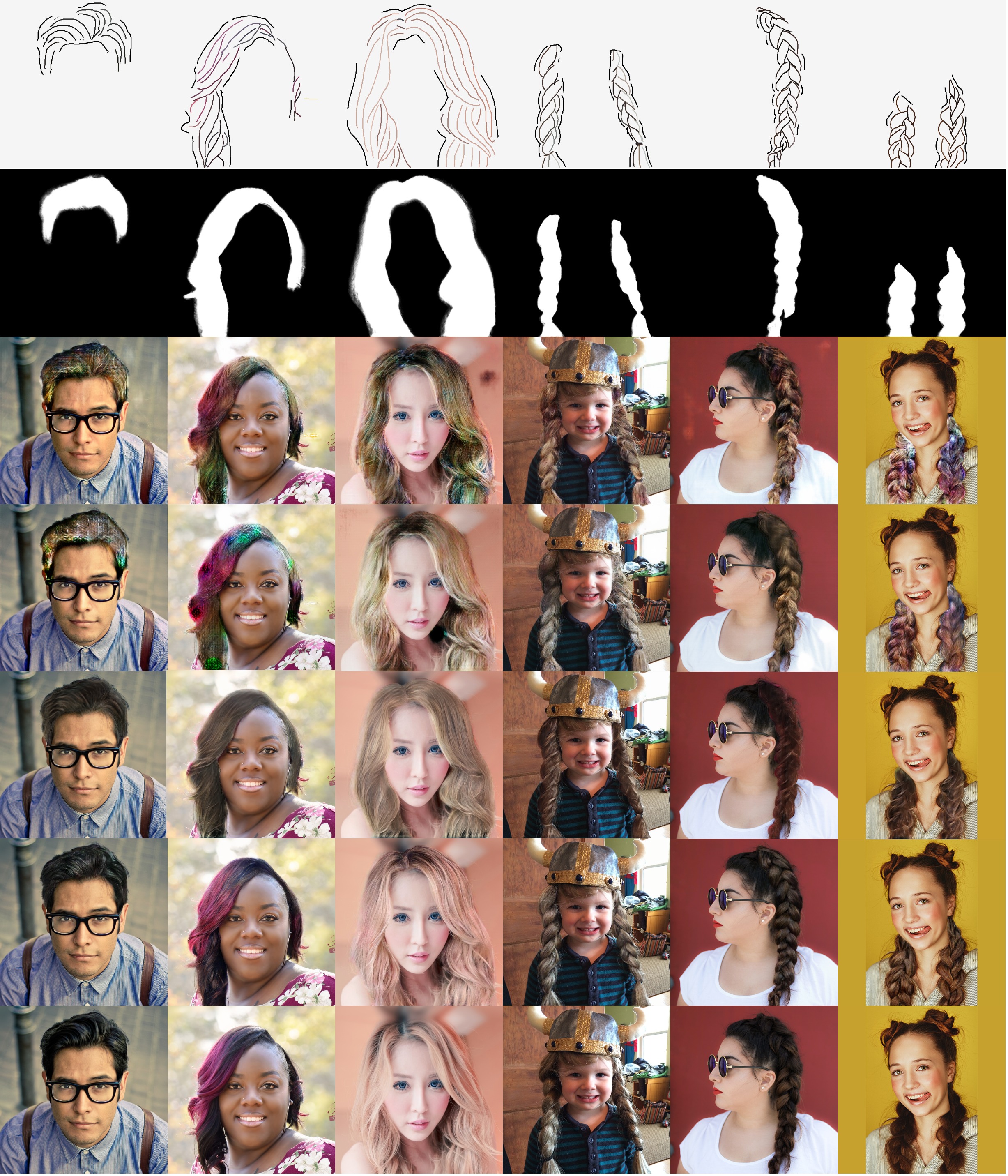}
     \end{minipage}
    \vspace{-3mm}
    
    \caption{
    Comparisons of structure reconstruction with the state-of-the-art methods given the same input sketches (Top Row) and the generated mattes via {S2M-Net} (Second Row) {as input}. 
    {The sketches contain hair strokes and non-hair strokes (in black), {and the latter} are only used for generating hair mattes but removed for hair image synthesis.}  {The left three columns are unbraided hairstyles, while the rest are braided hairstyles.} Please zoom in to better examine the quality of synthesized results by the compared methods against the ground truth. Original images courtesy of Juanjo Zanabria Masaveu, Amber Thomas, WKC, Juhan Sonin, Barion McQueen, and Konstantin Mishchenko.
    }
    \label{fig:comparison1}
\end{figure*}

{We trained and tested our proposed system \sysName~on a PC with Intel i7-8700 CPU, 32GB RAM, and a single 2080Ti GPU. Since the matte and {the} hair image to be generated are the targets of different nature, we separately train S2M-Net and S2I-Net on our dataset (augmented at each iteration). 
For S2M-Net, we train it on both of the unbraided and braided datasets. For S2I-Net, we first train the unbraided model on the unbraided dataset and then fine-tune {the unbraided model on the braided dataset to get the braided model}. 
During interactive design of hairstyles with our tool, it took around {0.5s} {on average} for synthesizing a single hair image (512 $\times$ 512) given a hair sketch, including the process of sketch to matte (Section \ref{sec:S2M}), sketch completion (Section \ref{sec:auto-comp}), and sketch to image (Section \ref{sec:S2I}). Our tool thus is able to provide {interactive} feedback to users{, though further optimization is still needed to achieve real-time feedback}. Please find more training details in the supplemental materials.
}

We have conducted extensive experiments to quantitatively and qualitatively evaluate the performance of our proposed \sysName. Below we first evaluate the quality of hair mattes generated from S2M-Net (Section \ref{exp:matte}). Then, we compare our sketch-based system with state-of-the-art image synthesis methods conditioned on {the same inputs} (Section \ref{exp:sota}). We further perform a {perception} user study and a usability user study of our system for the reconstruction results in terms of naturalness, faithfulness, and fidelity (Section \ref{exp:user_study}). A set of ablation studies are also conducted to evaluate {the effectiveness of orientation maps,} the importance of our new dataset, and the use of the self-attention modules (Section \ref{exp:ablation}).

\subsection{Quality of Hair Matte Generation}
\label{exp:matte}
To our knowledge, we are the first work that automatically predicts hair mattes given hair sketches. Thus, we focus on evaluating the quality of generated mattes. To do this, we simply compute the Sum of Absolute Differences (SAD) used in the matting work \cite{li2020natural}, to evaluate the accuracy of hair matte generation. Since the SAD metric is mostly dominated by the black and white region{s}, we also use Intersection over Union (IoU) to evaluate the border region accuracy by thresholding the generated mattes and ground truth.
The evaluation is conducted on our testing dataset (Section \ref{sec:data}). 
%\hbc{the text within the parentheses can be removed, since it is covered before?}\cfc{Yes, it was mentioned in Sec. 4. Removed.}
% (500 sketch-matte pairs, of which 400 for the unbraided type and the rest for the braided type)
To simulate users' input, we randomly generate non-hair strokes (in the same way for training). In addition, we compare S2M-Net with a variant model {given the same inputs but} without the self-attention modules.

Figure \ref{fig:matte_comparison} 
shows the generated mattes with their difference maps {against the ground truth}. It shows that our S2M-Net produces more plausible results  than the model without the self-attention modules. This is also confirmed by the analysis based on SAD {and IoU}: the results by our full model {(SAD=(mean: 49.782, SD: 16.07), IoU=0.882)} have lower SAD {and higher IoU values} than the model without the self-attention modules {(SAD=(mean: 52.40, SD: 16.0245), IoU=0.876)}. This is possibly because the attention modules help the network perceive hair sketches on a global level.

To show the effectiveness and the superiority of hair matte, we compare synthesized hair images based on the generated mattes with those based on the generated {binary} masks. Please find the qualitative comparison and user study analysis in the supplemental materials.

\subsection{Comparisons on {Sketch-based} Hair Image Synthesis}
\label{exp:sota}
We compare S2I-Net with the state-of-the-art methods for image synthesis conditioned on hair sketches, including pix2pix \cite{pix2pix2017}, HIS \cite{qiu2019two} and MichiGAN \cite{tan2020michigan} in terms of the visual quality of {generated hair images}. % hair image generation. 
For a fair comparison, the inputs to these methods are the same as those to S2I-Net, i.e., hair mattes, hair sketches, and background regions, though the ways to control the appearance are slightly different between MichiGAN and the other methods. Please find the details of training the compared methods in the supplemental materials.
The evaluation task performs like hair image reconstruction. All of the methods are trained and tested on our constructed dataset (with the same augmentation tricks) with the resolution 512 $\times$ 512. Note that we separately train the unbraided model and braided model for all the compared methods (the same way to train our models) until convergence. During testing, each stroke is assigned the average color of the corresponding pixels in the hair images along the stroke.

Qualitatively, Figure \ref{fig:comparison1} shows representative results given the same hair sketches and generated hair mattes from our S2M-Net. The three left columns are unbraided hairstyles, while the rest are braided hairstyles.
Our method can handle diverse hairstyles the best and produce visually pleasing results that faithfully respect the input sketches in terms of both structure and appearance. 
Taking a closer look at the details of the results, we find that for the orientation map-based methods, i.e., HIS and MichiGAN, their generated results (Figure \ref{fig:comparison1} (Rows 4 \& 5)) look flat and lack the occlusion structures depicted by the sketches (Row 1), especially for the braided hairstyles (Columns 4-6). It is mainly because the orientation maps may wash away the key global structures. For the appearance, MichiGAN could not restore well the local appearance due to the setting of its mean-feature extraction from the reference on the appearance encoder. HIS can avoid moderately this issue with the colored sketches in its second stage, but it seems to spray the sketch colors onto the images to approximate the desired appearance, due to the effect from the orientation maps.
In contrast, the results of pix2pix (Row 3) contain too noticeable artifacts at many places. For the unbraided hairstyles, the artifacts are always located at the hair regions not covered by the input strokes.
For the braided hairstyles, artifacts often appear at the junction structures, thus failing to restore the braided structures. 
It seems that pix2pix is hard to capture the local and global structures as well as the appearance from the sparse hair strokes without the self-attention modules used in ours.
{In addition, some artifacts leak out the hair region to the background. 
{This is mainly because for fair comparisons we concatenate the background region with the sketch as inputs (see the supplemental materials) and pix2pix could not perform well for jointly synthesizing the hair image and reconstructing the given background blended with the foreground. A similar issue also exists with HIS.}
}
More qualitative comparisons are shown in the supplemental materials.

Quantitatively, we compute Fr\'{e}chet Inception Distance (FID) to measure the distribution similarity, between the results and the ground truth, and the fidelity of the generated results to some extent. As shown in Table \ref{tab:fid_is}, our method {significantly} outperforms the compared methods in terms of FID, indicating that our results are the  most similar to the ground {truth} %truths 
on the deep features level. This is consistent with {our findings based on} the qualitative comparisons.

\begin{table}[htb]
\caption{Fréchet inception distance (FID) for different settings.}
\label{tab:fid_is}
% \small
\begin{tabular}{c|l|l}
\hline
Type                               & \multicolumn{1}{c|}{Model}    & \multicolumn{1}{c}{FID $\downarrow$} \\ \hline
\multirow{3}{*}{Different Methods} & Pix2Pix                       & 31.336                  \\ \cline{2-3} 
                                   & HIS                           & 29.857                  \\ \cline{2-3} 
                                   & MichiGAN                      & 24.327                  \\ \hline
\multirow{2}{*}{Ablation Study}    & Trained on Synthetic Dataset  & 21.825                  \\ \cline{2-3} 
                                   & w/o Attention Modules         & 23.902                  \\ \hline
\multirow{2}{*}{Different Inputs}  & Orientation Map               & 26.656                  \\ \cline{2-3} 
                                   & Orientation Map + Hair Sketch & 17.798                  \\ \hline
\multicolumn{2}{c|}{Ours}                                          & \textbf{15.553}         \\ \hline
\end{tabular}
\end{table}
\begin{figure}[htb]{
    \centering
    \subfloat[Naturalness]{
        \includegraphics[width=.35\linewidth]{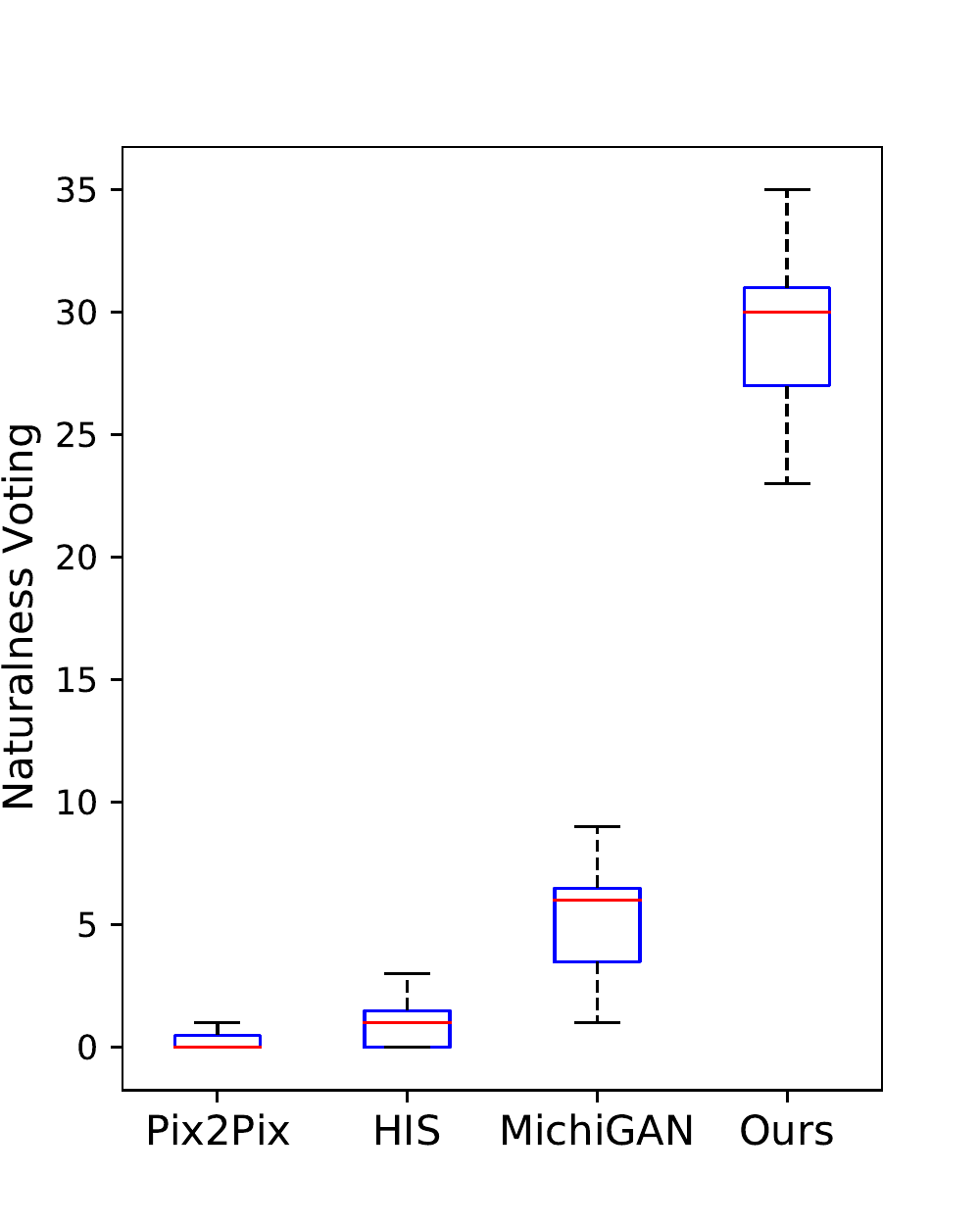}
    }\hspace{-4mm}
    \subfloat[{Faithfulness}]{
        \includegraphics[width=.35\linewidth]{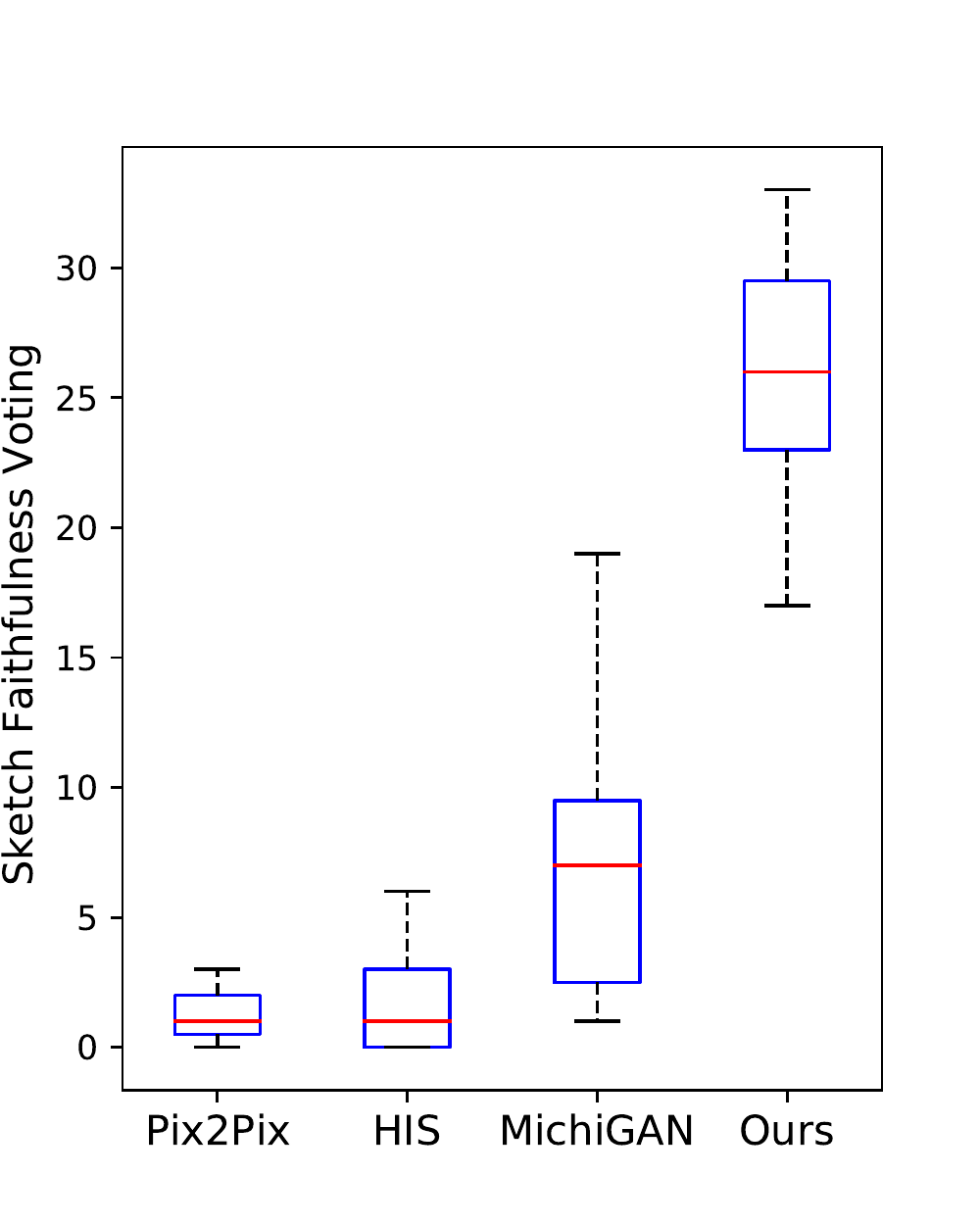}}\hspace{-4mm}
    \subfloat[Fidelity]{
        \includegraphics[width=.35\linewidth]{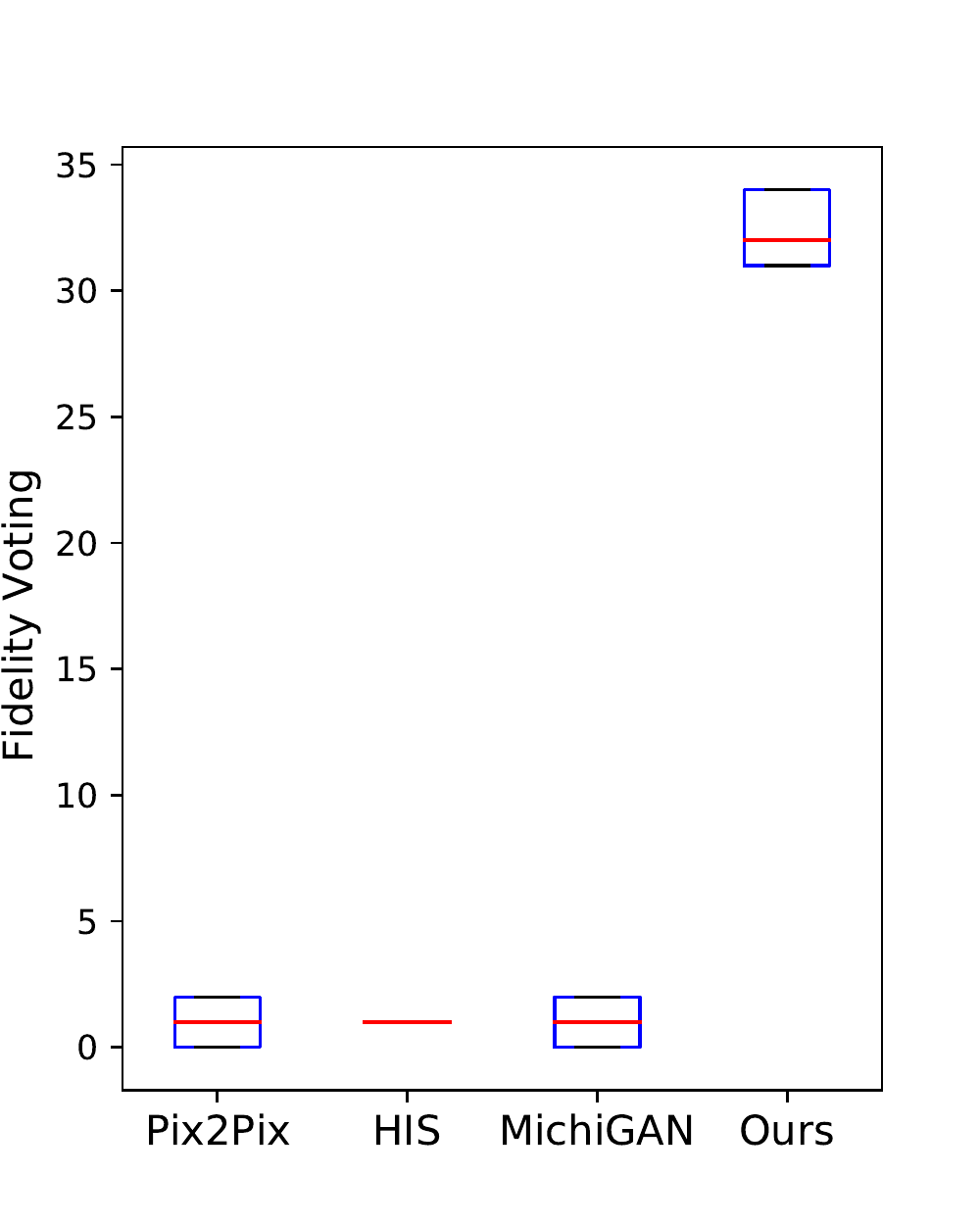}\hspace{-4mm}
    }
    \caption{Box plots of the average naturalness, {sketch faithfulness, and fidelity} voting over the {prepared questions for each method.}
    }
    \label{fig:box_plot}
    }
\end{figure}

\subsection{Perception and Usability Studies}
\label{exp:user_study}
We conducted two types of user studies to evaluate the performance of our system, including a {perception} study and a usability study. 
Some of the invited participants participated in both studies. We thus asked the users to finish the {perception} study ahead of the usability study since reversing the order of the two studies might introduce a bias due to the exposure of the users to the results by our system before the {perception} study.

{\textbf{Perception Study}.}
To evaluate the visual naturalness, 
the faithfulness of the generated hair images to the input sketches, and the fidelity of the generated results compared with their corresponding ground-truth hair images, we conducted a perception study, including three parts: \textit{naturalness study}, \textit{sketch faithfulness study}, and \textit{fidelity study}.

We prepared a set of input sketches (45 in total) with diverse hairstyles randomly picked 
(evenly covering straight, wavy, braided hairstyles) from our dataset.
The sketches were divided into three groups: 15 sketches for naturalness evaluation, 15 sketches for sketch faithfulness and the rest 15 for the fidelity evaluation. We first produced hair mattes given the sketches via our S2M-Net. 
Then, we applied the four image synthesis methods {(Section \ref{exp:sota})} to each input sketch with the corresponding synthesized matte to generate the hair images. Figure \ref{fig:comparison1} shows some representative examples used in our user study.

The evaluation was done through an online questionnaire. There were in total 36 participants for the three studies. Most of them had no professional training in drawing. 
In the naturalness study, we showed each participant sets of four images generated by the compared approaches in random order set by set. 
Each participant was asked to choose the best one in terms of visual naturalness. 
In the sketch faithfulness study, for each set of results, we showed each participant four pairs of images in random order, i.e., a hair sketch side-by-side placed with its generated image. Each participant was asked to select the best image pair where {the} synthesis result respects the input sketch the most faithfully. 
In the fidelity study, we also showed each participant four image pairs (for each set), consisting of a ground-truth hair image and {a} synthesized image by each compared method. Each participant was asked to select the best image pair where the generated result preserved the most fidelity of the {corresponding} ground-truth hair {image}. %images.
In total, we got 36 (participants) $\times$ 45 (15 sketches for naturalness, 15 for sketch faithfulness, and 15 for fidelity) = 1620 subjective evaluations.

Figure \ref{fig:box_plot} shows the statistics of the voting results. 
We conducted one-way ANOVA tests on the naturalness, faithfulness, and fidelity voting results, and found significant effects for naturalness ($F = 521.68$, $p \textless 0.001$), faithfulness ($F = 149.20$, $p\textless 0.001$), and fidelity ($F = 116.48$, $p \textless 0.001$). The further paired t-tests show that our method (mean: 29.40) got significantly more votes in the naturalness term than all the other methods, pix2pix (mean: 0.21; [$t = 33.40, p \textless 0.001$]), MichiGAN (mean: 5.33; [$t = 14.75, p \textless 0.001$] ), and HIS (mean: 1.43; [$t = 29.58, p \textless 0.001$]). In terms of faithfulness, our method also achieved the best performance (mean: 25.87) among the competitive approaches, MichiGAN (mean: 7.00; [$t = 7.70, p \textless 0.001 $]), pix2pix (mean: 2.00; [$t =20.53, p \textless 0.001$]), and HIS (mean: 1.13; [$t=17.47, p\textless 0.001$]). In term of fidelity, our method again achieved the best performance (mean: 29.80) among the compared approaches, pix2pix (mean: 1.00; [$t = 15.61, p \textless 0.001$]), HIS (mean: 1.13; [$t = 17.05, p \textless 0.001$]), and MichiGAN (mean: 4.07; [$t = 7.11, p \textless 0.001$]).

{\textbf{Usability Study.}} 
Since we are interested in checking whether our system would enable users
with little training in drawing or hairstyle design to create photo-realistic hairstyles, we invited six such users to test the usability of our system. 
At the beginning of the study, each of them was provided a Surface Pro 4 {device} {with its stylus} and a 10-minute tutorial on our system, including sketching hair strokes and non-hair strokes for generating a matte and {a synthesized} image, color brush for hair dyeing, sketch auto-completion for unbraided and braided hairstyles, etc. The study {consisted} of fixed-task and open-ended sections. For the fixed-task section, {each of the users was} %the users were 
given portrait images with specified hairstyles and asked to reproduce the hairstyles as much as he/she could in terms of both structure and appearance. The users were allowed to trace the key strokes over the given hair images as done for creating our dataset. For the open-ended task, the users were given portrait images with little or short hair so that they could focus on designing desired hairstyles almost from scratch instead of hairstyle editing.

\begin{figure}[t]{
    \includegraphics[width=\linewidth]{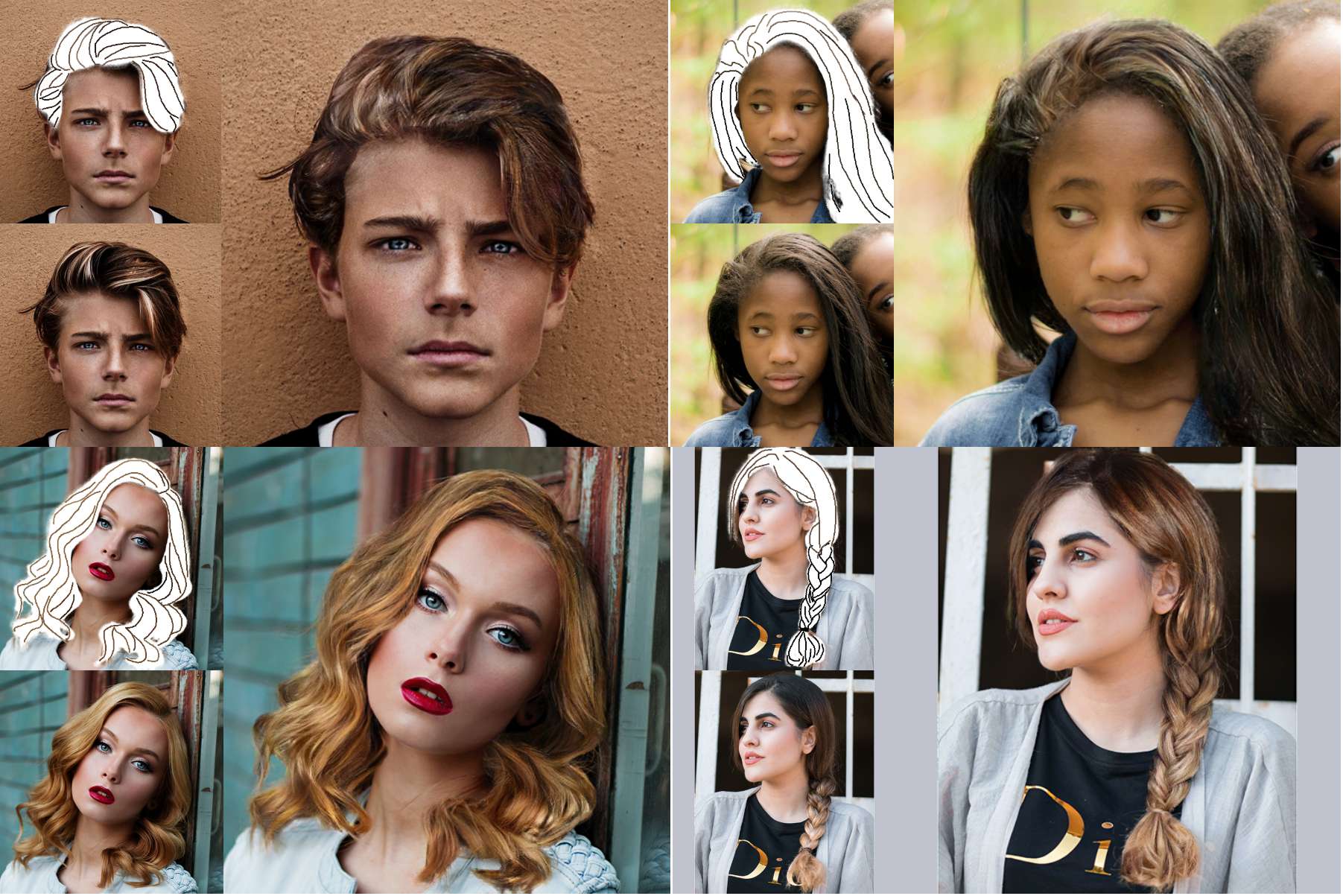}
     \caption{{Four group of results that the users finished in the fixed-task usability study. For each group, the left-top is the user-created sketch including the hair strokes and non-hair strokes ({omitted in the images}), and the left bottom is the target hairstyle the users were asked to {reproduce}, while the right one is the generated result by our system. Note that we asked the users to try to assign the closest colors with the target appearance to the hair strokes by themselves. {Original images courtesy of Oliver Ragfelt, Nastya Gepp, Jathan Johnston, and Sina Rezakhani.}}
    }
    \label{fig:user_study_recon}
}
\end{figure}
\begin{figure}[htb]{
    \centering
    \vspace{-2mm}
    \hspace{-3mm}
    \subfloat[]{
        \includegraphics[width=.425\linewidth]{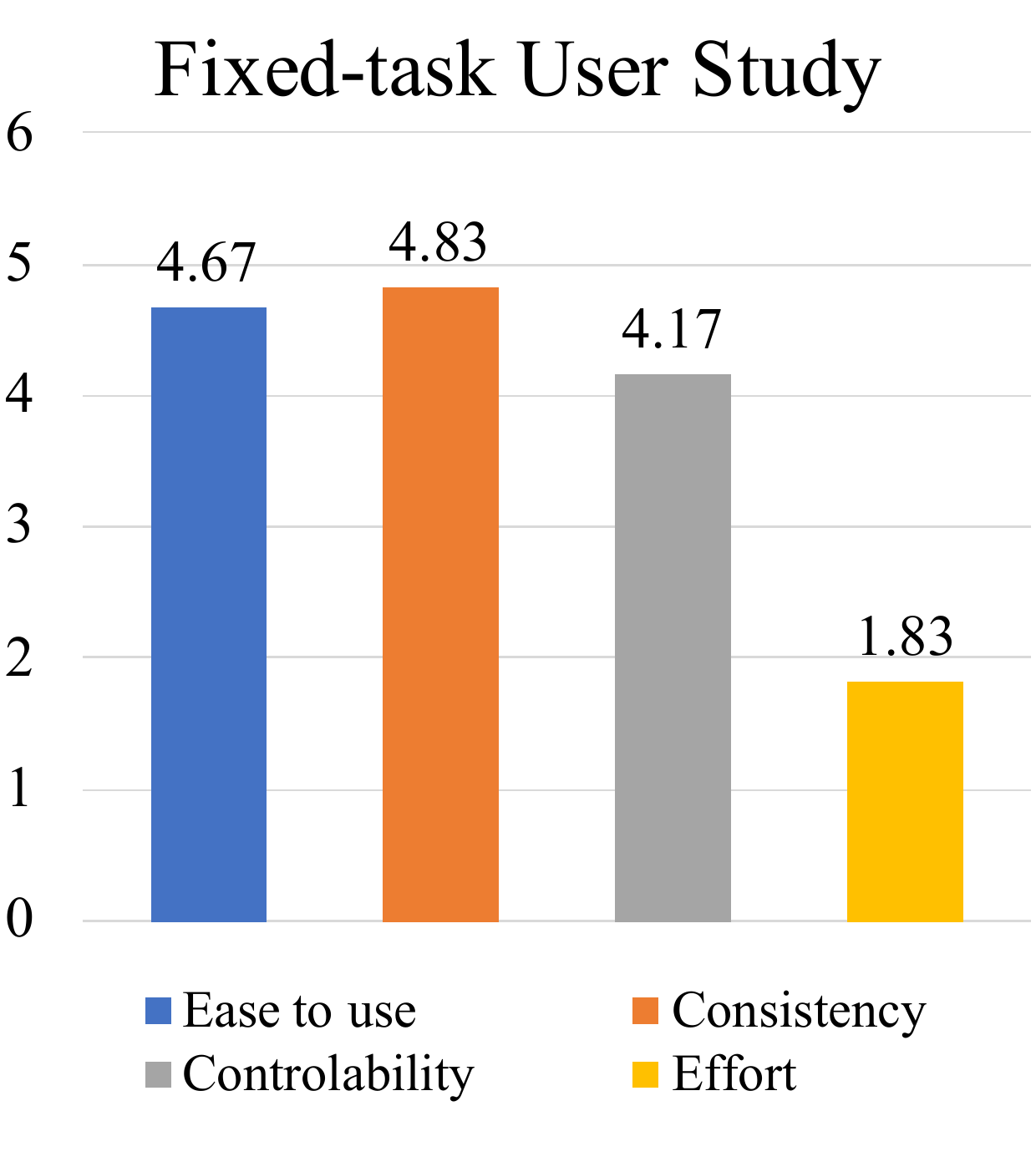}
    }\hspace{-2mm}
    \subfloat[]{
        \includegraphics[width=.565\linewidth]{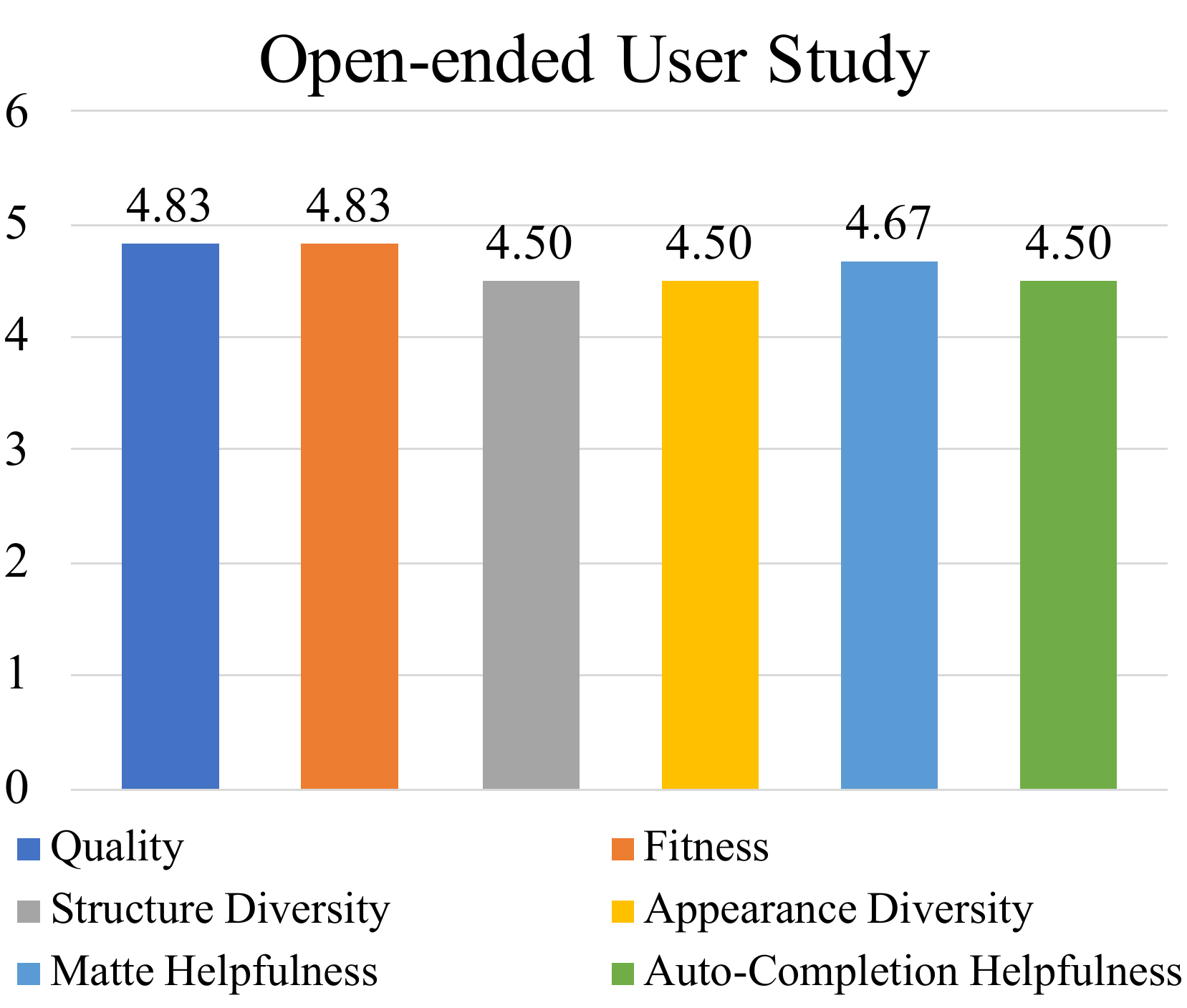}}\hspace{-4mm}
    % \caption{Box plot.}
    \caption{{The average {evaluation} scores of the fixed-task (a) and open-ended (b) studies {from different perspectives}. %in different terms. 
    (a) From left to right each bar is respectively for ease to use, the consistency of a {reproduced} %reconstructed 
    hairstyle with a given hairstyle, the controllability of structure and appearance, {and} the effort {for reproducing a given hairstyle}. %t for reconstruction. 
    (b) From left to right each bar is respectively for result quality, expectation fitness, structure diversity, appearance diversity, matte helpfulness, sketch auto-completion helpfulness for freely creating desired hairstyles.}}
    \label{fig:sus}
    }
\end{figure}

\begin{figure}[htb]{
    % \includegraphics[width=.331\linewidth]{fig/results_v2/user_study/design/sk/59113_my.png}
    % \hspace{-1.6mm}
    % \includegraphics[width=.331\linewidth]{fig/results_v2/limitation/R2_1543_sk.png}
    % \hspace{-1.6mm}
    % \includegraphics[width=.331\linewidth]{fig/results_v2/limitation/R2_1543_sk.png}
    % \\
    % \vspace{-.5mm}
    % \includegraphics[width=.331\linewidth]{fig/results_v2/user_study/design/sk/bald_1_my_t2.png}
    % \hspace{-1.6mm}
    % \includegraphics[width=.331\linewidth]{fig/results_v2/limitation/spiral_sk.png}
    % \hspace{-1.6mm}
    % \includegraphics[width=.331\linewidth]{fig/results_v2/limitation/spiral_sk.png}\\
    %  \vspace{-.5mm}
    % \includegraphics[width=.331\linewidth]{fig/results_v2/user_study/design/sk/58851_my_t1.png}
    % \hspace{-1.6mm}
    % \includegraphics[width=.331\linewidth]{fig/results_v2/limitation/spiral_sk.png}
    % \hspace{-1.6mm}
    % \includegraphics[width=.331\linewidth]{fig/results_v2/limitation/spiral_sk.png}\\
    %  \vspace{-.5mm}
    % \includegraphics[width=.331\linewidth]{fig/results_v2/user_study/design/sk/59113_my.png}
    % \hspace{-1.6mm}
    % \includegraphics[width=.331\linewidth]{fig/results_v2/limitation/spiral_sk.png}
    % \hspace{-1.6mm}
    % \includegraphics[width=.331\linewidth]{fig/results_v2/limitation/spiral_sk.png}\\
    %  \vspace{-.5mm}
    % \includegraphics[width=.331\linewidth]{fig/results_v2/user_study/design/sk/59113_my.png}
    % \hspace{-1.6mm}
    % \includegraphics[width=.331\linewidth]{fig/results_v2/limitation/spiral_sk.png}
    % \hspace{-1.6mm}
    \includegraphics[width=\linewidth]{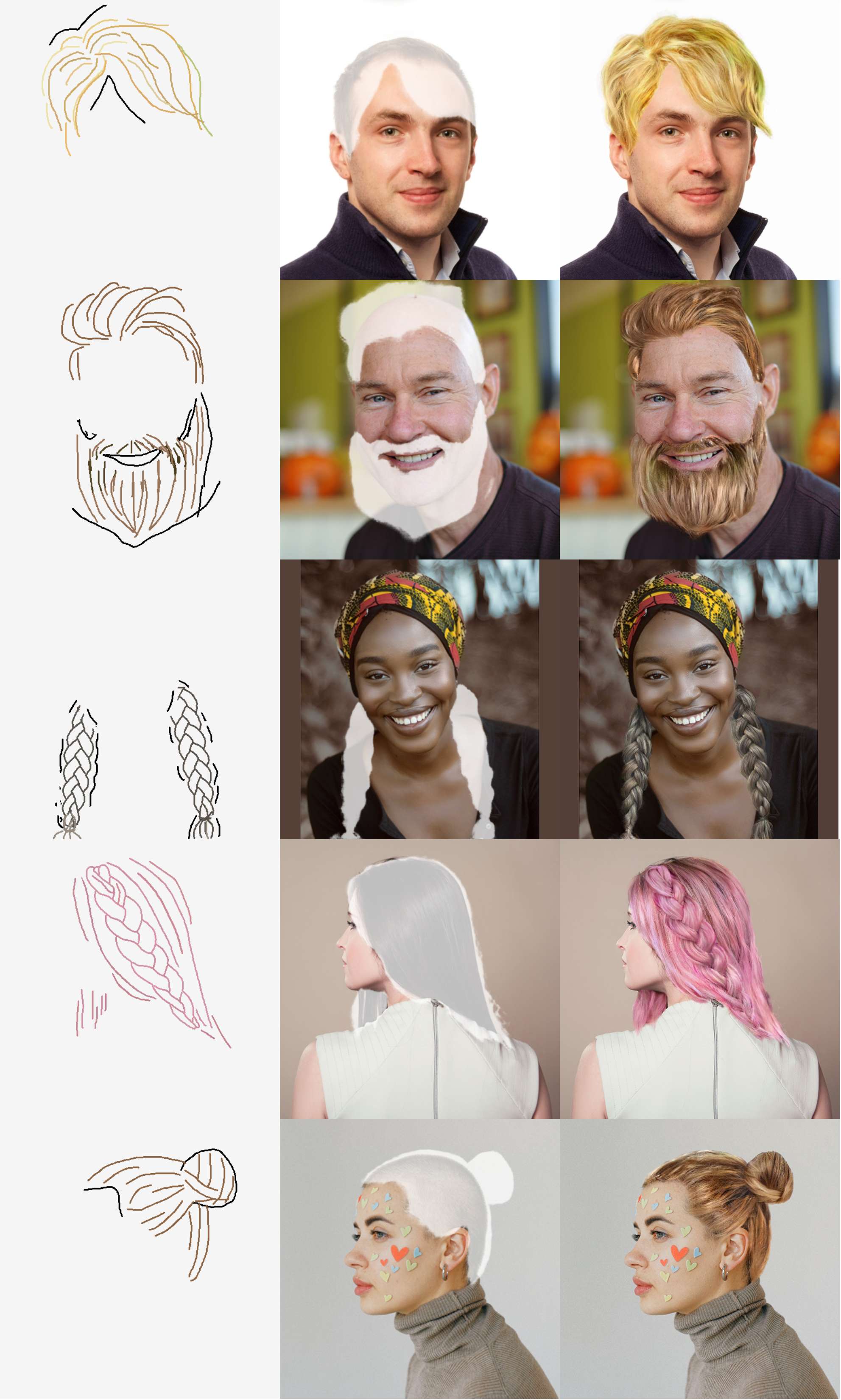}\\
    \qquad (a) Inputs \quad \qquad(b) Generated Mattes  \quad \qquad (c) Results \qquad \qquad 
     \caption{{Diverse hairstyles designed by the users in the open-ended usability study. The users sketched (a) the hair strokes and non-hair strokes (in black) and then our system \sysName~produced the hair mattes (b) and the realistic hair images (c). All the results are visually pleasing, even for the facial hair (i.e., beard, shown as Row 2), which is not included in our training dataset. {Original images courtesy of Sebastiaan Ter Burg, Chris Hunkeler, Prince Akachi, Khusen Rustamov, and Anna Shvets.}}
    }
    \label{fig:design}
}
\end{figure}

Figure \ref{fig:user_study_recon} shows the reproduced hairstyles by the users for the fixed-task section. It can be seen that they are very close to the given hairstyles in terms of structure and appearance. Figure \ref{fig:design} shows some examples of diverse hairstyles designed by the users for the open-ended task. All the resulting hairstyles are visually pleasing and faithfully respect the input sketches. All of the users felt that our system was powerful and efficient to design their hairstyles of interest using sparse sketches. They felt they had full control of the generated results.

At the end of the study, each participant was asked to fill in a questionnaire of customized five-point System Usability Scale (SUS, 1 = strongly disagree to 5 = strongly
agree) for the two tasks to evaluate the key components of our system. All of them believed that our system was efficient and helpful for creative design of hairstyles with {a} %the 
large diversity in structure and appearance. The time cost for creating a single hairstyle ranged from 32s to 4min, depending on the complexity of the hairstyles. Figure \ref{fig:sus} shows the average scores for different perspectives. Overall our system was rated positive for every perspective. 
There are three main features we asked the users to evaluate, including matte generation, hair dyeing, and sketch auto-completion. They particularly liked the sketch-to-matte feature since it was very helpful for them to produce desired hair shapes simply via sketching with a very few hair and non-hair strokes. They appreciated the color brush tool for hair dyeing, since they could easily try diverse alternative colors for their desired hairstyles. For sketch auto-completion, they found that the auto-completion for unbraided sketches was useful but not necessary, since it was not very difficult for them to create a more complete sketch by themselves. However, for braided sketches, the auto-completion feature helped them a lot, and they thought that this feature was amazing and helpful. We also compute the saved time of sketch auto-completion by asking the users to trace the auto-generated part. On average, 17.15s per-example was saved for unbraided auto-completion, while 47.94s per-example for braided auto-completion.

\subsection{Ablation Study}
\label{exp:ablation}
\begin{figure}
    \centering
    \begin{tabular}{c}
     \hspace{-3mm}
     \includegraphics[width=\linewidth]{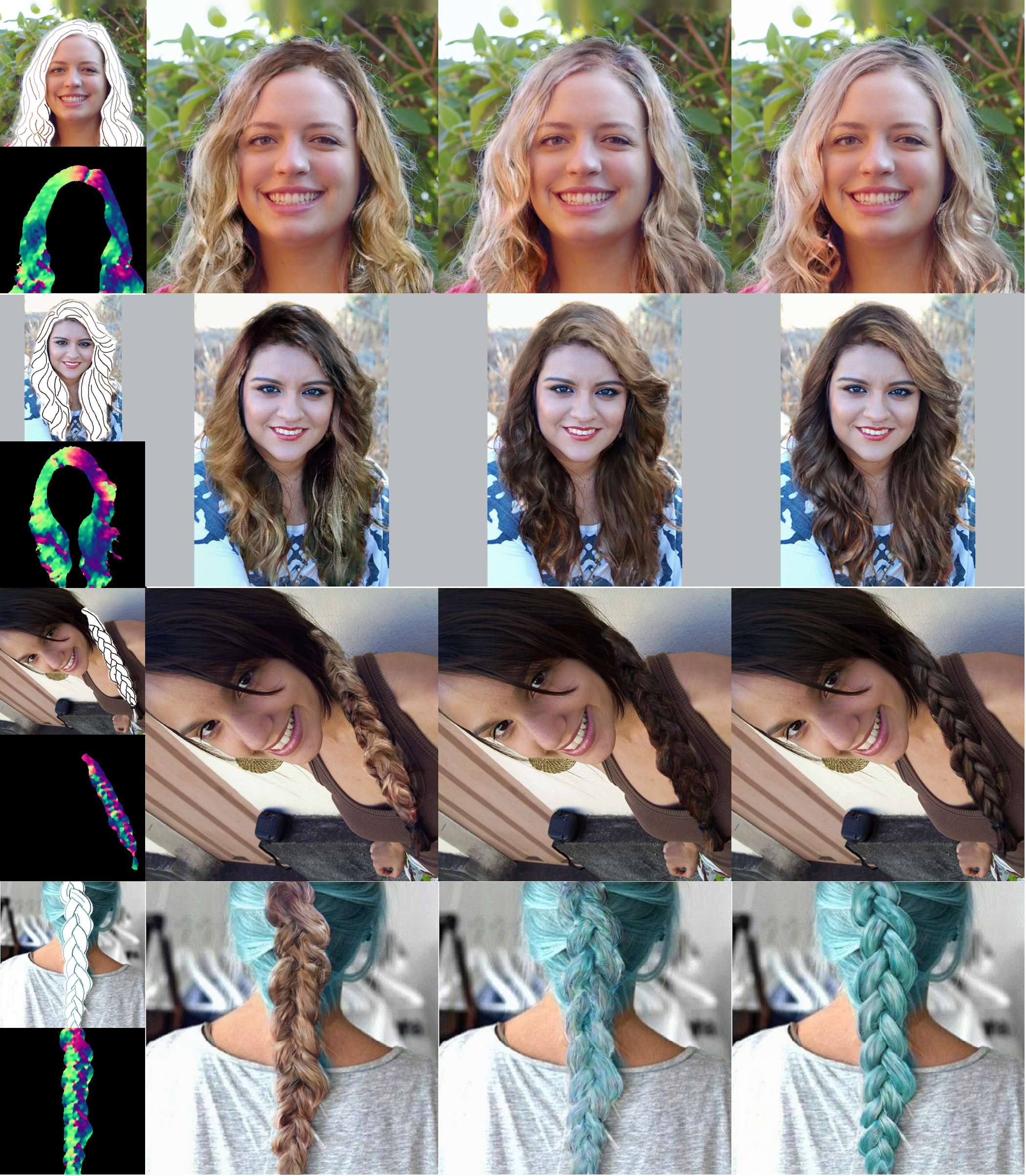}\\
     (a) \qquad  \qquad \quad (b)  \qquad  \qquad \qquad (c)\quad \qquad \qquad \qquad (d)  \qquad \qquad 
    \end{tabular}
    \caption{Comparisons of the results generated from our model incorporated with and without orientation maps: (b) using orientation maps alone; (c) using both sketches and orientation maps; (d) using sketches alone. For each {pair} of (a), the top one is the sketch and background inputs, while the bottom one is {a dense} %the smoothing
    orientation map {predicted from the sketch}. Note that the mattes used in this experiment are generated by our S2M-Net, with hair strokes and randomly selected non-hair strokes as input. {Original images courtesy of Lydia Liu, Dréa Rewal, Jipe Martins, and Wicker Paradise.}
    }
    \label{fig:ablation_input}
\end{figure}

\begin{figure}[htb]{
    \centering
    \begin{tabular}{c}
     \hspace{-3mm}
     \includegraphics[width=\linewidth]{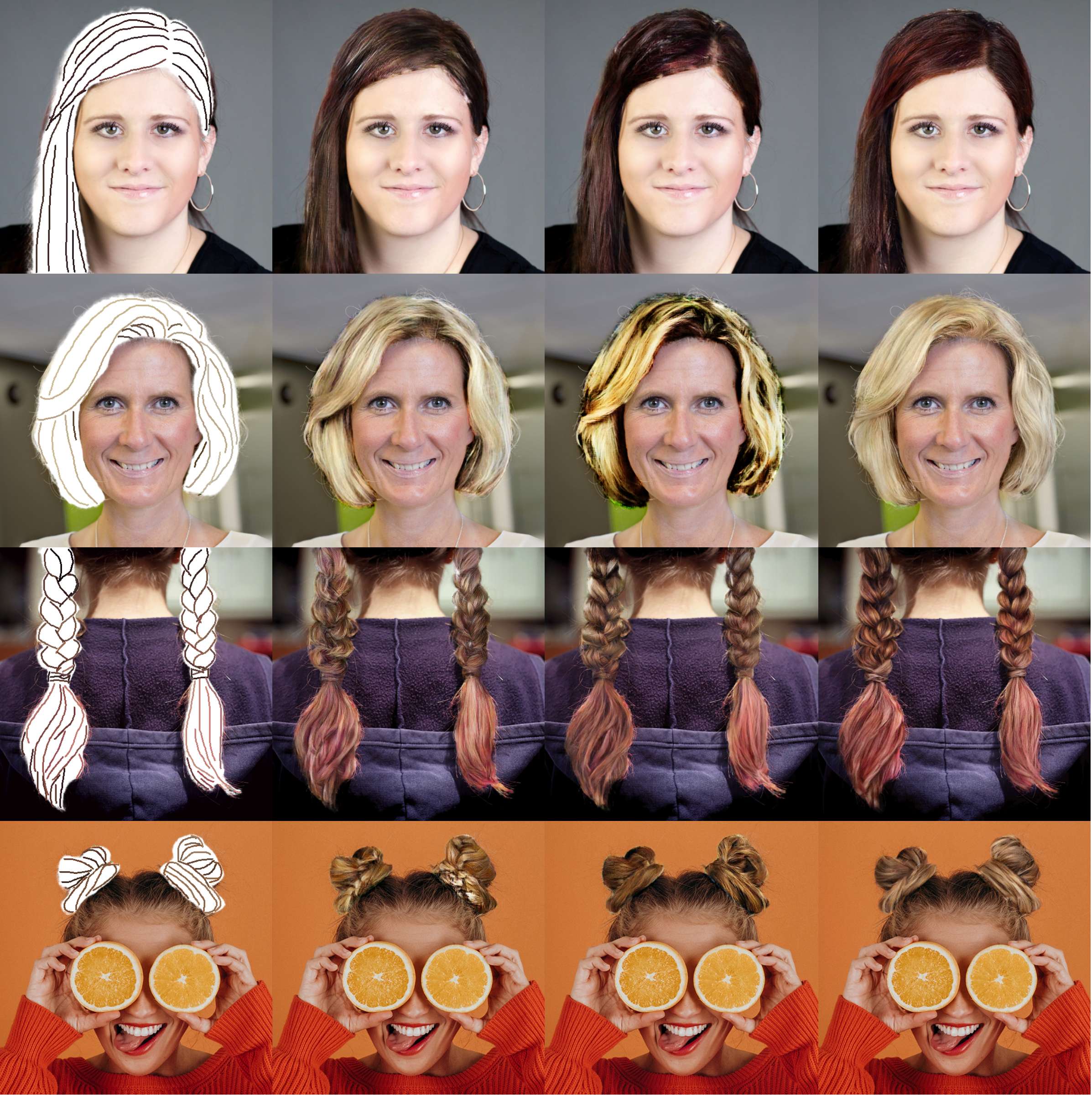}\\
    (a) \qquad  \qquad \quad (b)  \qquad  \qquad \qquad (c) \qquad \quad \qquad (d)  \qquad 
    \end{tabular}
     \caption{Comparison of the results generated from our model variants with different settings given the same inputs (a). (b) Trained on the synthetic dataset. (c) Without the attention modules, (d) Our full model. {Original images courtesy of Lucky Lynda, Microsoft Sweden, Emma Simpson, and Noah Buscher.}
     }
    \label{fig:ablation_variants}
    }
\end{figure}

In this subsection, we conducted two sets of ablation studies: 
one aims to verify the effectiveness of orientation maps for the generation of hair images with complex hairstyles (Figure \ref{fig:ablation_input}) and the other is to validate the key components of our method (Figure \ref{fig:ablation_variants}).

In our pipeline, we directly feed an input sketch to the generation network to produce a hair image, instead of predicting an orientation map from the sketch as an intermediate input as done in HIS \cite{qiu2019two} and MichiGAN \cite{tan2020michigan}. To evaluate the effectiveness of orientation map, we performed a set of ablation studies on our model with different inputs, including orientation map only, both sketch and orientation map, sketch only (ours).
Here, we predicted the dense %\hbc{`smoothing' sounds weird. dense?}\cfc{Sure, modified.}
orientation maps from hair sketches for training and testing, similar to MichiGAN.

As shown in Figure \ref{fig:ablation_input} ({b}), the model taking as input an orientation map only generates fluttering hair textures and blurry boundaries of hair-wisp occlusions. It indicates that the dense and local information in the orientation map would make the network focus too much on synthesizing local details but be {less sensitive} to the global structures. Using the orientation map only does not allow the control of hairstyle appearance. 
Incorporating both orientation map and sketch can alleviate these issues to some extent (Figure \ref{fig:ablation_input} ({c}). However, {although the stroke colors could be spread into the final results (conditioning the hairstyle appearance),} the orientation maps limit the hair sketches to express their global structures in the network. 
Figure \ref{fig:ablation_input} ({d}) shows our results that are highly faithful to the sketches in terms of appearance and the local and global structures.

Table \ref{tab:fid_is} also confirms our observation with the quantitative results, of which FID get worse when the input is merely an orientation map or a sketch incorporated with it. 
From both qualitative and quantitative results, we claim that orientation maps cannot benefit our task of sketch-based hair image synthesis, further confirming our key motivation of using sketches directly to predict hair images.

To further prove the effectiveness of our method {and our dataset}, we ablated one of the key components (i.e., manually annotated sketches and self-attention mechanism) alternatively to show their necessity and importance for training S2I-Net.
The manually annotated sketches as input play an essential role in our method to correctly capture the global structure{, orientation flow, and local appearance} of the underlying hair strands. To show their effects, we trained another model with the same architecture as S2I-Net, but using a synthetic dataset by auto-tracing strokes from orientation maps  \cite{olszewski2020intuitive}. Each auto-traced stroke is also color-coded (Figure \ref{fig:dataset} (c)), as ours does.
From the results {trained} based on the synthetic dataset, as shown in Figure \ref{fig:ablation_variants} (b), we can observe the hair wisps are not well organized to form junctions and occlusions but perform like flat effects, unfaithful to the hair sketches. {It seems that this model can learn the hair appearance from the colored strokes but is unaware of the structure. This is possibly because the synthetic sketches are not very correct to depict the hair structures for training the S2I-Net.}
In contrast, our manually annotated sketches lead to {the} results (Figure \ref{fig:ablation_variants} (d)) that more faithfully respect the structures depicted by the input sketches. 

Considering the limited receptive field of convolution neural networks (CNN), we adopt self-attention modules \cite{fu2019dual} in S2I-Net to efficiently learn the long-range dependency among hair strokes. 
When removing the attention modules from S2I-Net, we get rather rough results with mussy hair wisps (Figure \ref{fig:ablation_variants} (c)). It means the generated results could not well retain the coherent wisps or clear occlusions defined by {colored} hair sketches.

Besides the qualitative analysis, we also conducted the quantitative evaluation on the aforementioned ablation study to verify the contributions of different key components. It is shown from Table \ref{tab:fid_is} that our full model achieves {much} better performance than the compared alternatives. This is consistent with our {above} observations based on the qualitative results.

{\section{Conclusion and Discussions}} 
In this work we presented \sysName, a novel sketch-based system for users to design {photo-}realistic hair images with various hairstyles. Our tool is enabled by two key contributions: a reasonably large dataset of hair sketch-image pairs with corresponding hair mattes, and a two-stage framework consisting of the sketch-to-matte and sketch-to-image networks. We also present two sketch auto-completion mechanisms to reduce the workload of users.
Extensive experiments have shown that our proposed method outperforms the existing methods for hair synthesis and the alternative solutions. We will release the dataset and the code to the research community.

\begin{figure}[t]{
    \includegraphics[width=.331\linewidth]{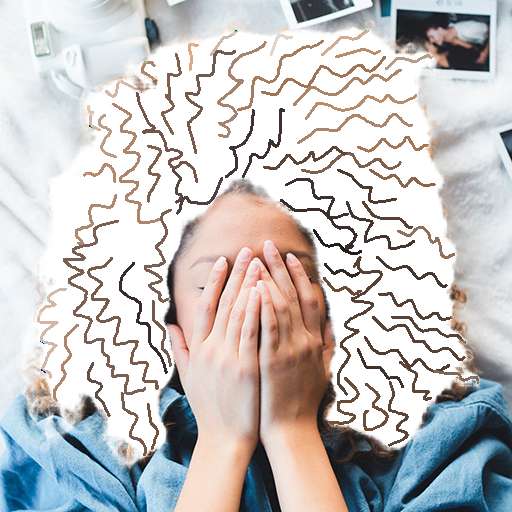}
    \hspace{-1.6mm}
    \includegraphics[width=.331\linewidth]{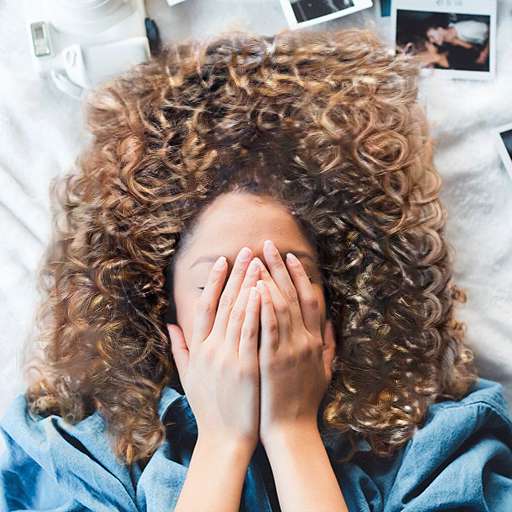}
    \hspace{-1.6mm}
    \includegraphics[width=.331\linewidth]{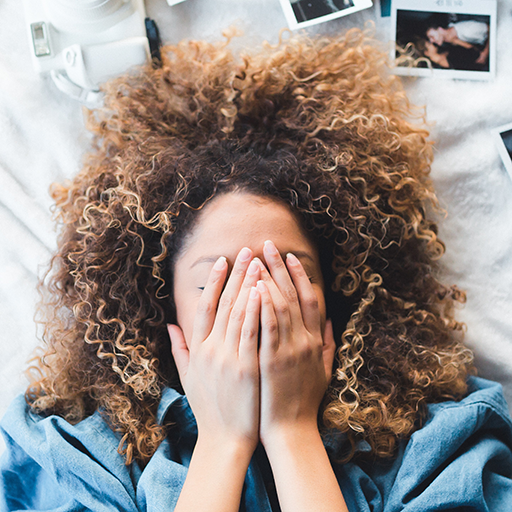}
    \\
    \vspace{-.5mm}
    \includegraphics[width=.331\linewidth]{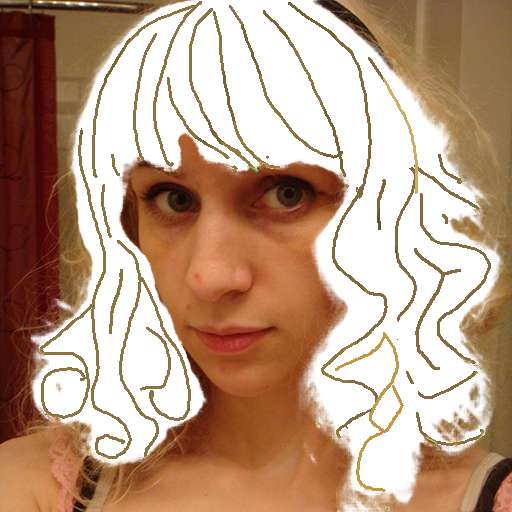}
    \hspace{-1.6mm}
    \includegraphics[width=.331\linewidth]{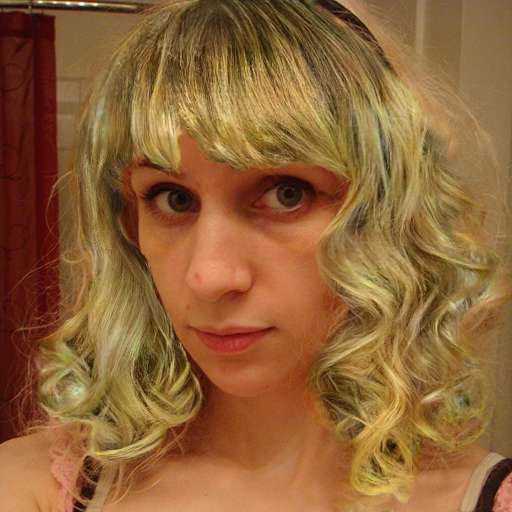}
    \hspace{-1.6mm}
    \includegraphics[width=.331\linewidth]{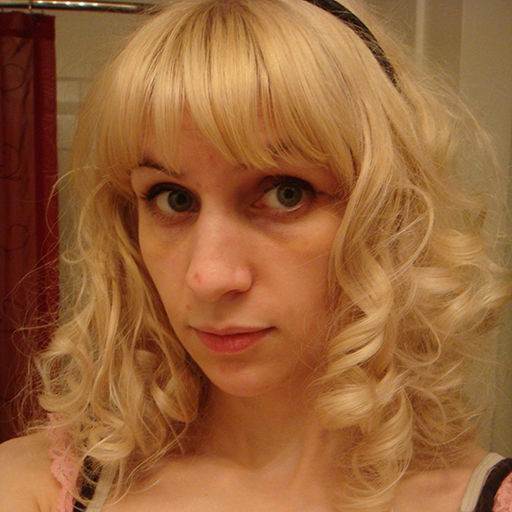}\\
     (a) Input Sketches \qquad \quad (b) Results \qquad \quad (c) Ground Truth
    \caption{Two less successful examples. The top row shows the unnatural result lacking enough layering effects, while the bottom row shows the failure case for restoring the self-occluded wisp strands. {Original images courtesy of Ian Dooley and Debs.}}
    \label{fig:fail_case}
}
\end{figure}

While our system is able to achieve impressive results with diverse hairstyles, it has several limitations.
First, for the hairstyles with too many layers shaped by a lot of wisps strands, our method could not restore such complicated layering effects merely from hair sketches without depth information, as shown in Figure \ref{fig:fail_case} (Top). For {certain hairstyles such as {dreadlocks,} spirals, and coils, which are hard to depict by 2D hair sketches}, {our method might lead to results that are visually similar to but structurally different from the desired results.} Figure \ref{fig:fail_case} (Bottom) shows such an example where we try to use {one single stroke to depict a self-occluded wisp (the spiral part as shown in the ground truth)} but our method fails to produce this type of structures and instead generates a wavy hairstyle (mostly learned from our dataset). This issue might be addressed by providing additional information (e.g., the depth along each stroke) to resolve ambiguities or preparing dedicated datasets containing a rich set of such hairstyles. 
Additionally, in some cases, the lighting condition is inconsistent between the generated hair and the background, causing some {noticeable} artifacts. We find that relighting the foreground, especially for hair, to harmonize the background is actually a challenging and open problem \cite{wang2020single}. A more elaborate image generator used as a post-process could potentially alleviate this issue. Another possible direction is to follow the state-of-{the-}art portrait relighting works \cite{wang2020single,pandey2021total}, though it might require solving a challenging problem of modeling 3D hair geometry.

Since our system is mainly for hairstyle design from scratch but not hairstyle editing, we directly take the original images as background inputs, without considering the existing hair there. Thus, users can only design new hairstyles completely covering the original hair. We are interested in extending our approach to hairstyle editing and anticipate at least two problems: how to reconstruct the existing hair using our sketch-based representation, and how to fill the missing background region originally blocked by the hair. In addition, we believe that our created dataset  can benefit other hair-related applications like %be used for 
3D hair modeling conditioned on hair sketches to create various 3D hairstyles, similar to \textit{DeepSketchHair} \cite{shen2020deepsketchhair}.
\leavevmode \\

\begin{acks}
{We thank the anonymous reviewers from SIGGRAPH 2021 and SIGGRAPH Asia 2021 for the constructive comments. }
This work was supported in part by unrestricted gifts from Adobe and grants from the Research Grants Council of the Hong Kong Special Administrative Region, China (No. CityU 11212119), City University of Hong Kong (No. 9667234, 7005176), the National Key Research \& Development Program of China (2018YFE0100900), the NSF China (No. 62172363, {61902334}), %, NSFC-61902334,
and the Centre for Applied Computing and Interactive Media (ACIM) of School of Creative Media, CityU.

\end{acks}

\bibliographystyle{ACM-Reference-Format}
\normalem
\bibliography{main}


\section*{Supplementary material}

# Supplemental Materials
# *SketchHairSalon*: Deep Sketch-based Hair Image Synthesis

CHUFENG XIAO, School of Creative Media, City University of Hong Kong
DENG YU, School of Creative Media, City University of Hong Kong
XIAOGUANG HAN, SSE, The Chinese University of Hong Kong, Shenzhen
YOUYI ZHENG, State Key Lab of CAD&CG, Zhejiang University
HONGBO FU*, School of Creative Media, City University of Hong Kong

## 1 METHOD

### 1.1 Self-Attention Maps

Due to the relatively interpretable property of the attention modules, we can visualize the attention maps generated in S2M-Net to find out what the network learns. As shown in Figure 1, the most attention is assigned to the regions surrounding the input strokes. This is reasonable to encourage the network to produce plausible hair shapes.

### 1.2 Network Architecture and Parameter Settings

We adopt a two-stage framework to synthesize hair images directly from sketches. The two main networks in our framework, S2M-Net and S2I-Net, share a similar encoder-decoder architecture, as shown in Figure 2. For S2M-Net, the encoder consists of eight vanilla convolution layers, while the decoder has eight deconvolution layers with self-attention modules. S2I-Net extends from S2M-Net by adding one background encoder for blending features between the background and foreground regions, followed by one more convolution layer. We utilize skip connections between the encoder and the decoder.

We trained and tested our proposed system *SketchHairSalon* on a PC with Intel i7-8700 CPU, 32GB RAM and a single 2080Ti GPU. Since the matte and the hair image to be generated are the targets of different natures, we separately train S2M-Net (batch size = 20) and S2I-Net (batch size = 4) for 200 epochs on our dataset (augmented at each iteration) using the Adam solver with the fixed learning rate 1e-4 via the PyTorch framework. For S2M-Net, we train it on both of the unbraided and braided datasets. For S2I-Net, we first train the unbraided model on the unbraided dataset and then fine-tune the braided model (400K iterations) upon it on the braided dataset. The whole training process took around 3 days including the two stages, i.e., around 12 hours for training S2M-Net, around 2 days for training S2I-Net on unbraided hairstyles, and around 12 hours for fine-tuning S2I-Net on braided hairstyles.

*Corresponding author.

Authors' addresses: Chufeng Xiao, School of Creative Media, City University of Hong Kong, chufeng.xiao@my.cityu.edu.hk; Deng Yu, School of Creative Media, City University of Hong Kong, deng.yu@my.cityu.edu.hk; Xiaoguang Han, SSE, The Chinese University of Hong Kong, Shenzhen, hanxiaoguang@cuhk.edu.cn; Youyi Zheng, State Key Lab of CAD&CG, Zhejiang University, youyizheng@zju.edu.cn; Hongbo Fu, School of Creative Media, City University of Hong Kong, hongbofu@cityu.edu.hk.

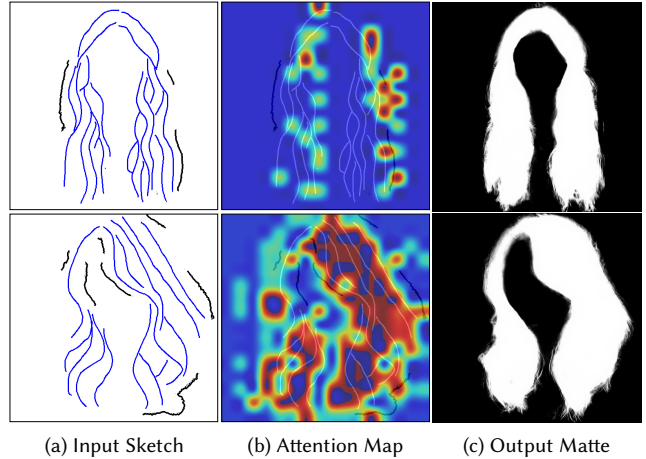

(a) Input Sketch  (b) Attention Map  (c) Output Matte

Fig. 1. Visualization of attention maps generated at the third self-attention module in S2M-Net. The generated attention maps (b) are upsampled and placed over the input sketch (a), showing where they guide S2M-Net to focus on for generating the hair mattes (c).

### 1.3 Braided Models

As describe in Section 5.3, we construct five parametric braid 3D models given users' control. We denote the control as follows:

$$\begin{cases} \triangle Y = (B_{Y0} + B_{Y1})/2 \\ \triangle X = (B_{X0} + B_{X1})/2 \\ a(t) = |B_{X0} - B_{X1}|/2 \\ t = [0, |\triangle Y|] \end{cases}, \quad (1)$$

where $B_0$ and $B_1$ are the two boundary strokes. We also provide a extra parameter $w$ directly from users, which can control the number of braided knots and the knot direction. With the parameters from users, the braid models can be reshaped. We provide five braided hairstyle models, including fishtail, rope, three-strand, four-strand and five-stand. Their detailed definitions are given below:

- Fishtail:

$$\begin{cases} L_0 : x = a(t)\sin(wt) + \triangle x, \ y = \triangle y, z = b\sin(2wt) \\ L_1 : x = a(t)\sin(wt + 2\pi/5) + \triangle x, y = \triangle y, z = b\sin(2(wt + 2\pi/5)) \\ L_2 : x = a(t)\sin(wt + 4\pi/5) + \triangle x, y = \triangle y, z = b\sin(2(wt + 4\pi/5)) \\ L_3 : x = a(t)\sin(wt + 6\pi/5) + \triangle x, y = \triangle y, z = b\sin(2(wt + 6\pi/5)) \\ L_4 : x = a(t)\sin(wt + 8\pi/5) + \triangle x, y = \triangle y, z = b\sin(2(wt + 8\pi/5)) \end{cases}, \quad (2)$$

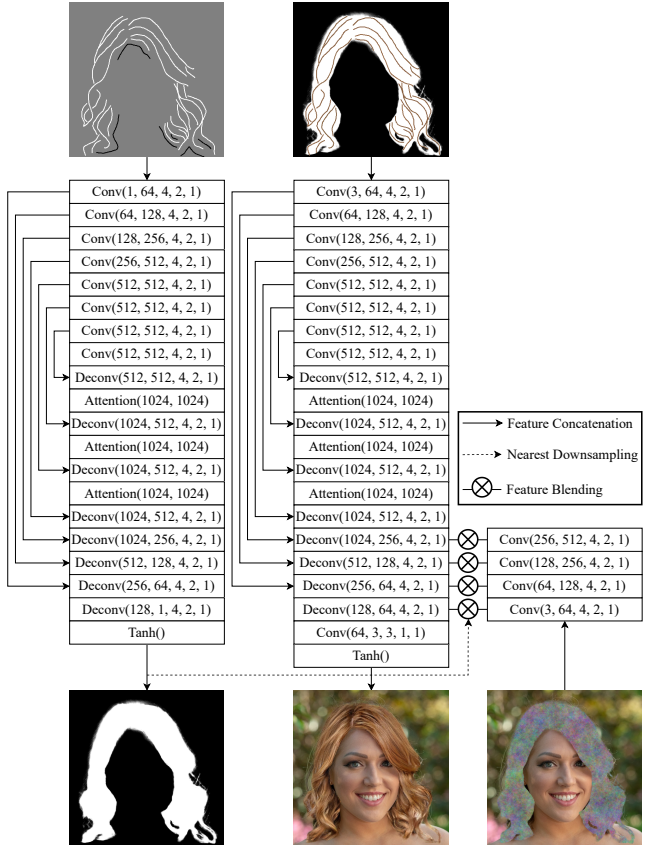

Fig. 2. Network architecture details. The left network is S2M-Net, while the right one is S2I-Net. The parameters of convolution and deconvolution layers are denoted as *Conv/Deconv(input_channel_number, output_channel_number, kernel_size, stride, padding_width)*, while those of the self-attention modules are *Attention(input_channel_number, output_channel_number)*. Original image courtesy of Charcharius.

- Rope:

$$\begin{cases} L_0 : x = a(t)\sin(wt) + \triangle x, y = \triangle y, z = b \cdot \sin(wt) \\ L_1 : x = a(t)\sin(wt + \pi) + \triangle x, y = \triangle y, z = b \cdot \sin(wt + 3\pi/4) \end{cases}$$ , (3)

- Three-strand:

$$\begin{cases} L_0 : x = a(t)\sin(wt) + \triangle x, \ y = \triangle y, z = b\sin(2wt) \\ L_1 : x = a(t)\sin(wt + 2\pi/3) + \triangle x, y = \triangle y, z = b\sin(2(wt + 2\pi/3)) \\ L_2 : x = a(t)\sin(wt + 4\pi/3) + \triangle x, y = \triangle y, z = b\sin(2(wt + 4\pi/3)) \end{cases}$$ , (4)

- Four-strand:

$$\begin{cases} L_0 : x = a(t)\sin(wt) + \triangle x, \ y = \triangle y, z = b \cdot f(t) \\ L_1 : x = a(t)\sin(wt + \pi/2) + \triangle x, y = \triangle y, z = b \cdot f(wt + \pi/2) \\ L_2 : x = a(t)\sin(wt + \pi) + \triangle x, y = \triangle y, z = b \cdot f(wt + \pi) \\ L_3 : x = a(t)\sin(wt + 3\pi/2) + \triangle x, y = \triangle y, z = b\sin(4(wt + 3\pi/2)) \end{cases}$$ , (5)

where

$$f(t) = \begin{cases} \sin(2t), \text{if } 2n\pi \leq t < 2n\pi + \pi, n \in \mathbb{Z} \\ \sin(4t), otherwise \end{cases}$$ (6)

- Five-strand:

$$\begin{cases} L_0 : x = a(t)\sin(wt) + \triangle x, \ y = \triangle y, z = b\sin(2wt) \\ L_1 : x = a(t)\sin(wt + 2\pi/5) + \triangle x, y = \triangle y, z = b\sin(4(wt + 2\pi/5)) \\ L_2 : x = a(t)\sin(wt + 4\pi/5) + \triangle x, y = \triangle y, z = b\sin(4(wt + 4\pi/5)) \\ L_3 : x = a(t)\sin(wt + 6\pi/5) + \triangle x, y = \triangle y, z = b\sin(4(wt + 6\pi/5)) \\ L_4 : x = a(t)\sin(wt + 8\pi/5) + \triangle x, y = \triangle y, z = b\sin(4(wt + 8\pi/5)) \end{cases}$$ , (7)

## 2 DATASET

Our dataset includes long (60%) and moderately short (40%) hair images on straight (1K), wavy (2K), and braided (1K) hairstyles on male (5%) and female (95%). We have many more female images since the female hairstyles exhibit significantly larger variety. Our system works for these hairstyles and their combinations on diverse structure and appearance. Figure 3 shows some representative examples in our dataset. Note that all of them are the testing results produced by our system given the sketches as input, including the generated mattes and the synthesized hair images. The original images are not shown due to the copyright issue.

## 3 EXPERIMENT

### 3.1 Matte VS. Mask

To show the effectiveness of hair mattes, we compare synthesized hair images based on the generated mattes with those based on the generated masks. We train two more networks with the same architectures (including the background blending module) as S2M-Net and S2I-Net, taking as input hair sketches to predict binary hair masks and further hair images. Figure 4 shows two examples of the matte-based and mask-based generation results. It is obvious that the generated mattes (Figure 4 (c)) can provide a more natural blending between hair and background regions, with fewer artifacts around their boundaries in the synthesized hair results (Figure 4 (e)), compared to the binary masks (Figure 4 (b)) and mask-based generated results (Figure 4 (d)), respectively.

In addition, we conducted one more perception study to evaluate the effectiveness of hair mattes for synthesizing hair images, compared to binary hair masks. We randomly picked 20 pairs of matte-based and mask-based generated hair images corresponding to the same sketch inputs, and invited 23 participants, each of which was asked to choose the better one in terms of visual naturalness from the two results in each pair (with the randomized presentation order of matte-based and mask-based results). Finally, we got 23 (participants) × 20 (pairs) = 460 subjective evaluations. The voting results show our matte-based method received significantly more votes (95.8% voting rate) than the mask-based one (4.2% voting rate). This is consistent with our findings on the qualitative results (Figure 4).

### 3.2 Comparisons on Sketch-based Hair Image Synthesis

We compare S2I-Net with the state-of-the-art methods for image synthesis conditioned on hair sketches, including pix2pix, HIS and MichiGAN in terms of the visual quality of hair image generation. For a fair comparison, the inputs to these methods are the same as those to S2I-Net, i.e., hair mattes, hair sketches and background

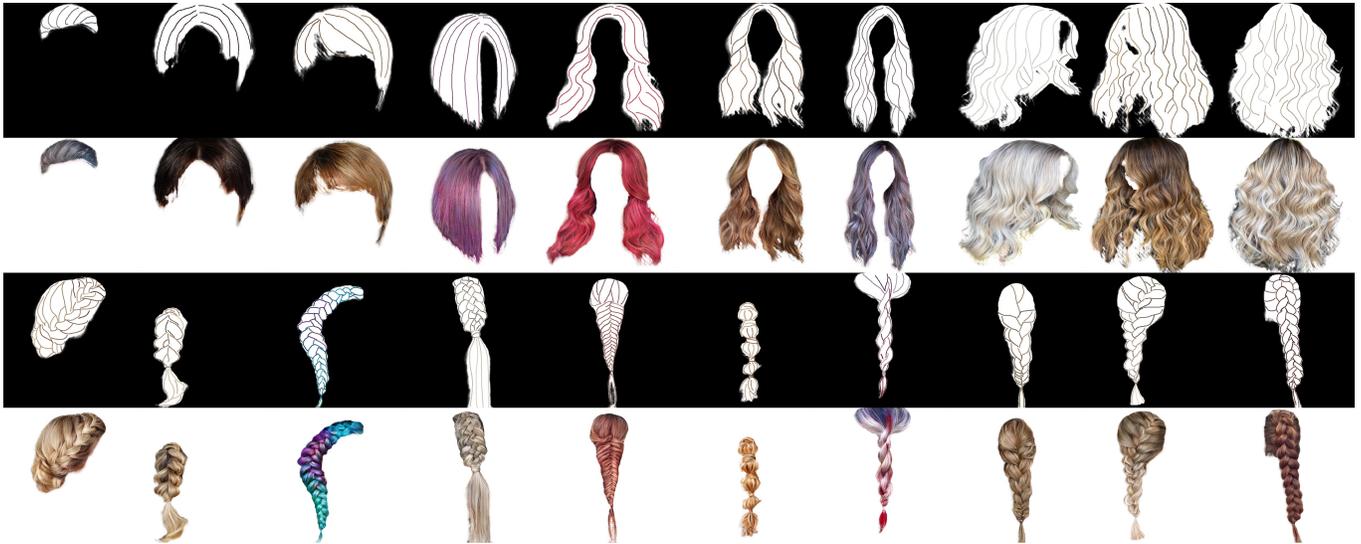

Fig. 3. Diverse hairstyles in our dataset. The first two rows are a set of unbraided hairstyles, while the last two rows are a set of braided hairstyles. In each set, the top row contains the input sketches and the predicted mattes, while the bottom row gives the generated hair images by our system. Due to the copyright issue, we show our generated hairstyles, which are very close to those in the original images.

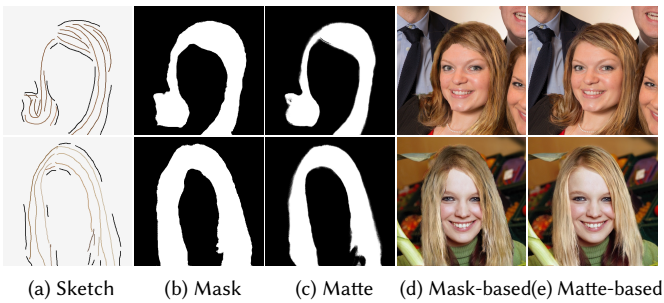

(a) Sketch  (b) Mask  (c) Matte  (d) Mask-based (e) Matte-based

Fig. 4. Comparison of mask-based (d) and matte-based (e) generated hair images. Note that both of the masks (b) and the mattes (c) are auto-generated given the same hair sketches (a). Please zoom in to examine the differences. Original images courtesy of The Cooperative Trust and Tim Reckmann.

regions, though the ways to control the appearance are slightly different between MichiGAN and the other methods.

To be specific, the binary hair masks of MichiGAN are replaced with our hair mattes. We find that the appearance encoder of MichiGAN could not be re-trained well from scratch due to our not very large-scale dataset. Figure 5 shows the testing results from MichiGANs trained on our dataset with different strategies, including trained from scratch and fine-tuned on the given pretrained model, given the ground truth as the reference. We also show the results generated by the pretrained model of MichiGAN and ours for reference. It shows that the pretrained model can perform better for restoring the appearance of the ground truth than that trained from scratch, due to the large-scale dataset for training the pretrained model. Thus, to achieve the best performance and mainly compare the quality of hair structure generation, we fine-tune the pre-trained MichiGAN on our dataset. Note that since MichiGAN actually controls local structures via orientation maps, we also re-train their sketch-to-orientation network on our dataset, of which the orientation map is extracted from the hair images like MichiGAN does. In addition, since MichiGAN requires a reference image as an appearance condition, we set the ground-truth images as the appearance inputs for training and testing.

Pix2pix and HIS take the totally same inputs as those to our S2I-Net, with the colored hair sketches and background regions. To remain their original network architectures, the two inputs are concatenated and fed together into their networks. HIS consists of two stages: its first stage is similar to pix2pix, i.e., colored sketch-to-image translation, while the second stage takes as inputs the colored sketches and the orientation maps as well as texture maps extracted from the first-stage output images.

Note that we separately train the unbraided model and braided model for all the compared methods (the same way to train our models) until convergence. Besides the qualitative results shown in our main paper, we provide more results for reference, as shown in Figure 6.

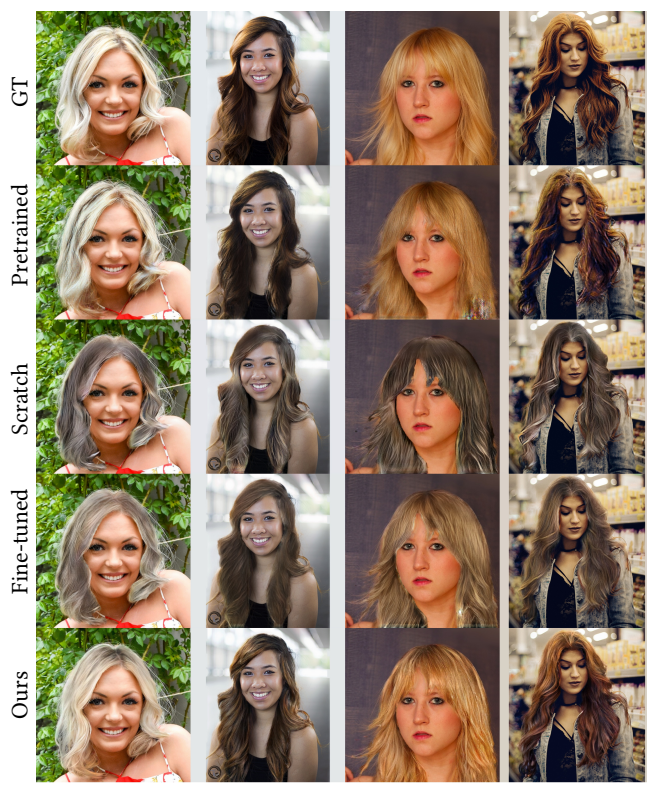

Fig. 5. Comparisons of the results generated by MichiGAN trained with different strategies, as well as the provided pretrained model, given the ground truth as the reference. We also show ours result for reference. Original images courtesy of Oscar Campos-Morales, Tony Au, Kerry Goodwin, and Allef Vinicius.


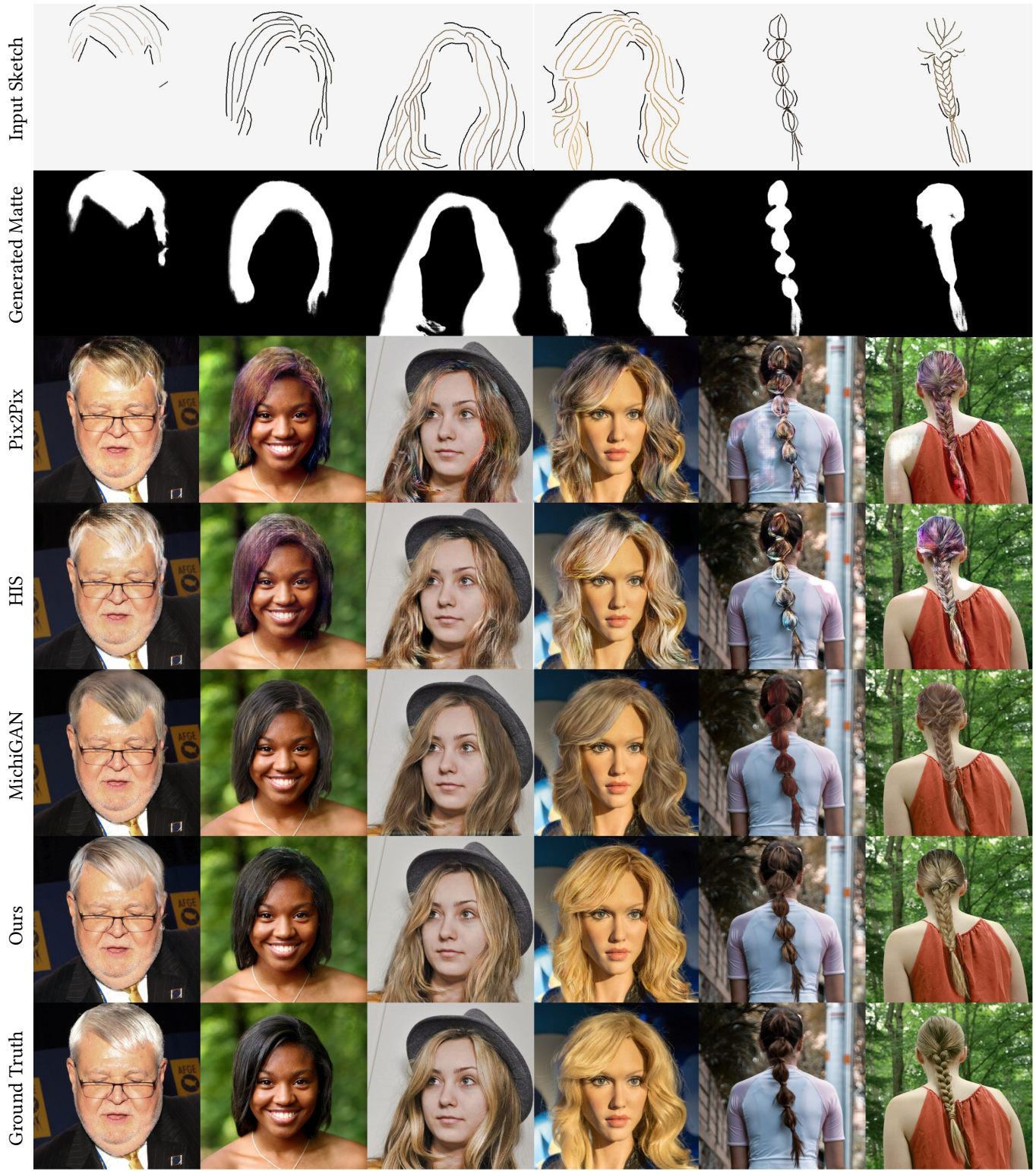

Fig. 6. Comparisons of structure reconstruction with the state-of-the-art methods given the same input sketches (Top Row) and their generated mattes S2M-Net (Second Row). The sketches contains hair strokes and non-hair strokes (in black), of which the non-hair strokes are only used for generating hair mattes but removed for hair image synthesis. The left four columns are unbraided hairstyles, while the rest are braided hairstyles. Please zoom in to better examine the quality of synthesized results by the compared methods against the ground truth. Original images courtesy of AFGE, Pawel Loj, Noah Diamond, Bigashb, Munbaik Cycling Clothing and Tsunami Green.


\end{document}